\documentclass{article}

\usepackage{microtype}
\usepackage{graphicx}
\usepackage{subcaption}
\usepackage{booktabs} % for professional tables

\usepackage{hyperref}

\usepackage[accepted]{icml2026}

\usepackage{amsmath}
\usepackage{amssymb}
\usepackage{mathtools}
\usepackage{amsthm}

\usepackage[capitalize,noabbrev]{cleveref}

\theoremstyle{plain}

\theoremstyle{definition}

\theoremstyle{remark}

\usepackage[textsize=tiny]{todonotes}
\usepackage{enumitem}
\usepackage{booktabs}       % professional-quality tables
\usepackage{amsfonts}       % blackboard math symbols
\usepackage{nicefrac}       % compact symbols for 1/2, etc.
\usepackage{microtype}      % microtypography
\usepackage{natbib}
\usepackage{algorithmic}
\usepackage{amssymb}
\usepackage{mathtools}
\usepackage{caption}
\usepackage{amsmath}
\usepackage{multirow}
\usepackage{adjustbox}
\usepackage{tabularx}
\usepackage{multirow}
\usepackage{color}
\usepackage{rotating}
\usepackage{enumitem}
\usepackage{makecell}
\usepackage{array}
\usepackage{etoc}
\usepackage{xcolor}

\renewcommand{\algorithmiccomment}[1]{\hfill\textcolor{gray}{\scriptsize// #1}}
\usepackage{comment}
\usepackage{subcaption}
\usepackage[acronym]{glossaries}
\usepackage{pifont}   % For check marks and X marks
\newcommand{\cmark}{\ding{51}}% Checkmark
\newcommand{\xmark}{\ding{55}}% Xmark
\glsdisablehyper
\definecolor{darkgray}{gray}{0.45}
\newcounter{rownumbers}
\newcommand{\rownum}{{\color{darkgray}\stepcounter{rownumbers}\arabic{rownumbers}.}}
\icmltitlerunning{Causal-Adapter: Taming Text-to-Image Diffusion for Faithful Counterfactual Generation}

\begin{document}

\twocolumn[
  \icmltitle{Causal-Adapter: Taming Text-to-Image Diffusion for \\ Faithful Counterfactual Generation}



  \icmlsetsymbol{equal}{*}

  \begin{icmlauthorlist}
    \icmlauthor{Lei Tong}{equal,affili1}
    \icmlauthor{Zhihua Liu}{equal,affili2,affili3}
    \icmlauthor{Chaochao Lu}{affili4}
    \icmlauthor{Dino Oglic}{affili1}
    \icmlauthor{Tom Diethe}{affili1}
    \icmlauthor{Philip Teare}{affili1}
    \icmlauthor{Sotirios A. Tsaftaris}{affili2,affili3}
    \icmlauthor{Chen Jin}{affili1}
  \end{icmlauthorlist}

  \icmlaffiliation{affili1}{Centre for AI, DS\&AI, Astrazeneca, UK}
  \icmlaffiliation{affili2}{Causality in Healthcare AI Hub (CHAI), UK}
  \icmlaffiliation{affili3}{Institute for Imaging, Data and Communications (IDCOM), School of Engineering, University of Edinburgh, Edinburgh, UK}
  \icmlaffiliation{affili4}{Shanghai Artificial Intelligence Laboratory}

  \icmlcorrespondingauthor{Chen Jin}{chen.jin@astrazeneca.com}

  \icmlkeywords{Machine Learning, ICML}

  \vskip 0.3in
]



\printAffiliationsAndNotice{}  % no special notice (required even if empty)

\begin{abstract}
  We present Causal-Adapter, a modular framework that adapts frozen text-to-image diffusion for counterfactual generation. Our method enables causal interventions on target attributes, consistently propagating their effects to causal dependents without altering the core identity of the image. In contrast to prior approaches that rely on prompt engineering without explicit causal mechanism, Causal-Adapter leverages structural causal modeling augmented with two attribute regularization strategies: prompt-aligned injection, which aligns causal attributes with textual embeddings for precise semantic control, and a conditioned token contrastive loss to disentangle attribute factors and reduce spurious correlations. Causal-Adapter achieves state-of-the-art performance on both synthetic and real-world datasets, with up to 91\% MAE reduction on Pendulum for accurate attribute control and 87\% FID reduction on ADNI for high-fidelity MRI generation. These results show that our approach enables robust, generalizable counterfactual editing with faithful attribute modification and strong identity preservation. Code, training data, and models are available on the \href{https://leitong02.github.io/causaladapter/}{project page}.

\end{abstract}

\section{Introduction}

\begin{figure}[t]
%\vspace{-18mm}
\centering
\includegraphics[width=\linewidth]{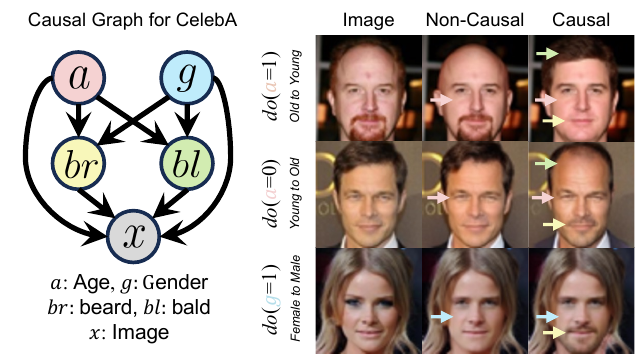}
%\vspace{-7mm}
\caption{Non-causal editing modifies only the target attribute (\textit{e.g.} age, gender); causal editing propagates changes to related attributes (\textit{e.g.} beard, baldness) enforced by the causal graph.}
\label{fig:ctf_generate}
\vspace{-3mm}
\end{figure}

Answering counterfactual questions~(\textit{e.g.} inferring what an event would have happened under an alternative action) requires understanding the cause–effect relationships among variables and performing hypothetical reasoning~\citep{pearl2009causality,scholkopf2021toward,weinberg2024causality}.
~Classical generative models typically tackle counterfactual-style tasks such as image editing or style transfer in a non-causal perspective. \textbf{Taking a pre-defined causal graph and semantic attributes as standard inputs}, subsequent work augments such models with explicit structural causal models (SCMs) to implement \textit{abduction–action–prediction}~\citep{pearl2013structural,pawlowski2020deep,pmlr-v202-de-sousa-ribeiro23a,komanduri2024causal,rasal2025diffusion}. 
This design drives advances in counterfactual image generation (Figure~\ref{fig:ctf_generate}) by enforcing edits consistent with a causal graph which enables critical domain-specific counterfactual generation applications, such as simulating medical images with fine-grained anatomical changes associated with aging or disease progression to improve clinical interpretation~\citep{starck2025diff}. Faithful counterfactual generation remains challenging, as real-world attributes are often causally entangled (\textit{e.g.}, only males can grow beards). This entanglement complicates disentangling factors and generalizing edits, making it nontrivial to ensure that interventions yield the intended visual effect while keeping non-intervened attributes invariant and preserving identity-specific details~\citep{komanduri2023identifiable}.
\begin{figure*}[ht]
    \centering
    \includegraphics[width=\linewidth]{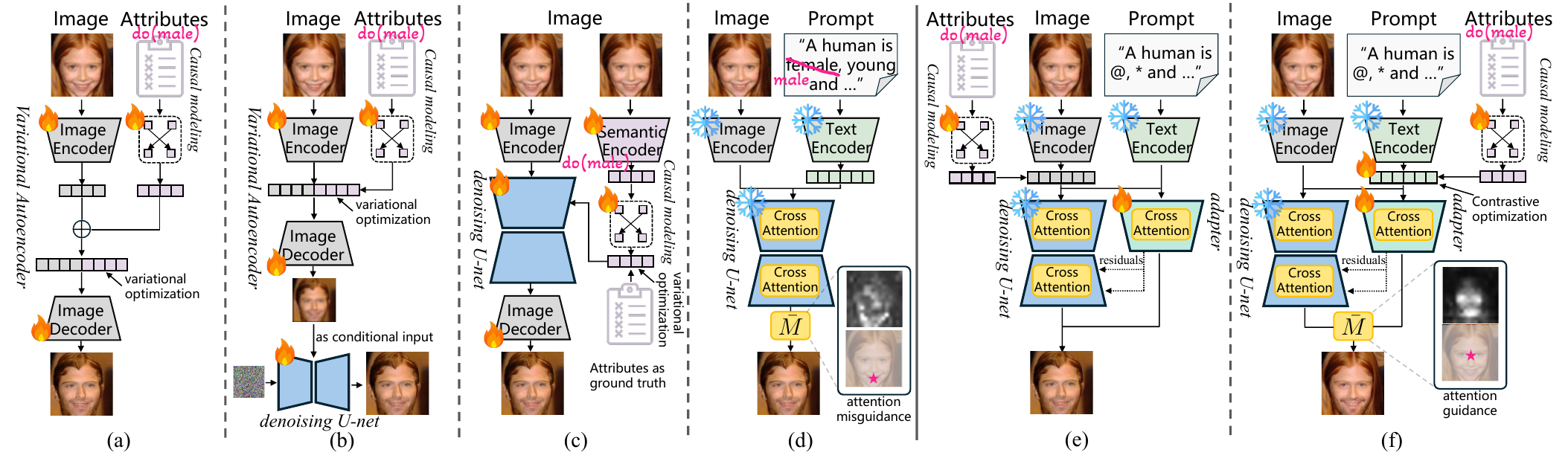}
    \vspace{-7mm}
    \caption{A sketch comparison of counterfactual image generation methods based on:~(a) \textbf{VAE or GAN}, which fail to achieve high-fidelity results. 
    (b) \textbf{Diffusion SCM} and (c) \textbf{Diffusion autoencoder}, which are sensitive to spurious correlations. 
    (d) \textbf{T2I based editing}, which requires heavy prompt engineering. 
    (e) \textbf{Vanilla Causal-Adapter}, which injects causal attributes into image-embedding. 
    (f) \textbf{Causal-Adapter with attribute regularization}, which injects causal attributes into learnable textual embeddings with contrastive optimization. Detailed discussion is presented in Appendix~\ref{Appendix:Related Works}.}
    \label{fig:key_comparison}
\end{figure*}

\textbf{Counterfactual Image Generation.}~Early approaches modeled counterfactual image generation using normalizing flows~\citep{papamakarios2021normalizing,winkler2019learning}, variational autoencoders (VAEs)~\citep{kingma2013auto,yang2021causalvae}, hierarchical VAEs~\citep{vahdat2020nvae,pmlr-v202-de-sousa-ribeiro23a}, and generative adversarial networks (GANs)~\citep{goodfellow2014generative,kocaoglu2017causalgan}, and encouraged attribute disentanglement through variational objectives~\citep{higgins2017beta}. However, variational optimization inevitably introduces uncertainty into the latent space, which can lead to posterior collapse of meaningful factors, creating a trade-off between image fidelity and controllable attribute manipulation (Figure~\ref{fig:key_comparison}a). Recent works integrate diffusion models with the SCM, capitalizing on its high perceptual quality to explore counterfactual identifiability~\citep{sanchez2022diffusion,rasal2025diffusion,komanduri2024causal,pan2024counterfactual,xia2025decoupled}. Despite domain-specific tuning, previous approaches perform disentanglement only in auxiliary encoders, which has limited effect on diffusion latents and leads to incomplete disentanglement (Figure~\ref{fig:key_comparison}b and c). This makes models prone to spurious correlations, where target factor interventions often cause unintended changes in non-intervened attributes.

%Despite domain-specific tuning, these methods remain prone to spurious correlations: interventions on a target factor often induce unintended changes in non-intervened attributes. Previous methods only perform disentanglement in auxiliary encoders, which has limited influence on diffusion latents with incomplete disentanglement (Figure~\ref{fig:key_comparison}(b) and (c)). 

\textbf{Text-to-Image based Editing.}~An alternative perspective is to view counterfactual image generation as a text-to-image (T2I) based editing, which typically relies on an inversion process~\citep{song2020denoising}. The aim is to generate a target edited image from projected latent states with condition manipulation~\citep{hertz2023prompttoprompt,ho2022classifier}. To reduce reconstruction error and preserve essential contents, several methods optimize either the unconditional text embedding~\citep{mokady2023null,xu2024inversion,miyake2025negative,ju2023direct,dong2023prompt}, or learn conditional concept embeddings for better attribute disentanglement~\citep{gal2023an,vinker2023concept,jin2024image}. 
However, generic T2I based editing remains insufficient for counterfactual generation. Existing methods heavily rely on carefully engineered inversion prompts to obtain reliable editing guidance. These approaches lack an explicit, learnable SCM over semantic attributes, making them difficult to guarantee both causal faithfulness and identity preserved counterfactual image generation~(Figure~\ref{fig:key_comparison}d). A broader discussion of related works can be found in Appendix~\ref{Appendix:Related Works}.
\begin{figure*}[ht]
    \centering
    \includegraphics[width=\linewidth]{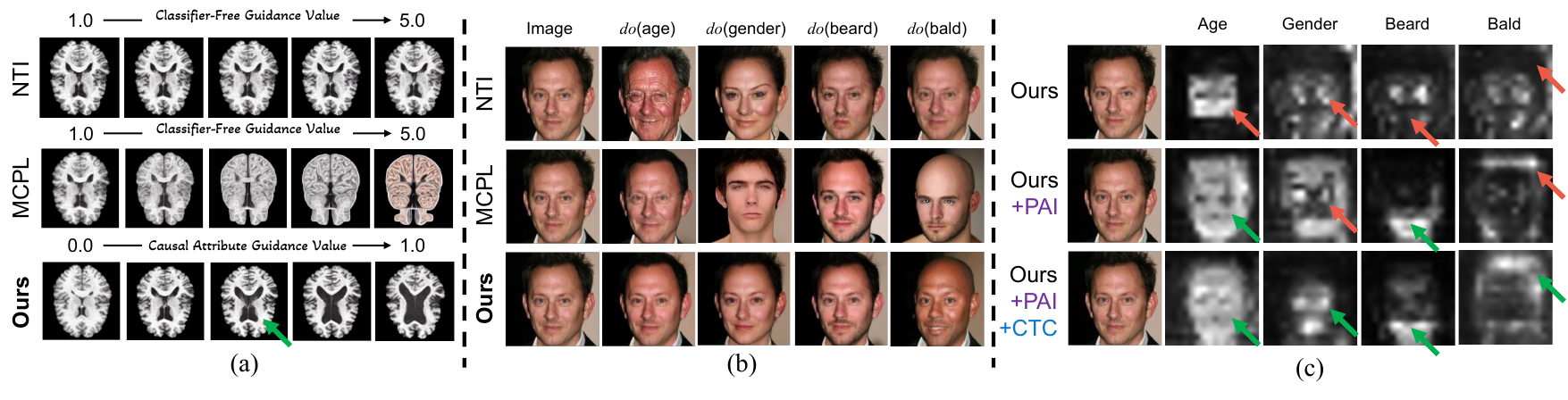}
     \caption{\textbf{Motivational study and preliminary counterfactual generation results between T2I methods and Causal-Adapter.} (a) Fine-grained anatomical counterfactual editing of brain ventricular volume using inversion-based editing (NTI~\citep{mokady2023null}), multi-concept prompt-learning editing (MCPL~\citep{jin2024image}), and our approach. (b) Comparison of counterfactual editing results on human faces. (c) Cross-attention maps: base vs. regularized Causal-Adapter. Details are presented in Appendix~\ref{Appendix:Full Motivational Study Results}.}
    \label{fig:motivational_study}
\end{figure*}

Herein, we propose~\textbf{Causal-Adapter}, an adaptive and modular framework that tames text-to-image diffusion model, such as Stable-Diffusion (SD)~\citep{rombach2022high}, for counterfactual generation. Unlike prior diffusion-based methods that require considerable re-training or fine-tuning ~\citep{sanchez2022diffusion,komanduri2024causal,pan2024counterfactual,rasal2025diffusion}, our method simply injects causal semantic attributes into a frozen backbone via a pluggable adapter~(Figure~\ref{fig:key_comparison}e). Inspired by advances in controllable diffusion~\citep{zhang2023adding,zhao2023uni,mou2024t2i,li2024dispose}, we investigate integrating semantic attributes with image embeddings as additional conditions. We find that naive fusion fails to achieve sufficient disentanglement or semantic alignment with spatial features in the diffusion latents. To address this, we introduce two regularization strategies that align causal semantic attributes with textual embeddings, improving disentanglement and causal faithfulness~(Figure~\ref{fig:key_comparison}f). Our main contributions are summarized as follows:
\begin{itemize}[leftmargin=*]
    \item Our motivational studies revealed that relying solely on a frozen text-to-image diffusion model with prompt tuning is insufficient for counterfactual image generation, as it fails to jointly represent causal semantic attributes and image embeddings, leading to imprecise reasoning and edits. This underscores the need to inject causal semantics into diffusion models.
    
    \item We propose Causal-Adapter, a  framework that employs an adapter encoder to learn causal interactions between semantic variables. These interactions are injected into a frozen diffusion and jointly optimized, introducing a dynamic prior for faithful counterfactual generation.
    
    \item To enhance causal disentanglement between semantic variables, we introduce two regularization strategies: Prompt-Aligned Injection (PAI) and Conditioned Token Contrastive Loss (CTC). These strategies separate token embeddings across conditions, enhancing causal representation learning and reducing spurious correlations for more precise counterfactual reasoning.
    
    \item We validate Causal-Adapter through extensive experiments on both synthetic and real-world datasets, including human face editing and medical image generation. Our method consistently achieves state-of-the-art performance across key metrics: counterfactual effectiveness (up to $50\%$ MAE reduction on ADNI), realism ($81\%$ FID reduction on CelebA), composition ($86\%$ LPIPS reduction on CelebA), and minimality ($4\%$ CLD reduction on ADNI).
\end{itemize}

\section{Methodology}
\label{sec:methodology}
% We first present the preliminaries in Section~\ref{subsec:preliminary}, followed by a motivational study in Section~\ref{subsec:motivational}. This study examines the systematic limitations of current T2I based editing methods for counterfactual generation and emphasizes the need to leverage causal semantic attributes for producing faithful counterfactuals (Figure~\ref{fig:motivational_study}). Motivated by these findings, we further introduce the proposed Causal-Adapter in Section~\ref{subsec:causal-adapter} and describe our regularization strategies in Section~\ref{subsec:conditional_disentanglement}, which further enhance counterfactual generation through semantic disentanglement.

\subsection{Preliminaries}
\label{subsec:preliminary}
\paragraph{Structural Causal Model (SCM).}
SCM provides a formal framework for modeling causal relationships between variables~\citep{pearl2010causal}. An SCM is defined as a triplet $\mathcal{S} = \langle Y, F, U \rangle$, where $Y = \{y_i\}_{i=1}^K$ denotes a set of $K$ endogenous (observed) variables, $U = \{u_i\}_{i=1}^K$ is a set of exogenous (latent) variables, and $F = \{f_i\}_{i=1}^K$ is a set of deterministic functions defining $y_i = f_i(\mathrm{Pa}_i, u_i)$, with $\mathrm{Pa}_i \subseteq Y\setminus \{y_i\}$ denoting the parent variables of $y_i$. The structural assignments of the SCM induce a directed acyclic graph (DAG) $\mathcal{G}$, where each node represents a variable $y_i$, and each edge $\mathrm{Pa}_i \rightarrow y_i$ represents a direct causal dependency between variables $\mathrm{Pa}_i$ and $y_i$. A SCM is called Markovian if the exogenous variables are mutually independent $p(U) = \prod_{i=1}^K p(u_i)$. A Markovian causal model induces the unique joint observational distribution that satisfies the causal Markov condition $p_{\mathcal{S}}(Y) = \prod_{i=1}^K p(y_i \mid \mathrm{Pa}_i)$. 

\paragraph{Counterfactual Reasoning.}
Counterfactual reasoning aims to answer queries ``Given observation $Y$, what would $y_i$ have been if its parents $\mathrm{Pa}_i$ had been different?'' This is formalized by the \textit{abduction–action–prediction} procedure~\citep{pearl2013structural,pawlowski2020deep,shen2022weakly}: 1. \emph{Abduction}: inferring posterior over exogenous variables consistent with observation, $p_{\mathcal{S}}(U \mid Y)$. 2. \emph{Action}: perform an intervention $do(y_i = \tilde{y_i})$, which replaces the structural function $f_i$ with a constant assignment. This yields a modified SCM $\tilde{\mathcal{S}} = \langle \tilde{Y}, \tilde{F}, U \rangle$, where $\tilde{Y}$ and $\tilde{F}$ denote the modified endogenous variables and structural mechanisms respectively. 3. \emph{Prediction}: compute the counterfactual outcome by evaluating $p_{\tilde{\mathcal{S}}}(\tilde{Y})$ under the modified structural mechanism and the inferred exogenous noise.
% \begin{enumerate}[leftmargin=*]
%     \item \emph{Abduction}: inferring posterior over exogenous variables consistent with observation, $p_{\mathcal{S}}(U \mid Y)$.
    
%     \item \emph{Action}: perform an intervention $do(y_i = \tilde{y_i})$, which replaces the structural function $f_i$ with a constant assignment. This yields a modified SCM $\tilde{\mathcal{S}} = \langle \tilde{Y}, \tilde{F}, U \rangle$, where $\tilde{Y}$ and $\tilde{F}$ denote the modified endogenous variables and structural mechanisms respectively.

%     \item \emph{Prediction}: compute the counterfactual outcome by evaluating $p_{\tilde{\mathcal{S}}}(\tilde{Y})$ under the modified structural mechanism and the inferred exogenous noise.
% \end{enumerate}
\begin{figure*}
    \centering
    \includegraphics[width=\linewidth]{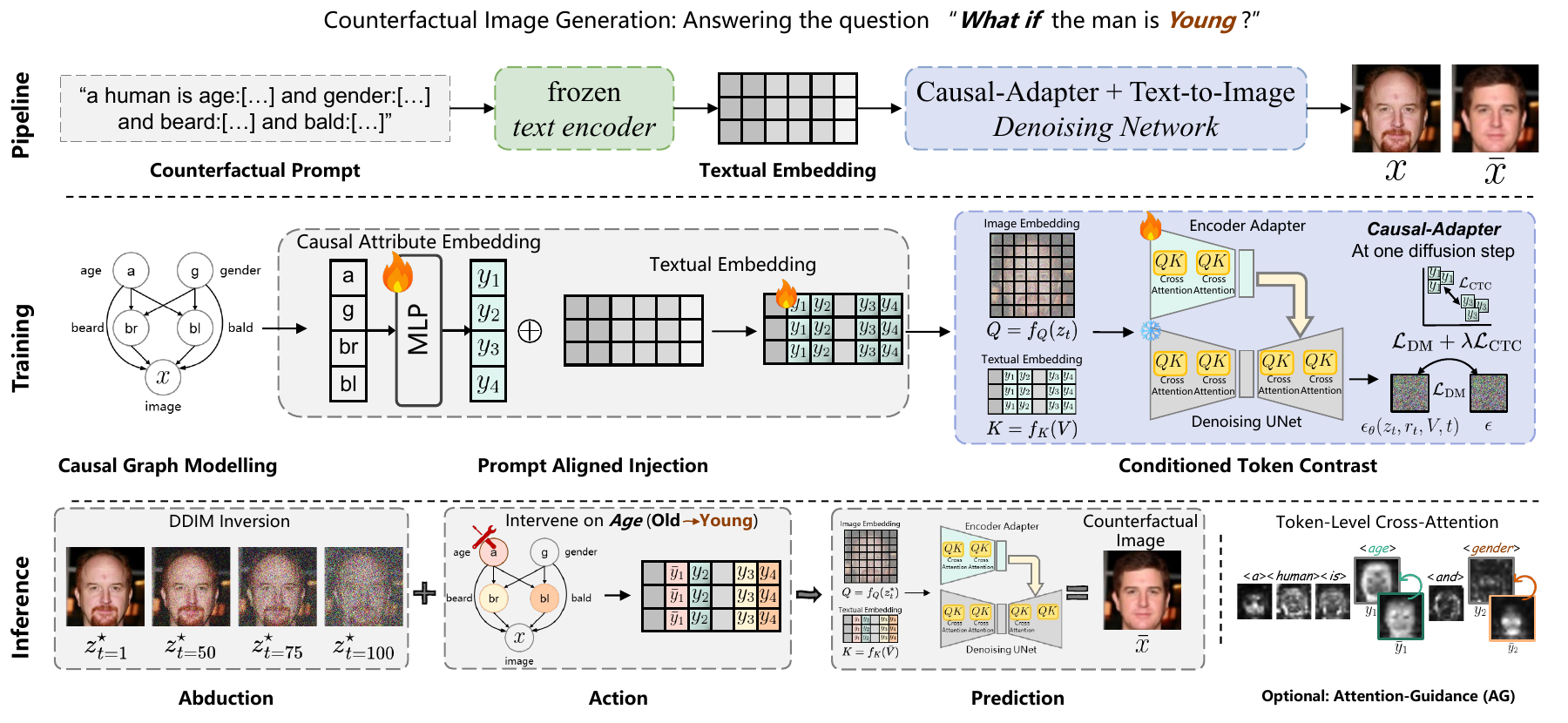}
    \caption{Method overview. A counterfactual prompt and input image $x$ are fed into a pretrained text-to-image diffusion model with a learnable Causal-Adapter $\ddot{\epsilon}_\psi$. Causal mechanisms, modeled over a known causal graph and attributes ${y_i}$, are injected into token embeddings via \textit{Prompt-Aligned Injection (PAI)} to align semantic and spatial features. The adapter $\ddot{\epsilon}_\psi$ operates alongside the frozen diffusion U-Net $\epsilon_\theta$, optimized with MSE $\mathcal{L}_{\text{DM}}$ and a \textit{Conditioned Token Contrastive (CTC)} loss $\mathcal{L}_{\text{CTC}}$ to enforce disentanglement. At inference, interventions on $y_i$ update token embeddings, and the counterfactual $\bar{x}$ is generated using the abducted exogenous noise $z^{\star}_{t}$. Optionally, Attention Guidance (AG) updates the cross-attention map of intervened tokens (\textit{e.g.} age, beard, bald) to achieve localized editing and preserving non-intervened attributes identity (\textit{e.g.} human, gender).}
    %Optionally, Attention Guidance (AG) localizes edits by updating intervened tokens while preserving non-intervened attention weights.}
    \label{fig:framework}
\end{figure*}
\subsection{Motivational Study}
\label{subsec:motivational}

To examine the generalization limits of existing text-to-image (T2I) based models for counterfactual image generation, we conducted a motivational study on two real-world datasets: CelebA for human faces and ADNI for brain MRI. Preliminary results are shown in Figure~\ref{fig:motivational_study}, including counterfactual image generation based on different conditional signals in brain volume MRI, as well as interventions on different target attributes in the CelebA dataset. Our study reveals two systematic limitations of directly applying T2I based models to counterfactual generation:
\begin{itemize}[leftmargin=*]
    \item \textbf{Text-only prompting is inadequate for counterfactual generation.} Counterfactual generation should yield consistent visual changes when a single attribute is intervened. However, current T2I based models ignore continuous attributes, making fine-grained edits infeasible, and hence particularly concerning for safety-critical domains such as medical imaging (Figure~\ref{fig:motivational_study}a).
    
    \item \textbf{Existing T2I based counterfactual generation suffer from attribute entanglement.} Current methods often confuse (entangle) unrelated attributes or require instance-specific fine-tuning, underscoring the need for explicit causal modeling and controllable semantic representations for faithful counterfactual generation (Figure~\ref{fig:motivational_study}b).
\end{itemize}

\subsection{Overview of Causal-Adapter}
\label{subsec:causal-adapter}
Recent advances on controllable diffusion methods~\citep{zhang2023adding,mou2024t2i,li2024dispose} show that a frozen T2I diffusion backbone can be steered by auxiliary control signals (\textit{e.g.}, segmentation masks or human poses) supplied through a trainable side module. We adopt the same high-level recipe by treating causal semantic attributes as the auxiliary control signals. The causal mechanism between attributes is then learned and injected explicitly into the diffusion backbone via a compact modular encoder that we term as \emph{Causal-Adapter} (Figure~\ref{fig:framework}).

\paragraph{Causal Mechanism Modeling.}  
Let $x$ denote an image and $Y = \{y_i\}_{i=1}^K$ denote a vector of semantic variables, where each $y_i$ represents a scalar value corresponding to a high-level semantic attribute. We assume a known causal graph $\mathcal{G}$ encodes the causal relationships among the variables in $Y$. Let $A \in \{0,1\}^{K \times K}$ be the binary adjacency matrix of $\mathcal{G}$, where the $i$-th row $A_i \in \{0,1\}^K$ indicates the parent variables $\mathrm{Pa}_i$ of the $i$-th attribute $y_i$, \textit{i.e.}, $A_{ij} = 1$ if and only if $y_j \in \mathrm{Pa}_i$. We model each causal mechanism $f_i$ using a nonlinear additive noise model such that:
\begin{equation}
\label{eqn:anm}
    \bar{y}_i \coloneqq f_i(\mathrm{Pa}_i, u_i) = f_i(A_i \odot Y; \omega_i) + u_i, \quad u_i \sim \mathcal{N}(0, \sigma_i^2),
\end{equation}
where $\omega_i$ are the learnable parameters and $\odot$ denotes element-wise multiplication. $\bar{y}_i$ is the model’s prediction of $y_i$ under the change of parent variables $\mathrm{Pa}_i$. For root nodes ($A_i=0$), we simply set $\bar{y}_i=y_i$ as no causal parents exist.  
The row vector $A_i$ acts as a binary mask on $Y$, passing only the true parents of $y_i$ to $f_i$.  
To estimate the parameters of mechanisms $F=\{f_i\}_{i=1}^{K}$ and the noise variances $\{\sigma_i\}_{i=1}^K$, we minimize the negative log-likelihood of the observations:
\begin{equation}
\label{eq:nll}
\begin{aligned}
\mathcal{L}_{\text{NLL}}
&= 
-\sum_{i=1}^{K}\!\log p\bigl(y_i \mid f_i(\mathrm{Pa}_i, u_i)\bigr)
\\
&=\;
\frac{1}{2}\sum_{i=1}^{K}\frac{\lVert y_i-\bar{y}_i\rVert_2^{2}}{\sigma_i^{2}}
\;+\; \log(2\pi\sigma_i^2).
\end{aligned}
\end{equation}

Unlike prior methods that infer endogenous and exogenous from latent image features~\citep{yang2021causalvae,komanduri2023learning}, we operate directly in the observed semantic attribute space.  
This reduces reliance on the quality of learned visual representations and enables exact, interpretable $do$-interventions on attributes. The mechanism modeling can be transferred across domains by simply replacing the attribute set and causal graph $\mathcal{G}$. Moreover, it is modular that any differentiable alternatives (\textit{e.g.}, DeepSCM~\citep{pawlowski2020deep}) can be seamlessly integrated.

\paragraph{Integrating Causal Conditions into a Frozen T2I Backbone.}
Prior attempts have treated causal attributes as the conditioning signal to the diffusion backbone, which is computationally expensive as it requires retraining the entire model when switching across datasets~\citep{sanchez2022diffusion,komanduri2024causal,rasal2025diffusion}. To overcome this limitation, we introduce a branch module that delivers causal guidance without modifying the backbone parameters to preserve the flexibility and scalability of a pretrained T2I model. Given the image latent representation $z_t=\mathcal{E}(x,t)$ at diffusion step $t$ and the text embeddings $V$, we instantiate a half-scale replica $\ddot{\epsilon}_{\psi}$ of the denoiser $\epsilon_\theta$. This replica specializes $\ddot{\epsilon}_{\psi}$ in injecting control signals and is parameterized by $\psi$. The causal attributes $Y$ are added to $z_t$ and processed by $\ddot{\epsilon}_{\psi}$: 
\begin{equation}
\label{eq:controlnet}
r_t \coloneqq \ddot{\epsilon}_{\psi}\bigl(z_t \oplus Y,\,t,\,V\bigr),
\end{equation}
the residuals $r_t$ are injected into the mid- and up-sampling blocks of the frozen denoiser $\epsilon_{\theta}$. During training, we keep all backbone parameters fixed and optimize only $\psi$ with a T2I noise prediction objective.
\paragraph{Counterfactual Image Generation.} We implement counterfactual generation under the abduction–action–prediction procedure via DDIM inversion and sampling~\citep{song2020denoising}. Let $H_{\theta}(z_t,V,t)$ denote the DDIM inversion operator and let $H_{\theta}^{-1}$ be its generative inverse. The procedure is as follows: (1)~\emph{Abduction:} given an observed image $x$, we inject the condition residuals $r_t$ and run $H_{\theta}(z_{t=0}, r_t, V, t)$ to recover the corresponding exogenous noise $z_T^\star$.  (2)~\emph{Action:} we intervene $y_i$ by setting $y_i\!\gets\!\bar{y_i}$ via $do(y_i=\bar{y_i})$, propagate the attribute change through Eqn.~\ref{eqn:anm} and Eqn.~\ref{eq:controlnet} to obtain the updated $\bar{Y}=\{\bar{y_i}\}_{k=1}^K$ and its residuals $\bar{r_t}$. (3)~\emph{Prediction:} the counterfactual outcome is generated with $H_{\theta}^{-1}(z_{t=T}^{\star},\bar{r}_t,V,t)$. Optionally, classifier-free guidance (CFG)~\citep{ho2022classifier} can be applied to amplify the counterfactual signal with guidance weight $\alpha$:
\begin{equation}
\label{eq:cfg}
\tilde{\epsilon}_{\theta}(z_t,\bar{r_t},t,V,\varnothing)=\alpha\,\epsilon_{\theta}(z_t,\bar{r_t},t,V)+(1-\alpha)\,\epsilon_{\theta}(z_t,r_t^{\varnothing},t,\varnothing),
\end{equation}
where $\varnothing = c_{\phi}(``\ ")$ is the null-text embedding, and
$
r_t^{\varnothing} \coloneqq \ddot{\epsilon}_{\psi}(z_t \oplus \bar{Y},\, t,\, \varnothing)
$
are the residuals computed under the null-text condition.

\subsection{Regularizing the Causal-Adapter}
\label{subsec:conditional_disentanglement}

A key challenge in counterfactual generation is ensuring that interventions modify only the intended attribute while leaving others unaffected. Prior work highlights that achieving such disentanglement in the latent space is crucial for faithful counterfactuals~\citep{yang2021causalvae,shen2022weakly,komanduri2023learning,pmlr-v202-de-sousa-ribeiro23a,komanduri2024causal,rasal2025diffusion}.  To assess the degree of disentanglement in a multi-modal setting, we use cross-attention maps to reveal attribute separation in latent space.
Figure~\ref{fig:motivational_study}c visualizes averaged cross-attention maps across total denoising time steps among all Causal-Adapter variants.
Plain Causal-Adapter, which directly injects causal attributes into the image embedding, fails to align attribute semantics with spatial features and disrupts attribute independence, resulting in entangled, off-target edits. To tackle this issue, we introduce two regularization terms that reinforce the conditional causal attribute disentanglement.

\paragraph{Prompt Aligned Injection.}
% Inspired by previous findings~\citep{yang2024diffusion}, the highly 'disentangled' words can induce cross attention layer to align the semantic and spatial features in to diffusion latents and let generation process can do semantic aligned generation. 
Causal conditions $Y$ are numeric attributes that lack direct alignment with pixel-level representations but are semantically closer to text embeddings. Inspired by prior finding~\citep{yang2024diffusion} that disentangled tokens in the prompt can guide cross-attention to align semantics with spatial structure in diffusion latents, we inject $Y$ through the prompt channel. This allows the cross-attention module to propagate attribute semantics into spatial image features during generation. We introduce a prompt-aligned injection (PAI) mechanism that maps each causal attribute to a learnable token embedding. Let $C =[\,c_1,\ldots,c_K\,]^{\top} \in \mathbb{R}^{K \times d}$ denote the embeddings of $K$ placeholder tokens with dimension $d$. Each attribute $y_i$ is mapped into the text space via a linear projector $g_i:\mathbb{R}\!\to\!\mathbb{R}^{d}$. 
We form the attribute-injected token embeddings:
\begin{equation}
\label{eq:pai_mechan}
\begin{aligned}
v_i(y_i) &= c_i + g_i(y_i), \qquad i=1,\ldots,K,\\
V(Y) &= \bigl[\,v_1(y_1),\ldots,v_K(y_K)\,\bigr]^{\top} \in \mathbb{R}^{K \times d}.
\end{aligned}
\end{equation}

The set $V(Y)$ serve as the conditioning input to both the frozen denoiser and the adapter via cross-attention.%, linking high-level causal attributes to image features. 
~During training, we jointly optimize the adapter $\psi$, the placeholders token embeddings $C$, and the projectors $G=\{g_i\}_{i=1}^K$ using the standard noise prediction loss $\mathcal{L}_{\text{DM}}$:
\begin{equation}
\label{eq:pai_trainable}
\min_{\psi,\,G,\,C}\;
\mathbb{E}_{z,r,Y,\epsilon,t}
\left[
\Big\|
\epsilon - \widehat{\epsilon}_{\theta,\psi}\!\big(z_t, r_t, V(\bar{Y}), t\big)
\Big\|_2^2\right],
\end{equation}
where $\widehat{\epsilon}_{\theta,\psi}$ denotes the U-Net prediction modulated by the adapter with PAI. At inference, the learned placeholders $C$ and projectors $G$ are reused to construct $V(\bar{Y})$ for counterfactual query. Notably, PAI only updates token-level embeddings without fine-tuning the CLIP tokenizer or pretrained text encoder, ensuring compatibility between counterfactual queries and existing T2I backbones.

\textbf{Conditioned Token Contrast (CTC).}
%Motivated by prior efforts that contrastive objectives improve concept separation and reduce cross-factor leakage~\citep{jin2024image,liu2025segment}, we introduce a token-conditioned contrastive loss that enforces each placeholder token to capture only a single causal factor, thereby reducing attribute entanglement and suppressing spurious correlations. It achieves this by enforcing within-token invariance (same attribute token across samples) and cross-token separation (different attribute tokens). 
Motivated by prior efforts that contrastive objectives improve concept separation and reduce cross-factor leakage~\citep{jin2024image,liu2025segment}, we introduce a token-conditioned contrastive loss that enforces each placeholder token to capture only a single causal factor. Given a batch with $B$ samples and $K$ attributes (tokens), PAI produces a batch of textual embeddings $\{v_{b}^{k}\}_{b=1,k=1}^{B,K}$. For a specific anchor $(b,k)$, positive pairs are same attribute token across samples $\{v_{b'}^{k}\mid b'\neq b\}$ and negative pairs are different attribute tokens across samples $\{v_{b}^{k'}\mid k'\neq k\}$. By enforcing inter-token invariance (positive pairs) and intra-token separation (negative pairs), CTC reduces attribute entanglement and suppresses spurious correlations. We implement the objective using InfoNCE~\citep{oord2018representation,chen2020simple}:
% \begin{equation}
% \label{eq:patc_total}
% \begin{aligned}
% \mathcal{L}_{\text{CTC}}
% &= \frac{1}{BK} \sum_{k=1}^{K} \sum_{b=1}^{B}
% \Bigg[
% -\frac{\operatorname{sim}\!\left(v_{b}^{k},\, v_{b'}^{k}\right)}{\tau}
% \\
% &\qquad\qquad
% + \log\!\Bigg(
% \sum_{\substack{k'=1\\ k'\neq k}}^{K}
% \sum_{\substack{b'=1\\ b'\neq b}}^{B}
% \exp\!\left(\frac{\operatorname{sim}\!\left(v_{b}^{k},\, v_{b'}^{k'}\right)}{\tau}\right)
% \Bigg)
% \Bigg]
% \end{aligned}
% \end{equation}

\begin{equation}
\label{eq:patc_total}
\resizebox{0.5\textwidth}{!}{$
\mathcal{L}_{\text{CTC}}
= \frac{1}{BK} \sum_{k=1}^{K} \sum_{b=1}^{B}
\Bigg[
-\frac{\operatorname{sim}\!\left(v_{b}^{k},\, v_{b'}^{k}\right)}{\tau}
+ \log\!\Bigg(
\sum_{\substack{k'=1\\ k'\neq k}}^{K}
\sum_{\substack{b'=1\\ b'\neq b}}^{B}
\exp\!\left(\frac{\operatorname{sim}\!\left(v_{b}^{k},\, v_{b'}^{k'}\right)}{\tau}\right)
\Bigg)
\Bigg]
$}
%\end{aligned}
\end{equation}

% \begin{equation}
% \label{eq:patc_total}
% \mathcal{L}_{\text{CTC}}
% =
% -\frac{1}{BK}\sum_{k=1}^{K}\sum_{b=1}^{B}
% \log\!\left(
% \frac{
% \exp\!\left(\operatorname{sim}\!\left(v_b^{k},\, v_{b'}^{k}\right)/\tau\right)
% }{
% \sum\limits_{\substack{k'=1\\ k'\neq k}}^{K}
% \sum\limits_{\substack{b'=1\\ b'\neq b}}^{B}
% \exp\!\left(\operatorname{sim}\!\left(v_b^{k},\, v_{b'}^{k'}\right)/\tau\right)
% }
% \right).
% \end{equation}
where $\operatorname{sim}$ denotes the cosine similarity and $\tau$ is the temperature. Our final training objective becomes:
\begin{equation}
\label{eq:final_loss}
\mathcal{L}
=
\mathcal{L}_{\text{DM}}
+
\lambda\,\mathcal{L}_{\text{CTC}},
\end{equation}
$\lambda$ denotes a scaling coefficient. The proposed regularizers improve semantic–spatial alignment in the latent space (Figure~\ref{fig:motivational_study}c) and reduce spurious correlations (Figure~\ref{fig:ablation_study_main}c). The learned attention maps can be leveraged for localized editing through attention-guided manipulation. Following ~\citet{ju2023direct}, interventions are applied only to cross-attention weights corresponding to targeted tokens, while attention for non-intervened tokens is preserved. This ensures that edits remain spatially localized (e.g., gender) without altering unrelated attributes (e.g., hairstyle),  improving identity preservation without introducing extra training objectives. Full algorithm is presented in Appendix~\ref{Appendix:sub_full_regularization_Algorithm}.

\section{Experiments}
\label{sec:experiments}
We conduct experiments on four counterfactual image generation datasets across different domains: the synthetic \textbf{Pendulum} dataset~\citep{yang2021causalvae}, the human-face dataset \textbf{CelebA}~\citep{liu2015faceattributes} and its high-resolution restoration~\textbf{CelebA-HQ}~\citep{karras2017progressive}, and the medical imaging dataset \textbf{Alzheimer’s Disease Neuroimaging Initiative (ADNI)}~\citep{petersen2010alzheimer}. We follow the benchmarking experimental settings~\citep{melistas2024benchmarking,komanduri2024causal,rasal2025diffusion} for fair comparison. During evaluation, we follow the official setups for each of the datasets and report:~(1) Effectiveness: whether the intervention succeeds, measured by pretrained classifiers using F1-score or MAE depending on the attribute type.~(2) Composition: reconstruction quality under null interventions, measured by MAE and LPIPS distance.~(3) Realism: visual quality of counterfactual images, evaluated with Fréchet Inception Distance (FID).~(4) Minimality: non-intervened attributes remain minimally affected, assessed with the Counterfactual Latent Divergence (CLD) metric. Implementation details are presented in Appendix~\ref{Appendix:data_implement}.
\begin{figure}[t]
\centering
\includegraphics[width=\linewidth]{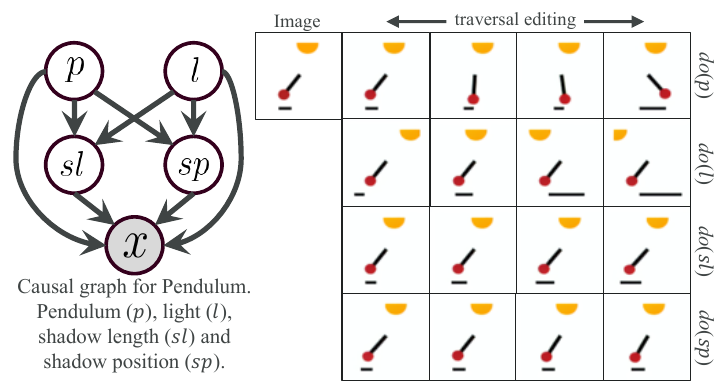}
\caption{Qualitative results on the Pendulum dataset demonstrating counterfactual traversals. When intervening on the pendulum angle or light position, our approach induces causally consistent changes in the descendant shadow attributes (length and position), in accordance with the underlying causal laws.}
\label{fig:pendulumn_quality1}
\end{figure}
\subsection{Results}

\setcounter{rownumbers}{0}
\begin{table*}[!]
\centering
\caption{
Intervention effectiveness on Pendulum test set. 
We report MAE from pretrained regressors under four interventions. w/ CM: with causal mechanisms; w/o CM: without causal mechanisms; w/ GT: ground-truth labels injected. “$\sim$” denotes descendant attributes remain unaffected. The table follows the causal graph: e.g., under the ``Pendulum $(p)$ MAE'', $do(p)$ or $do(l)$ report the pendulum MAE after intervening on $p$ or $l$.
}
\label{tab:pend_benchamrk_effectiveness}
\resizebox{\textwidth}{!}{
\begin{tabular}{rlcccccccc cccccccc}
\toprule
&\multirow{2}{*}{Method}
 & \multicolumn{4}{c}{\textbf{Pendulum $(p)$ MAE $\downarrow$}} 
 & \multicolumn{4}{c}{\textbf{Light $(l)$ MAE $\downarrow$}} 
 & \multicolumn{4}{c}{\textbf{Shadow Length $(sl)$ MAE $\downarrow$}} 
 & \multicolumn{4}{c}{\textbf{Shadow Position $(sp)$ MAE $\downarrow$}} \\
\cmidrule(lr){3-6}\cmidrule(lr){7-10}\cmidrule(lr){11-14}\cmidrule(lr){15-18}
&& $do(p)$ & $do(l)$ & $do(sl)$ & $do(sp)$ 
& $do(p)$ & $do(l)$ & $do(sl)$ & $do(sp)$ 
& $do(p)$ & $do(l)$ & $do(sl)$ & $do(sp)$ 
& $do(p)$ & $do(l)$ & $do(sl)$ & $do(sp)$  \\
\midrule
\rownum &CausalVAE     
& 24.86 & 23.03 & 20.47 & 11.58 
& 34.20 & 26.01 & 35.49 & 47.06 
& 1.946 & 1.430 & 2.020 & 1.720 
& 52.52 & 72.50 & 57.03 & 32.78 \\
\rownum &DisDiffAE        
& 0.668 & 0.648 & 0.647 & 0.647 
& 0.656 & 0.654 & 0.630 & 0.651 
& 0.550 & 0.527 & 0.560 & 0.516 
& 0.474 & 0.475 & 0.479 & 0.534 \\
\rownum &CausalDiffAE  
& 0.297 & 0.132 & \textbf{0.031} & \textbf{0.034} 
& 0.045 & 0.434 & \textbf{0.035} & \textbf{0.064} 
& 0.136 & 0.322 & 0.492 & \textbf{0.082} 
& 0.146 & 0.303 & 0.064 & 0.471 \\
\midrule
\rownum &$\text{Ours}_{\text{w/~CM}}$        
& \textbf{0.014} & \textbf{0.035} & 0.043 & 0.259
& \textbf{0.045} & \textbf{0.041} & 0.058 & 0.120 
& \textbf{0.028} & \textbf{0.051} & \textbf{0.489} & 0.110 
& \textbf{0.030} & \textbf{0.033} & \textbf{0.041} & \textbf{0.336} \\
\midrule
\rownum &$\text{Ours}_{\text{w/o~CM}}$       
& 0.159 & 0.183 & $\sim$ & $\sim$
& 0.060 & 0.173 & $\sim$ & $\sim$
& 0.143 & 0.235 & $\sim$ & $\sim$ 
& 0.086 & 0.155 & $\sim$ & $\sim$ \\
\rownum &$\text{Ours}_{\text{w/~GT}}$       
& 0.013 & 0.033 & $\sim$ & $\sim$
& 0.043 & 0.039 & $\sim$ & $\sim$
& 0.025 & 0.036 & $\sim$ & $\sim$ 
& 0.028 & 0.035 & $\sim$ & $\sim$ \\
\bottomrule
\end{tabular}
}
\end{table*}
% ---- CelebA: compact single-column table (two-line headers) ----
\setcounter{rownumbers}{0}
\begin{table}[t]
\centering
\caption{CelebA results. We report target attribute effectiveness and include composition, realism, and minimality metrics.}
\label{tab:benchmark_celeba}
{\footnotesize
\setlength{\tabcolsep}{2.4pt}
\renewcommand{\arraystretch}{0.95}
\resizebox{\linewidth}{!}{%
\begin{tabular}{rl cccc c c c}
\toprule
&\multirow{2}{*}{Method}
& \textbf{(a) F1$\uparrow$}
& \textbf{(g) F1$\uparrow$}
& \textbf{(br) F1$\uparrow$}
& \textbf{(bl) F1$\uparrow$}
& \textbf{LPIPS$\downarrow$}
& \textbf{FID$\downarrow$}
& \textbf{CLD$\downarrow$}
\\
\cmidrule(lr){3-3}\cmidrule(lr){4-4}\cmidrule(lr){5-5}\cmidrule(lr){6-6}\cmidrule(lr){7-7}\cmidrule(lr){8-8}\cmidrule(lr){9-9}
&& $do(a)$  & $do(g)$ &$do(br)$  &$do(bl)$ & (Comp.) & (Real.) & (Min.) \\
\midrule
\rownum & VAE
& 35.0 & 90.9 & 29.6 & 41.2
& 0.282 & 59.397 & \textbf{0.299} \\
\rownum & HVAE
& \textbf{65.4} & 94.9 & 44.1 & \textbf{61.1}
& 0.122 & 35.712 & 0.305 \\
\rownum & GAN
& 41.3 & 98.2 & 23.3 & 49.2
& 0.276 & 27.861 & 0.304 \\
\midrule
\rownum & Ours
& 58.5 & \textbf{99.9} & \textbf{52.1} & 58.8
& \textbf{0.017} & 8.152 & 0.310 \\
\midrule
\multicolumn{9}{l}{\emph{with attention guidance for localized editing}} \\
\rownum & Ours (AG)
& 57.1 & 99.7 & 48.2 & 51.4
& -- & \textbf{5.213} & 0.301 \\
\bottomrule
\end{tabular}
} % resizebox
} % footnotesize
\end{table}

\paragraph{Synthetic Imaging Counterfactuals.} 
We first evaluate our approach on the Pendulum dataset, which consists of four continuous causal variables (pendulum angle, light position, shadow length, and shadow position). We compare against CausalVAE~\citep{yang2021causalvae}, as well as diffusion-based methods DisDiffAE~\citep{preechakul2022diffusion} and CausalDiffAE~\citep{komanduri2024causal}, using the causal graph shown in Figure~\ref{fig:pendulumn_quality1}. As reported in Table~\ref{tab:pend_benchamrk_effectiveness}, our method (row 4) achieves state-of-the-art intervention performance across most attributes. In particular, under $do(l)$ (light intervention), we obtain up to a 91\% reduction in MAE for light prediction (from 0.434 to 0.041), indicating highly accurate control over light movement. Moreover, when intervening on light, our approach correctly preserves the pendulum angle while inducing causal changes in the descendant shadow attributes (shadow length and position), consistent with the real-world physical law. We further conduct an investigation into the role of causal mechanisms. Without explicit causal mechanisms (row 5), the model exhibits larger intervention errors, underscoring the necessity of updating semantic variables with causal dependencies during generation. In contrast, injecting synthetic ground-truth labels (row 6) yields performance close to ours (row 4), indicating that our causal mechanism provides a principled approximation to ground-truth causal reasoning, generating faithful counterfactuals over all attributes as shown in Figure~\ref{fig:pendulumn_quality1}.

\begin{figure}[t]
        \centering
        \includegraphics[width=\linewidth]{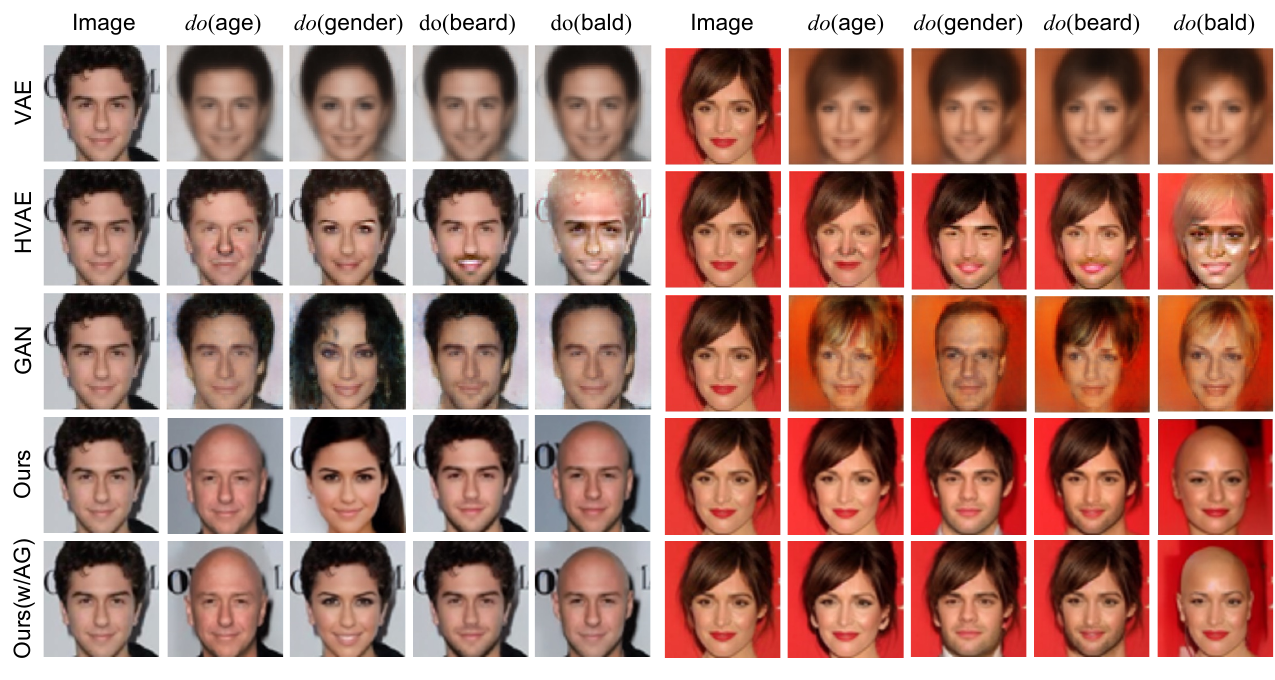}
        
        \caption{Qualitative comparison on the CelebA dataset. The results demonstrate that our method preserves core identity features while effectively enforcing causal interventions.} 
        \label{fig:benchmark_celeba_quality}
    \end{figure}

\paragraph{Human Face Counterfactuals.}
Following the benchmarking of~\citet{melistas2024benchmarking}, we evaluate Causal-Adapter on CelebA test set for human face counterfactual generation across four categorical attributes (age, gender, beard, bald) with the causal graph shown in Figure~\ref{fig:ctf_generate}. We also incorporate attention guidance to perform localized editing and assess the utility of our learned attention maps. Table~\ref{tab:benchmark_celeba} reports intervention effectiveness, composition, realism, and minimality. Our method achieves best performance across most interventions, including up to an $86\%$ reduction in LPIPS (composition, from 0.122 to 0.017) and an $79\%$ reduction in FID (realism, from 27.861 to 8.152). HVAE achieves competitive effectiveness via post-training classifier optimization but produces classifier-biased artifacts and reduced fidelity (Figure~\ref{fig:benchmark_celeba_quality}). Further localized editing with attention guidance balances intervention effectiveness and identity preservation, achieving the best FID (from 8.152 to 5.213). Qualitative results confirm that the learned attention maps enable precise, localized detailed editing (e.g., modifying gender without altering hairstyle), enable Causal-Adapter to preserve core identity while enforcing causal interventions. Full results are in Appendix~\ref{Appendix:extra_celeba}.

\setcounter{rownumbers}{0}
\begin{table}[t]
\centering
\caption{Intervention target effectiveness on ADNI.}
\label{tab:effectiveness_benchmark_adni}
{\footnotesize
\setlength{\tabcolsep}{2.4pt}
\renewcommand{\arraystretch}{0.95}
\resizebox{\linewidth}{!}{%
\begin{tabular}{rl ccc c c c}
\toprule
&\multirow{2}{*}{Method}
& \textbf{(b) MAE$\downarrow$}
& \textbf{(v) MAE$\downarrow$}
& \textbf{(s) F1$\uparrow$}
& \textbf{LPIPS$\downarrow$}
& \textbf{FID$\downarrow$}
& \textbf{CLD$\downarrow$}
\\
\cmidrule(lr){3-3}\cmidrule(lr){4-4}\cmidrule(lr){5-5}\cmidrule(lr){6-6}\cmidrule(lr){7-7}\cmidrule(lr){8-8}
&& $do(b)$ & $do(v)$ & $do(s)$ & (Comp.) & (Real.) & (Min.) \\
\midrule
\rownum & VAE
& 0.17 & 0.20 & 0.46
& 0.306 & 278.245 & 0.352 \\
\rownum & HVAE
& 0.09& 0.04 & 0.41
& 0.101 & 74.696 & 0.347 \\
\rownum & GAN
& 0.17 & 0.22 & 0.05
& 0.268 & 113.749 & 0.353 \\
\midrule
\rownum & Ours
& \textbf{0.09} & \textbf{0.03} & \textbf{0.48}
& \textbf{0.035} & 9.130 & 0.346 \\
\midrule
\multicolumn{8}{l}{\emph{with attention guidance for localized editing}} \\
\rownum & Ours (AG)
& 0.10 & 0.04 & 0.46
& -- & \textbf{9.066} & \textbf{0.332} \\
\bottomrule
\end{tabular}
} % resizebox
} % footnotesize
\end{table}

\begin{figure}[t]
    \centering
    \includegraphics[width=1.05\linewidth]{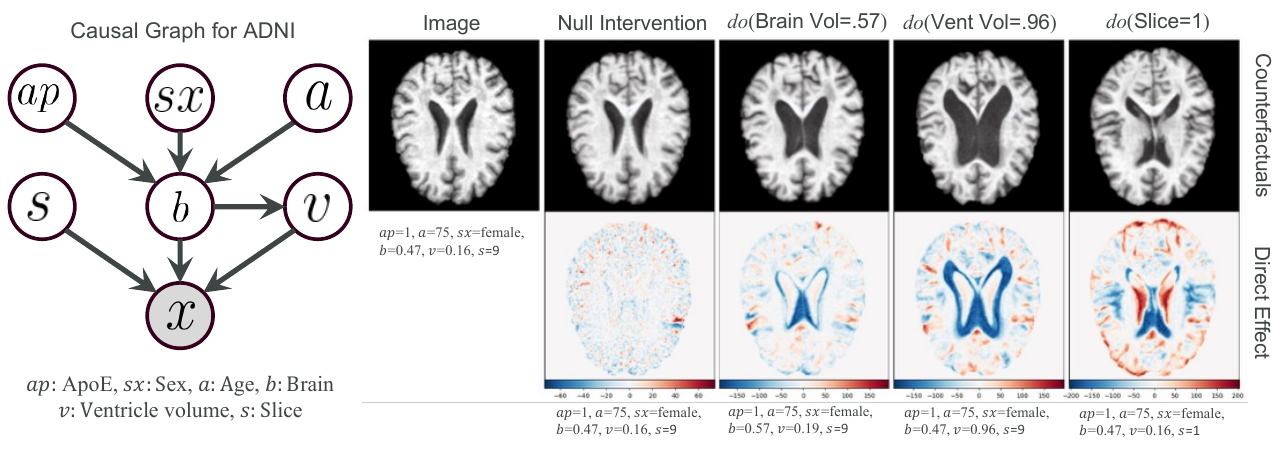}
    \caption{ADNI brain MRI counterfactual results from Causal-Adapter. Direct causal effects are shown (red: increase; blue: decrease). The results show sharp, localized interventional changes consistent with the causal graph (left), while preserving the observation’s identity.}
    \label{fig:adni_quality1}
\end{figure}

\setcounter{rownumbers}{0}
\begin{table}[t]
\centering
\caption{CelebA-HQ results. Intervention effectiveness, reversibility, and identity preservation for eyeglasses and smiling.}
\label{tab:celeba-hq}
{\footnotesize
\setlength{\tabcolsep}{2.4pt}
\renewcommand{\arraystretch}{0.95}
\resizebox{\linewidth}{!}{%
\begin{tabular}{rl c c c c c c}
\toprule
&\multirow{2}{*}{Method}
& \textbf{(g) F1$\uparrow$}
& \textbf{(g) $L_1$$\downarrow$}
& \textbf{(g) LPIPS$\downarrow$}
& \textbf{(s) F1$\uparrow$}
& \textbf{(s) $L_1$$\downarrow$}
& \textbf{(s) LPIPS$\downarrow$}
\\
\cmidrule(lr){3-3}\cmidrule(lr){4-4}\cmidrule(lr){5-5}
\cmidrule(lr){6-6}\cmidrule(lr){7-7}\cmidrule(lr){8-8}
&& $do(g)$ & (Rev.) & (IDP) & $do(s)$ & (Rev.) & (IDP) \\
\midrule
\rownum & VCI
& 3.39  & --    & --
& 33.81 & --    & -- \\
\rownum & HVAE
& 65.31 & --    & --
& 75.33 & --    & -- \\
\rownum & DiffCounter
& 96.86 & 0.185 & 0.096
& \textbf{94.93} & 0.183 & 0.066 \\
\rownum & Ours
& \textbf{99.26} & \textbf{0.049} & 0.084
& 94.15 & \textbf{0.028} & \textbf{0.028} \\
\midrule
\rownum & Ours (DiT)
& 97.39 & 0.086 & \textbf{0.060}
& 94.71 & 0.089 & 0.035 \\
\bottomrule
\end{tabular}
} % resizebox
} % footnotesize
\end{table}

\paragraph{Brain Imaging Counterfactuals.}
We further evaluate our method on ADNI dataset, which includes six attributes in both categorical variables (ApoE, Sex, Slice) and continuous variables (Age, Brain Volume, Ventricular Volume). %The corresponding causal graph is shown in Figure~\ref{fig:adni_quality1}.
Following~\citet{melistas2024benchmarking}, we intervene on three generative conditioning attributes (Brain Volume, Ventricular Volume, Slice) and report the results in Table~\ref{tab:effectiveness_benchmark_adni}. Our approach achieves best performance in intervention effectiveness (up to $50\%$ MAE reduction in fine-grained edits) and minimality, while also delivering strong realism ($87\%$ FID reduction) even without attention guidance. Qualitative results (Figure~\ref{fig:adni_quality1}) further show that our model produces sharp and localized interventional changes consistent with the causal graph. In particular, fine-grained edits
to Ventricular Volume visibly enlarge the ventricle region while preserving subject identity. Full results are in Appendix~\ref{Appendix:extra_adni}.

\begin{figure}[t]
    \centering
    \includegraphics[width=0.95\linewidth]{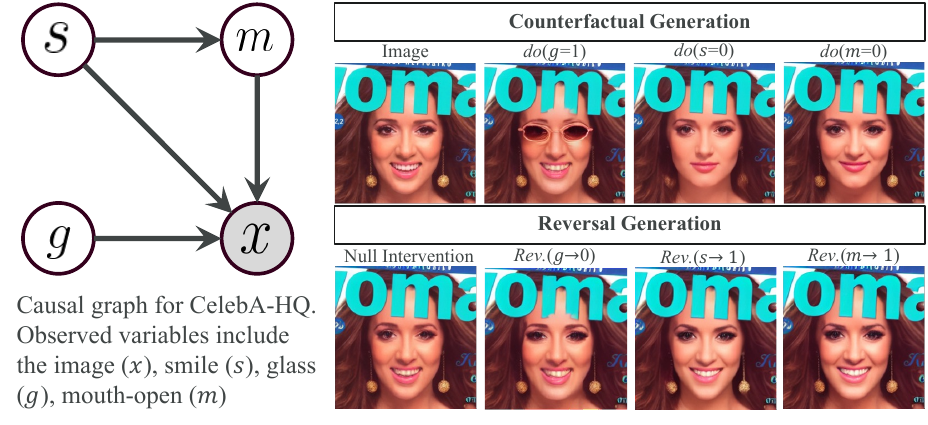}
    \caption{CelebA-HQ counterfactuals, together with reversible cases that map back to the original observations, illustrating identity preservation and reversibility.}
    \label{fig:main_celebqhq_qualiti}
\end{figure}

\paragraph{High-Resolution Face Counterfactuals.} We further compare Causal-Adapter with recent SOTA methods, VCI~\citep{wu2024counterfactual}, HVAE~\citep{pmlr-v202-de-sousa-ribeiro23a}, and DiffCounter~\citep{rasal2025diffusion} on high-resolution human face dataset CelebA-HQ. We follow the same settings in DiffCounter for fair comparison, focusing on three categorical variables (eyeglass, smiling, mouth-open). Quantitative results are reported in Table~\ref{tab:celeba-hq}, with qualitative examples shown in Figure~\ref{fig:main_celebqhq_qualiti}. Our method (row 4) achieves intervention effectiveness comparable to DiffCounter, while substantially improving both reversibility and identity preservation. We further evaluate generalization by instantiating Causal-Adapter with a diffusion transformer (DiT) backbone~\citep{esser2024scaling}. 
As shown in Table~\ref{tab:celeba-hq} (row~5) and Appendix~\ref{Appendix:extra_celebahq}, Causal-Adapter transfers effectively across distinct T2I backbones and produces causally faithful, high-resolution counterfactuals.

\begin{figure}[t]
\centering
\includegraphics[width=0.94\linewidth]{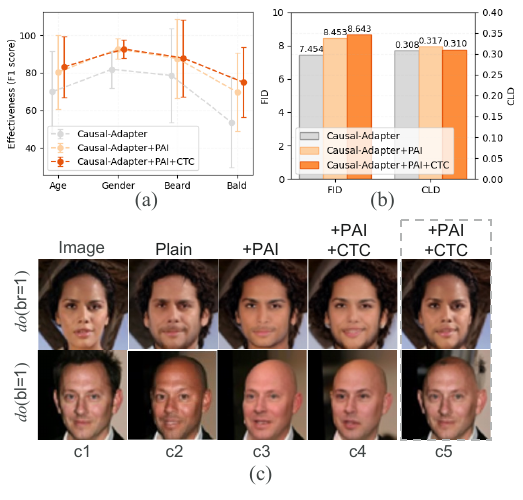}
    \caption{Ablation study on CelebA validation set. (a) Average intervention effectiveness. (b) Realism and minimality. (c) Qualitative examples, with dotted boxes indicating results of localized editing. Both quantitative and qualitative results demonstrate the necessity of the proposed PAI and CTC.}
    \label{fig:ablation_study_main}
\end{figure}
\paragraph{Ablation Study.}
We conduct an ablation study on the CelebA validation set to evaluate the contribution of each regularizer. As shown in Figures~\ref{fig:ablation_study_main}(a) and (b), the plain adapter achieves the lowest intervention effectiveness. Adding the PAI module yields a consistent average gain of $+11.6\%$ F1 across all attributes with slight increases in FID and CLD, indicating that aligning causal semantics with spatial features in the diffusion latents enables more effective edits. Incorporating CTC further improves intervention effectiveness by enforcing token embedding disentanglement, while also reducing CLD (from 0.317 to 0.310) and keeping FID stable (from 8.453 to 8.643), resulting more faithful counterfactual generation. Figure~\ref{fig:ablation_study_main}c presents qualitative ablation results. Spurious correlations arise in plain adapter,~\textit{e.g.}, beard interventions in females induce male facial features (c2). PAI alleviates attribute entanglement (c3), while the complete regularization with CTC produces bearded female, demonstrating mitigation of such correlations (c4). Under bald interventions, the baselines alter skin color or age (c2–c3), whereas the full regularization edits baldness with only minor facial changes (c4). Finally, attention guidance enhances identity preservation with localizing edits (c5). Extended results are presented in Appendix~\ref{Appendix:extra_ablation}. 

\paragraph{Beyond Pre-defined Graphs.}
Existing counterfactual generation methods often assume a pre-defined causal graph~\citep{pearl2009causality,pawlowski2020deep,yang2021causalvae}, which can be restrictive when the graph is misspecified, partially known, or unavailable. 
To relax this assumption, we make the adjacency matrix $A$ learnable and regularize it with an acyclicity constraint following differentiable DAG methods such as NOTEARS~\citep{zheng2018dags} and DAGMA~\citep{bello2022dagma}. 
This extension requires no architectural changes and only adds a structural loss term.

\begin{figure}[t]
    \centering
    \includegraphics[width=\linewidth]{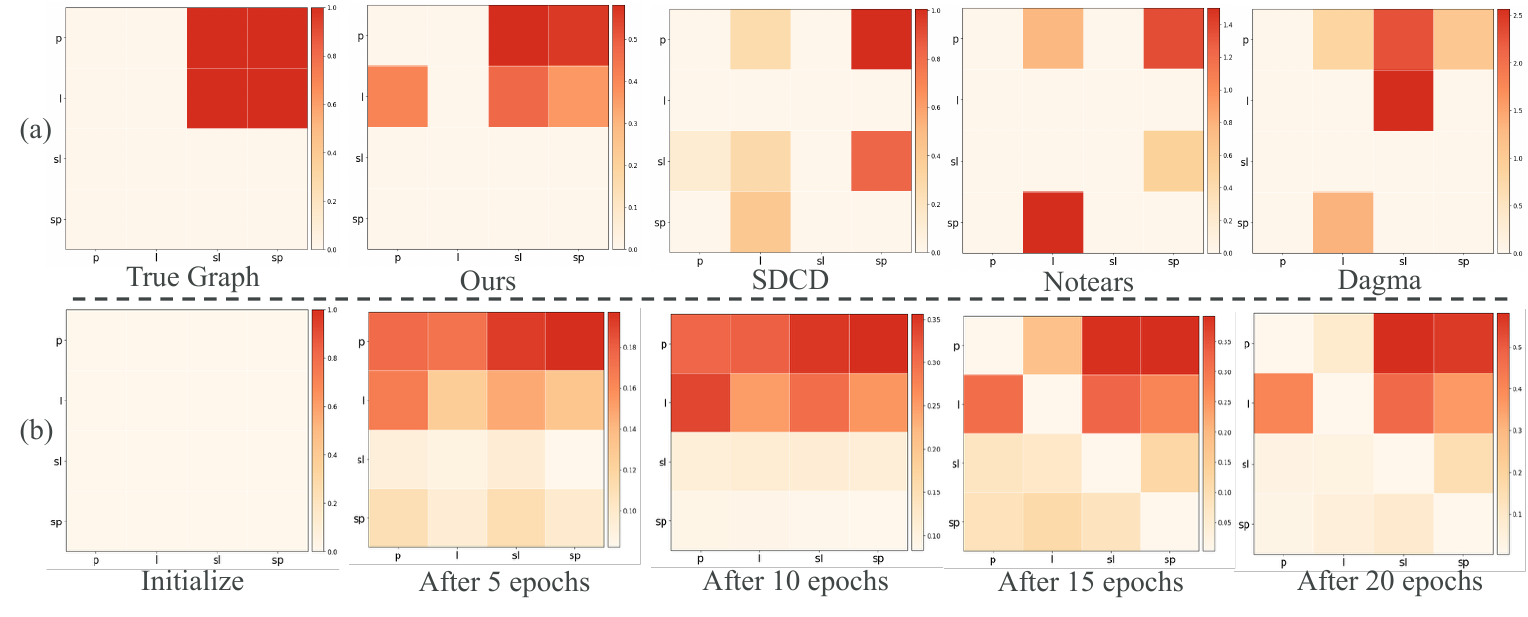}
    \caption{Causal discovery performance on the Pendulum dataset. 
    (a) Comparison of Causal-Adapter with SDCD, NOTEARS, and DAGMA. 
    (b) Learning trajectory of the adjacency matrix $A$.}
    \label{fig:main-causal-discovery}
\end{figure}

As shown in Figure~\ref{fig:main-causal-discovery}(a), this extension enables Causal-Adapter to perform causal structure learning and achieve superior performance compared with SOTA differentiable causal discovery methods, including SDCD~\citep{nazaret2024stable}.  
Moreover, our method recovers more true causal edges than competing baselines.  
The learning dynamics of $A$ during training are visualized in Figure~\ref{fig:main-causal-discovery}(b), showing stable convergence toward the ground-truth graph and demonstrating the interpretable behavior of our model.  
Overall, these results suggest that Causal-Adapter has the potential to unify causal discovery and counterfactual generation within a single, simple, and efficient framework.  
Extended results are presented in Appendix~\ref{subsec:mitigate_causal_graph}.

\paragraph{Ablation on Causal Graphs.}
\setcounter{rownumbers}{0}
\begin{table}[t]
\centering
\caption{
Ablation study on different causal graphs for SCM modeling. 
w/o~SCM denotes the setting without using a causal graph, while w/~SCM1 and w/~SCM2 correspond to the causal graphs shown in Figure~\ref{fig:main-ablation-pend}.
}
\label{tab:main_pend_graph_ablation}
{\footnotesize
\setlength{\tabcolsep}{2.4pt}
\renewcommand{\arraystretch}{0.95}
\resizebox{\linewidth}{!}{%
\begin{tabular}{rlcccccccc}
\toprule
&\multirow{2}{*}{Method}
& \multicolumn{2}{c}{\textbf{$(p)$ MAE $\downarrow$}}
& \multicolumn{2}{c}{\textbf{$(l)$ MAE $\downarrow$}}
& \multicolumn{2}{c}{\textbf{$(sl)$ MAE $\downarrow$}}
& \multicolumn{2}{c}{\textbf{$(sp)$ MAE $\downarrow$}} \\
\cmidrule(lr){3-4}\cmidrule(lr){5-6}\cmidrule(lr){7-8}\cmidrule(lr){9-10}
&& $do(p)$ & $do(l)$
& $do(p)$ & $do(l)$
& $do(p)$ & $do(l)$
& $do(p)$ & $do(l)$ \\
\midrule
\rownum &$\text{w/o~SCM}$
& 0.159 & 0.183
& 0.060 & 0.173
& 0.143 & 0.235
& 0.086 & 0.155 \\
\rownum &$\text{w/~SCM1}$
& \textbf{0.011} & 0.040
& 0.048 & 0.062
& 0.039 & 0.104
& 0.031 & 0.033 \\
\rownum &$\text{w/~SCM2}$
& 0.014 & \textbf{0.035}
& \textbf{0.045} & \textbf{0.041}
& \textbf{0.028} & \textbf{0.051}
& \textbf{0.030} & \textbf{0.033} \\
\bottomrule
\end{tabular}
}
}
\end{table}

\begin{figure}[t]
    \centering
    \includegraphics[width=\linewidth]{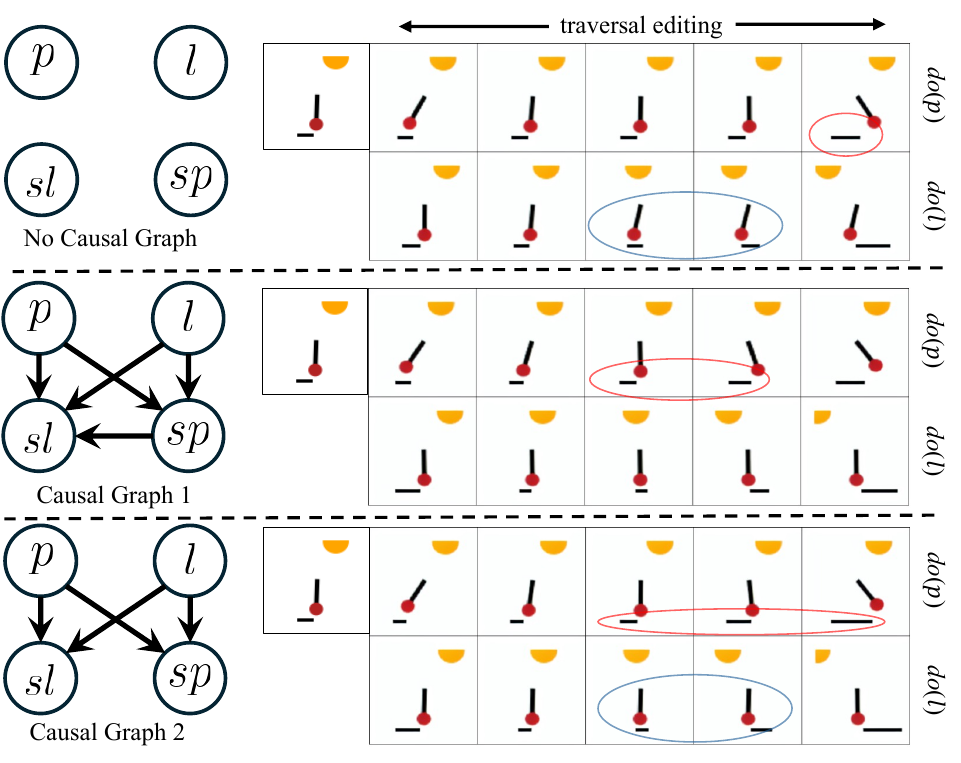}
    \caption{
    Pendulum counterfactuals under three graph settings: no causal graph, an incorrect causal graph (SCM1), and the correct causal graph (SCM2).
    }
    \label{fig:main-ablation-pend}
\end{figure}

To validate the robustness of Causal-Adapter to causal graph specification, we conduct an ablation study under different graph settings. 
As shown in Table~\ref{tab:main_pend_graph_ablation} and the results in Figure~\ref{fig:main-ablation-pend}, we compare three settings: 
(1) without SCM, 
(2) with an incorrect causal graph, denoted as SCM1, and 
(3) with the correct causal graph, denoted as SCM2. 
Without SCM, interventions can still affect shadow-related variables because pendulum, light, and shadow attributes are strongly correlated in the training data. 
However, these changes are primarily correlation-driven rather than graph-faithful: intervening on one factor may spuriously modify another non-intervened factor, such as changing the light while also moving the pendulum. 
With SCM1, the incorrectly specified causal edge leads to an inconsistent propagation pattern and results in a different shadow evolution behavior. 
In contrast, with the correct graph SCM2, the intervention effects propagate more consistently with the underlying physical mechanisms.

Importantly, our design separates SCM modeling from conditional disentanglement. 
This allows the framework to adapt to different causal graphs at inference time by replacing the SCM, without retraining the diffusion backbone. 
These results show that, even without a causal graph, the model can still function as a disentangled editing method; however, with the correct causal graph, the generated counterfactuals better match the underlying data-generating structure and achieve stronger quantitative performance. Extended results are presented in Appendix~\ref{subsec:mitigate_causal_graph}.

\paragraph{Generalization to Broader Settings.}
To further assess the practical effectiveness of Causal-Adapter, we compare it with recent commercial foundation editing models, including Nano Banana 2, Qwen-Image-Edit, and Flux.1-Kontext-Dev, on our three benchmark datasets and a less tightly aligned full-body human image dataset. Results in Appendix~\ref{Appendix:compare_commercial_model} and Appendix~\ref{Appendix:human_body_generalization} support its practical applicability across broader image domains.

\section{Conclusion}
We introduced Causal-Adapter, a modular framework to tame Text-to-Image diffusion models for counterfactual generation. Our motivational study revealed that current Text-to-Image diffusion based editing approaches lack an explicit structural causal model for attribute control, making it difficult to generate faithful counterfactual images. In contrast, Causal-Adapter is a simple yet effective framework that leverages a frozen diffusion backbone and injects causal semantic attributes through a pluggable adapter network to explicitly learn causal semantics. We further proposed prompt aligned injection and conditioned token contrastive optimization, which align attribute semantics with spatial features and promote disentanglement in the latent space, reducing spurious correlations while preserving identity for generations. Causal-adapter achieves superior counterfactual generation performance on multiple datasets. Extensive evaluation across diverse settings further confirm that Causal-Adapter provides a robust, scalable, and practical alternative for enabling causal editing in modern T2I systems.

\clearpage
\section*{Impact Statement}
This paper presents work whose goal is to advance the field of Machine Learning. There are many potential societal consequences of our work, none which we feel must be specifically highlighted here.

\section*{Acknowledgment}
We thank the area chair and reviewers for their constructive feedbacks. Z.Liu and S.Tsaftaris were supported by the UKRI AI programme and the Engineering and Physical Sciences Research Council, for CHAI - Causality in Healthcare AI Hub [grant number EP/Y028856/1].

% This work utilizes two real-world datasets, CelebA~\citep{liu2015faceattributes} and ADNI~\citep{petersen2010alzheimer}, in conjunction with pretrained text-to-image generation models. We emphasize that our research is conducted solely for scientific purposes, and we explicitly oppose any misuse of generative AI that harms individuals, violates privacy, or spreads misinformation. While our approach demonstrates capabilities in generating human faces and MRI images, we acknowledge the potential for dual-use risks. To mitigate these concerns, we adhere to strict ethical standards, including compliance with applicable legal and institutional frameworks, preservation of data privacy, and a commitment to promoting socially beneficial applications of generative models.
\bibliography{example_paper}
\bibliographystyle{icml2026}

%%%%%%%%%%%%%%%%%%%%%%%%%%%%%%%%%%%%%%%%%%%%%%%%%%%%%%%%%%%%%%%%%%%%%%%%%%%%%%%
%%%%%%%%%%%%%%%%%%%%%%%%%%%%%%%%%%%%%%%%%%%%%%%%%%%%%%%%%%%%%%%%%%%%%%%%%%%%%%%
% APPENDIX
%%%%%%%%%%%%%%%%%%%%%%%%%%%%%%%%%%%%%%%%%%%%%%%%%%%%%%%%%%%%%%%%%%%%%%%%%%%%%%%
%%%%%%%%%%%%%%%%%%%%%%%%%%%%%%%%%%%%%%%%%%%%%%%%%%%%%%%%%%%%%%%%%%%%%%%%%%%%%%%
\newpage
\appendix
\onecolumn

% \etocdepthtag.toc{mtappendix}
% \etocsettagdepth{mtappendix}{subsection}
% \etocsettagdepth{mtchapter}{none}

% {
%     \hypersetup{linkcolor=black}
%     \tableofcontents
% }
%\definecolor{tocblue}{HTML}{04146E}
\definecolor{tocblue}{HTML}{111671}
\section*{Appendix Table of Contents} % Use \section* to format it as a title without numbering
\addcontentsline{toc}{section}{Appendix Table of Contents} % Ensures it appears in TOC
\label{Appendix:table_of_contents} % Correct label placement

\vskip 4mm
\hrule height .5pt
\vskip 4mm

\begin{itemize}[label={},leftmargin=*]
    \item \textbf{\textcolor{tocblue}{\hyperref[Appendix:Related Works]{Appendix A - Extended Related Works}}} \dotfill \pageref{Appendix:Related Works}
    % No figures in Appendix A.

    \item \textbf{\textcolor{tocblue}{\hyperref[Appendix:Full Motivational Study Results]{Appendix B - Full Motivational Study Results}}} \dotfill \pageref{Appendix:Full Motivational Study Results}
    % No figures in Appendix B.

    \item \textbf{\textcolor{tocblue}{\hyperref[Appendix:data_implement]{Appendix C - Dataset and Implementation}}} \dotfill \pageref{Appendix:data_implement}
    % No figures in Appendix C.
    \begin{itemize}[label={},leftmargin=*]
        \item \textit{\textcolor{tocblue}{\hyperref[Appendix:sub_implement_details]{\ref{Appendix:sub_implement_details}: Implementation Details}}}
          \dotfill \pageref{Appendix:sub_implement_details}
          
          \item \textit{\textcolor{tocblue}{\hyperref[Appendix:sub_implement_metrics]{\ref{Appendix:sub_implement_metrics}: Metrics}}}
          \dotfill \pageref{Appendix:sub_implement_metrics}
          
          \item \textit{\textcolor{tocblue}{\hyperref[Appendix:sub_full_regularization_Algorithm]{\ref{Appendix:sub_full_regularization_Algorithm}: Full Regularization Algorithm}}}
          \dotfill \pageref{Appendix:sub_full_regularization_Algorithm}
    \end{itemize}

    \item \textbf{\textcolor{tocblue}{\hyperref[Appendix:extra_ablation]{Appendix D - Extended Ablation Results}}} \dotfill \pageref{Appendix:extra_ablation}
    \begin{itemize}[label={},leftmargin=*]
        \item \textit{\textcolor{tocblue}{\hyperref[extra_ablation:effect_guidance_scale]{\ref{extra_ablation:effect_guidance_scale}: Effect of Guidance Scale}}}
          \dotfill \pageref{extra_ablation:effect_guidance_scale}
          
          \item \textit{\textcolor{tocblue}{\hyperref[extra_ablation:effect_ddim_steps]{\ref{extra_ablation:effect_ddim_steps}: Effect of DDIM Steps}}}
          \dotfill \pageref{extra_ablation:effect_ddim_steps}
          
          \item \textit{\textcolor{tocblue}{\hyperref[extra_ablation:investigation_attentions]{\ref{extra_ablation:investigation_attentions}: Investigation of Attention Guidance}}}
          \dotfill \pageref{extra_ablation:investigation_attentions}

          \item \textit{\textcolor{tocblue}{\hyperref[extra_ablation:qualitative_evidence]{\ref{extra_ablation:qualitative_evidence}: Qualitative Evidence}}}
          \dotfill \pageref{extra_ablation:qualitative_evidence}

    \end{itemize}

    \item \textbf{\textcolor{tocblue}{\hyperref[Appendix:addition_results]{Appendix E - Additional Results}}} \dotfill \pageref{Appendix:addition_results}
    
    \begin{itemize}[label={},leftmargin=*]
        \item \textit{\textcolor{tocblue}{\hyperref[Appendix:extra_pend]{\ref{Appendix:extra_pend}: Pendulum}}}
          \dotfill \pageref{Appendix:extra_pend}
          
          \item \textit{\textcolor{tocblue}{\hyperref[Appendix:extra_celeba]{\ref{Appendix:extra_celeba}: CelebA}}}
          \dotfill \pageref{Appendix:extra_celeba}
          
          \item \textit{\textcolor{tocblue}{\hyperref[Appendix:extra_adni]{\ref{Appendix:extra_adni}: ADNI}}}
          \dotfill \pageref{Appendix:extra_adni}

          \item \textit{\textcolor{tocblue}{\hyperref[Appendix:extra_celebahq]{\ref{Appendix:extra_celebahq}: CelebA-HQ}}}
          \dotfill \pageref{Appendix:extra_celebahq}
          
          \item \textit{\textcolor{tocblue}{\hyperref[Appendix:additional_attention_maps]{\ref{Appendix:additional_attention_maps}: Attention Maps}}}
          \dotfill \pageref{Appendix:additional_attention_maps}

          \item \textit{\textcolor{tocblue}{\hyperref[Appendix:compare_commercial_model]{\ref{Appendix:compare_commercial_model}: Comparison with Commercial Foundation Models}}}
          \dotfill \pageref{Appendix:compare_commercial_model}

          \item \textit{\textcolor{tocblue}{\hyperref[Appendix:human_body_generalization]{\ref{Appendix:human_body_generalization}: Generalization Beyond Aligned Images}}}
          \dotfill \pageref{Appendix:human_body_generalization}
    \end{itemize}

    \item \textbf{\textcolor{tocblue}{\hyperref[Appendix:Counterfactual Identifiability]{Appendix F - Counterfactual Identifiability}}} \dotfill \pageref{Appendix:Counterfactual Identifiability}

    \item \textbf{\textcolor{tocblue}{\hyperref[subsec:mitigate_causal_graph]{Appendix G - Relaxing the Assumption of a Known Causal Graph}}} \dotfill \pageref{subsec:mitigate_causal_graph}

\end{itemize}

\vskip 4mm
\hrule height .5pt
\vskip 10mm

\clearpage
\section{Extended Related Works \footnotesize{\textbf{\hyperref[Appendix:table_of_contents]{[Back to Contents]}}}}
\label{Appendix:Related Works}
\paragraph{Counterfactual Image Generation}aims to synthesize images that reflect the visual effect of a hypothetical intervention while preserving non-intervened attributes according to an underlying causal graph and maintaining instance-specific identity details~\citep{komanduri2023identifiable,melistas2024benchmarking}.
Existing approaches often augment generative models with explicit structural causal models (SCMs) and follow Pearl’s \textit{abduction–action–prediction} paradigm~\citep{pearl2009causality,pearl2013structural,pawlowski2020deep,shen2022weakly,sanchez2022diffusion,pmlr-v202-de-sousa-ribeiro23a,wu2024counterfactual,spyrou2025causally}. Early work predominantly relied on \textbf{VAEs or GANs}~\citep{kingma2013auto,goodfellow2014generative,kocaoglu2017causalgan,pawlowski2020deep}, as their noise-injection and variational sampling mechanisms offered a natural way to represent exogenous uncertainty, while their objectives encouraged disentanglement of causal factors, inspired by $\beta$-VAE~\citep{higgins2017beta}. \citet{pmlr-v202-de-sousa-ribeiro23a} extends this line by incorporating hierarchical VAEs~\citep{vahdat2020nvae} to estimate direct, indirect, and total causal effects, whereas \citet{wu2024counterfactual} combined variational Bayesian inference with adversarial training to improve abduction and preserve identity. Other works~\citep{yang2021causalvae,shen2022weakly,komanduri2023learning,li2024causal} integrates SCM priors directly into the latent space of VAEs, enabling implicit causal reasoning on image encodings.
However, variational optimization inevitably introduces uncertainty into the learned representations, which can lead to posterior collapse or the loss of semantically meaningful factors. This results in an inherent trade-off between high-fidelity image synthesis and flexible attribute control~\citep{higgins2017beta,alemi2018fixing}. \citet{sanchez2022diffusion} introduces \textbf{DiffSCM}, the first framework to integrate diffusion with SCMs for counterfactual generation. DiffSCM employed DDIM inversion~\citep{song2020denoising} for abduction and conditioned the generative process on causal labels, but it was limited to small parent sets and simple causal structures. 
\citet{pan2024counterfactual} extends DiffSCM by combining a VAE for preliminary counterfactual generation with diffusion-based refinement, shifting the conditioning from attribute labels to pre-generated images to produce refined counterfactuals. \citet{chao2024modeling} models structural equations directly using diffusion processes, enabling counterfactual sampling on a predefined causal graph. While DCM also answers interventional queries, it trains a dedicated diffusion model per causal node. Recent works~\citep{komanduri2024causal,rasal2025diffusion,xia2025decoupled} build on \textbf{Diffusion Autoencoder} framework~\citep{preechakul2022diffusion}, equipping diffusion models with variational encoders to inject semantic attributes into diffusion latents. However, these methods require heavy post-training or fine-tuning and remain prone to spurious correlations. A key limitation is that disentanglement is typically enforced through auxiliary encoders with limited influence on the intrinsic latent space of diffusion models, failing to align semantic attributes with disentangled spatial representations of images and leaving causal disentanglement incomplete~\citep{wu2024counterfactual,yang2024diffusion}.

To address these limitations, we propose \textbf{Causal-Adapter}, an adaptive and modular framework that employs an adapter encoder to explicitly learn causal interactions between semantic attributes. We further introduce two regularization strategies: Prompt Aligned Injection (PAI) and Conditioned Token Contrastive Loss (CTC). These strategies align semantic attributes with spatial features in the diffusion latents and separate token embeddings across conditions, thereby enhancing causal representation learning and reducing spurious correlations, all while preserving the pre-trained diffusion weights.

\paragraph{Text-to-Image based Editing} aims to manipulate existing images according to user-provided natural language instructions~\citep{brooks2023instructpix2pix}. Most approaches rely on an inversion process~\citep{song2020denoising}, where the image is projected into a latent state and then synthesized under modified conditions~\citep{hertz2023prompttoprompt,ho2022classifier}. However, DDIM inversion and classifier-free guidance often often interfere with each other, leading to a trade-off between preserving essential content and achieving faithful edits~\citep{ju2023direct,huberman2024edit,kynkaanniemi2024applying,cao2025causalctrl}. To mitigate this, Null-text inversion~\citep{mokady2023null} learns a null embedding to account for reconstruction discrepancies, while ~\citet{dong2023prompt} optimizes conditional embeddings to reduce information loss in unconditional guidance. Yet, both methods require costly per-sample optimization. Subsequent works~\citep{ju2023direct,xu2024inversion,miyake2025negative} improve efficiency by recording residual losses between conditional and unconditional embeddings and re-injecting them during editing, to stabilize edits and preserving content. Textual inversion~\citep{gal2023an} improves content preservation by disentangling single concepts: it learns new text embeddings from a few personalized images to represent objects in novel contexts. Extending this, \citet{vinker2023concept} decomposes concept embeddings into sub-concepts via learned vectors injected into the latent space of T2I model, while \citet{jin2024image} performs multi-concept disentanglement using adjective-based prompts and contrastive optimization. \citet{lyu2024one} propose a one-dimensional non-invasive adapter that modulates concept semi-permeability in frozen diffusion models for concept erosion, and \citet{yeganeh2025latent} address domain shift in medical imaging, supporting counterfactual-like edits such as aging or disease progression without explicit structural constraints. Despite these advances, generic T2I editing remains insufficient for causal counterfactual generation. Existing methods depend heavily on prompt manipulation and do not incorporate a learnable structural causal model (SCM). They make no use of observed semantic attributes or a predefined causal graph, and therefore cannot enforce that edits follow correct causal dependencies. As a result, current T2I-based editing techniques struggle to simultaneously maintain causal faithfulness and identity preservation in counterfactual image generation.

As shown in our motivational study (Appendix~\ref{Appendix:Full Motivational Study Results}), we  evaluate the adaptation of text-to-image diffusion models for counterfactual image generation. Our findings show that relying solely on a frozen diffusion backbone with prompt-tuning is insufficient, as it fails to jointly represent causal semantic attributes and image embeddings. As a result, the model struggles to achieve precise counterfactual reasoning and generation. This highlights the need for an adaptive mechanism within the diffusion model, enhanced with injected causal semantics. Following the standard formulation of counterfactual image generation~\citep{pearl2009causality,pmlr-v202-de-sousa-ribeiro23a,wu2024counterfactual,rasal2025diffusion}, where a predefined causal graph is treated as the structural prior, our Causal-Adapter models the SCM directly on the observed semantic attributes. This enforces correct causal dependencies during intervention and enables the generation of counterfactual images that are both visually plausible and causally faithful. When the SCM module is omitted, the framework naturally reduces to a standard conditional generation setting for that attribute, as illustrated in Figure~\ref{fig:ctf_generate}.

\paragraph{Controllable Diffusion Models}extend T2I diffusion frameworks by incorporating additional user-specified signals to guide generation~\citep{huang2025diffusion}. One approach is to train diffusion models from scratch with multi-conditional objectives~\citep{huang2023composer,puglisi2024enhancing}, which achieves strong controllability and high-quality synthesis but at huge computational cost. More recently, adapter-based methods~\citep{zhang2023adding,li2023gligen,zhao2023uni,mou2024t2i,li2024dispose} have emerged as a scalable alternative. By attaching lightweight, trainable modules to a frozen Stable Diffusion backbone, these methods enable the model to incorporate auxiliary control signals such as segmentation masks or pose skeletons, significantly reducing training overhead while maintaining stability and fidelity. We adopt the same high-level recipe by treating causal semantic attributes as the auxiliary control signals, and employ an adapter encoder to
explicitly learn causal interactions between high-level variables. These interactions are then injected
into a a frozen diffusion backbone, and jointly optimized with partial textual embeddings. The design of Causal-Adapter introduces a dynamic and learnable prior, enabling the
effective adaptation of a frozen diffusion network for realistic and faithful counterfactual image
generation.

\newpage
\section{Full Motivational Study Results \footnotesize{\textbf{\hyperref[Appendix:table_of_contents]{[Back to Contents]}}}}
\label{Appendix:Full Motivational Study Results}
To assess the feasibility of using a pretrained text-to-image (T2I) model for counterfactual generation, we conduct a motivation study to answer the following three questions.

\begin{figure*}[htp]
    \centering
    \includegraphics[width=\linewidth]{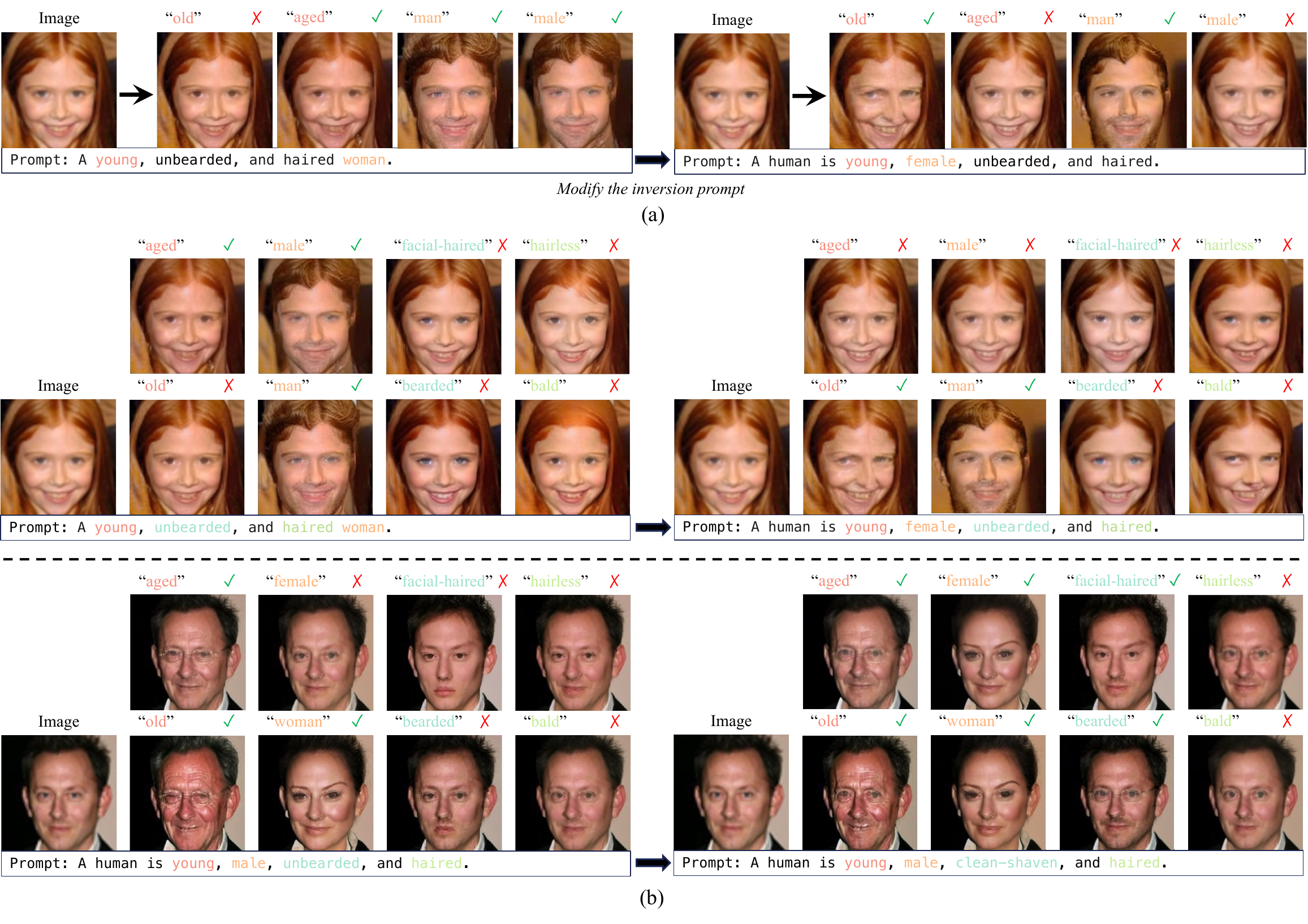}
    \caption{Null-Textual Inversion (NTI) relies heavily on prompt engineering, where minor word changes can determine editing success. 
    (a) Illustration of our study: two inversion prompts are given for the same image, and attribute words are replaced with different synonyms. Successful edits are marked with \textcolor{green}{\cmark} and failures with \textcolor{red}{\xmark} (human evaluation). Results show that editing outcomes are highly sensitive to prompt wording and reveal weak counterfactual faithfulness. 
    (b) Full results of this investigation.}
    \label{fig:motivation_NTI_1}
\end{figure*}

\textbf{1. Can existing T2I based editing methods perform counterfactual generation?}  
We begin by examining a representative text-to-image (T2I) editing method, Null-Textual Inversion (NTI)~\citep{mokady2023null}. Our findings highlight two fundamental issues:  (1) \textit{Heavy reliance on prompt engineering and instability.}  
As shown in Figure~\ref{fig:motivation_NTI_1}a, the success of NTI-based edits is highly sensitive to prompt wording. For example, when editing age, the token “old” may fail while the synonym “aged” succeeds under one inversion prompt, yet the opposite occurs under another prompt. This indicates that editing success and visual quality depend not only on the choice of attribute word but also on the particular inversion prompt. Even minor wording changes that preserve semantic meaning can alter tokenization and cross-attention patterns, leading to inconsistent success rates and perceptual artifacts. Moreover, to preserve identity while achieving the desired edit, practitioners must manually try multiple prompt variants, requiring extensive hand-crafted effort. (2) \textit{Weak counterfactual faithfulness.}  Beyond instability, prompt-based editing fails to reliably reflect the intended intervention. In Figure~\ref{fig:motivation_NTI_1}b, using the word ``man'' to edit gender (woman case) or ``old'' to edit age (man case) yields different counterfactual characteristics depending on the inversion prompt: one ``old'' edit introduces glasses while another does not. Such variability arises purely from prompt formulation, introducing extraneous variance unrelated to the target attribute. This illustrates weak counterfactual faithfulness: the same intervention should yield coherent edits of the intended factor, without unintended changes in other attributes. Hence, purely text-driven control is inadequate for reliable counterfactual generation.  

\begin{figure*}[htp]
    \centering
    \includegraphics[width=\linewidth]{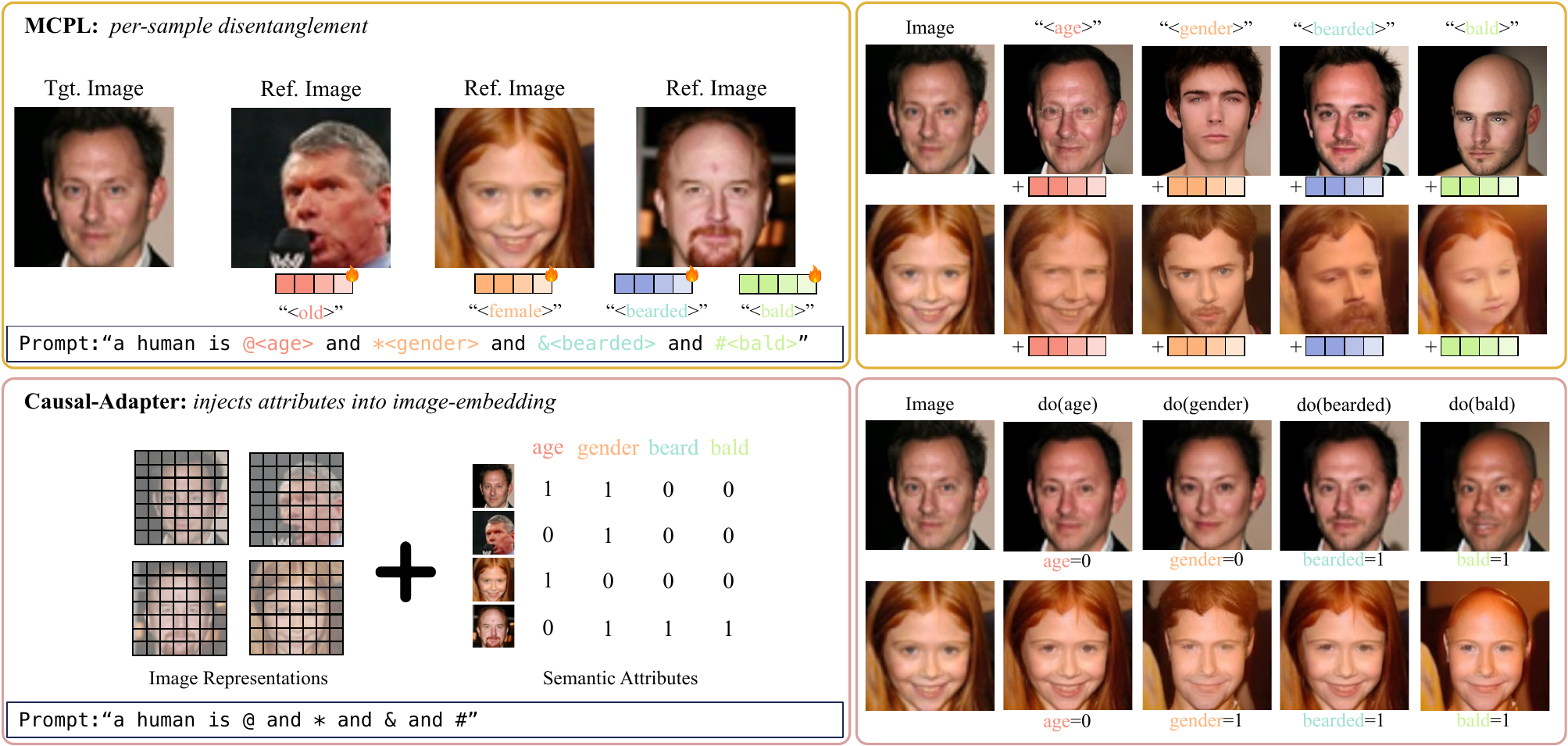}
    \caption{Multi-Concept Prompt Learning (MCPL) as a representative prompt-learning baseline. Each placeholder token (\textit{e.g.}, $@$ for ``young'', $*$ for ``male'') is initialized with a pretrained embedding and jointly optimized across multiple concepts. At test time, counterfactuals are generated by swapping embeddings with target embeddings (\textit{e.g.}, ``old'', ``female''). MCPL can achieve some edits missed by NTI (e.g., baldness, beard), but often entangles unrelated attributes, alters backgrounds, and requires per-instance optimization. In contrast, our vanilla Causal-Adapter injects causal attributes into image embeddings, supporting batch optimization and counterfactual generation via direct attribute interventions. }
    \label{fig:motivation_mcpl_2}
\end{figure*}

\begin{figure*}[h]
    \centering
    \includegraphics[width=\linewidth]{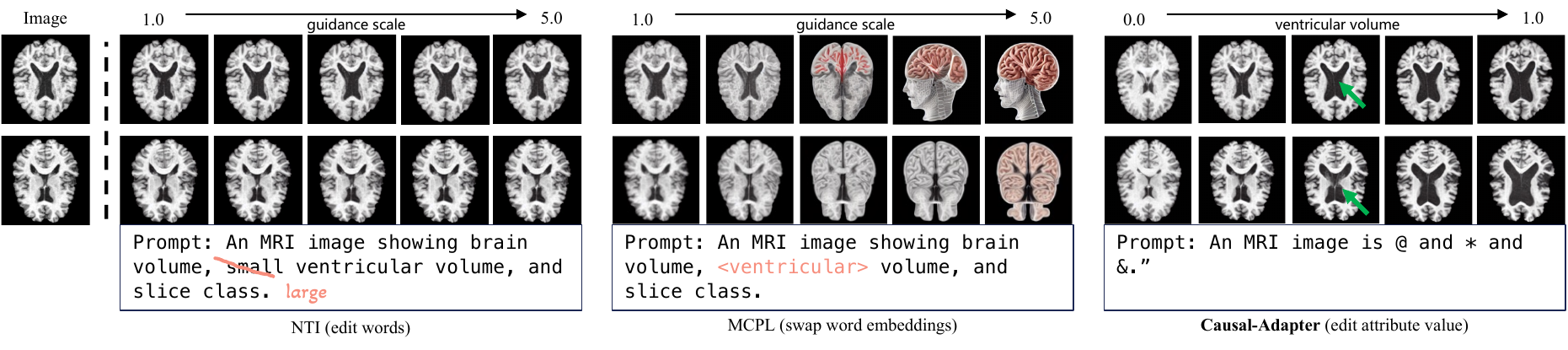}
    \caption{Fine-grained anatomical counterfactual editing of brain ventricular volume. NTI and MCPL cannot achieve fine-grained control with text-only prompts. In contrast, our Causal-Adapter produces sharp, localized interventions that smoothly adjust ventricular volume from small to large while preserving subject identity.}
    \label{fig:motivation_ADNI_3}
\end{figure*}

\textbf{2. Can multi-concept prompt learning yield disentangled attribute control?} Prompt-learning approaches such as Textual Inversion~\citep{gal2023an}, Inspiration Tree~\citep{vinker2023concept} and Multi-Concept Prompt Learning (MCPL) framework~\citep{jin2024image} to learn conditional concept embeddings for attribute disentanglement. We adopt MCPL framework as our baseline. As shown in Figure~\ref{fig:motivation_mcpl_2}, MCPL can perform certain edits that NTI fails to capture (e.g., editing baldness for men or adding beards for women) by equipping text embeddings learned from multiple samples. However, we observe two key limitations: (1) \textit{Entangled edits and lack of faithfulness.}  
Edits often induce changes in unrelated attributes, heavily altering the background or unintended regions. This leads to a loss of fidelity and identity preservation, thus violating the faithfulness requirement of counterfactual generation. (2)  \textit{Per-sample optimization is still required.}  
Similar to NTI, prompt-learning approaches need separate fine-tuning for each image instance. This makes them impractical for scalable, real-world causal reasoning tasks. Our findings show that relying solely on a frozen diffusion backbone with prompt-tuning is insufficient, as it fails to jointly represent causal
semantic attributes and image embeddings. As a result, the model struggles to achieve precise
counterfactual reasoning and generation. This highlights the need for an adaptive mechanism within
the diffusion model, enhanced with causal semantics.
\clearpage
\textbf{3. Can existing T2I based methods do fine-grained editing?} 
In causal editing, interventions on a single factor (\textit{e.g.}, increasing object size) should consistently induce predictable and proportional changes in the generated image. However, with current T2I based methods, numerical attributes must be mapped to discrete linguistic tokens, making it difficult to express edits along a continuous range. In practice, the primary way to adjust editing strength is by tuning the classifier-free guidance scale~\citep{ho2022classifier}. As shown in Figure~\ref{fig:motivation_ADNI_3}, both NTI and MCPL fail to achieve fine-grained anatomical counterfactual editing of brain ventricular volume, even with different guidance scales.  Their text-only control mechanisms cannot reliably translate numeric interventions into gradual visual changes. This limitation is particularly problematic in medical imaging domains, where precise, numerically controlled edits (\textit{e.g.}, adjusting brain or ventricular volume in MRI scans) are essential for simulating disease progression.  

\clearpage
\section{Dataset and Implementation \footnotesize{\textbf{\hyperref[Appendix:table_of_contents]{[Back to Contents]}}}}
\label{Appendix:data_implement}

\subsection{Implementation Details}
\label{Appendix:sub_implement_details}

We build our method on the backbone of Stable Diffusion v1.5~\citep{rombach2022high}, using the same hyperparameter settings as the original implementation. The pre-trained weights are trained at a resolution of $256 \times 256$.
Our adapter network $\ddot{\epsilon}_{\psi}$ is designed as a half-copy of the diffusion U-Net backbone, consisting of the encoder and bottleneck layers.
%We obtain the pre-trained SD weights from a open-source repository\footnote{\url{https://huggingface.co/lambdalabs/miniSD-diffusers}}, which are trained at a resolution of $256 \times 256$ for fast and lightweight computation. 

Unless specified, experimental setups on CelebA and ADNI follow the benchmark protocol in~\citep{melistas2024benchmarking}, and CelebA-HQ experiments follow the setup in~\citep{rasal2025diffusion}. To ensure fairness, we adopt the same causal mechanism (normalizing flow) as described in the benchmark implementation. Experimental setup on Pendulum follows the configuration in CausalDiffAE~\citep{komanduri2024causal}. Since both approaches assume linear causal modeling, we use our constructed simple MLP-based causal mechanism.  

We construct prompts in the form of ``\texttt{a human is @ and ...}'' or ``\texttt{an MRI image is @ and ...}'', where the placeholder tokens like “@” is aligned with semantic attributes in the embedding space via PAI. Prefix words (\textit{e.g.}~``\texttt{a human is}'' and ``\texttt{an image is}'') are kept fixed across datasets, as we observed no significant performance differences when switching them. 

During training, we apply the proposed CTC loss with a temperature $\tau=0.2$ and scaling factor $\lambda=0.0005$ for all datasets. At inference, all images are first generated at $256 \times 256$ resolution and then down-sampled via bicubic interpolation to match the resolution of the original benchmarks.

For counterfactual image generation, we employ DDIM inversion~\citep{song2020denoising}. To best preserve identity, we set the inversion guidance scale to $1.0$ (no classifier-free guidance (CFG)). For the forward process, we use CFG: CelebA uses $\alpha=3.0$ and CelebA-HQ uses $\alpha=1.5$ for effective edits, while ADNI and Pendulum perform well without CFG. Optionally, we incorporate token-level attention guidance manipulation following~\citep{ju2023direct}, replacing only the intervened token attentions while keeping the others fixed.

All experiments are executed on a single NVIDIA A100 GPU. Training completes within one day, and generating a single counterfactual image takes approximately 5–7 seconds. Table~\ref{tab:para_summary} presents detailed training configurations.

\begin{table}[H]
\caption{Experimental settings for our Causal-Adapter across three datasets.}
\tiny
\label{tab:para_summary}
\centering
\setlength{\tabcolsep}{2.4pt}
\renewcommand{\arraystretch}{0.95}
\resizebox{0.9\textwidth}{!}{
\begin{tabular}{l|cccc}
\toprule
 & \textsc{CelebA} & \textsc{ADNI} & \textsc{Pendulum} & \textsc{CelebA-HQ} \\
\midrule
\textsc{Train Set Size} & 162,770  & 10,780 & 5,000  &24000 \\
\textsc{Validation Set Size} & 19,867 & 0 & 500 &3000 \\
\textsc{Test Set Size} & 19,962 & 2,240 & 2,000 &3000 \\
\textsc{Resolution} & $256 \times 256 \times 3$ & $256 \times 256 \times 3$ & $256 \times 256 \times 3$ &$256 \times 256 \times 3$ \\
\textsc{Downsampled Resolution} & $64 \times 64 \times 3$ & $192 \times 192 \times 1$ & $94 \times 94 \times 3$ & $64 \times 64 \times 3$\\
\textsc{Batch Size} & 40 & 40 & 40 &40 \\
\textsc{Training Steps} & 200k & 100k & 50k &100k \\
\midrule
\textsc{Num of Attributes} & 4 & 6 & 4 & 7\\\textsc{Prompt Template} & 
\makecell[l]{\texttt{``a human is @}\\\texttt{and * and \& and \#''}} & 
\makecell[l]{\texttt{``an MRI image is @}\\\texttt{and * and \&''}} & 
\makecell[l]{\texttt{``an image is @}\\\texttt{and * and \& and \#''}} & \makecell[l]{\texttt{``a human is @}\\\texttt{and * and \& and \#}\\\texttt{and ! and ? and \%''}} \\
\midrule
\textsc{CTC Temperature $\tau$} & 0.2 & 0.2 & 0.2 & 0.2 \\
\textsc{CTC Scale $\lambda$} & 0.0005 & 0.0005 & 0.0005 & 0.0005 \\
\textsc{DDIM Steps $T$} & 100 & 100 & 100 & 100 \\
\textsc{Guidance Scale $\alpha$} & 3.0 & 1.0 & 1.0 & 1.5 \\
\midrule
\textsc{Learning Rate} & \multicolumn{4}{c}{1e-5} \\
\textsc{Optimizer} & \multicolumn{4}{c}{AdamW (weight decay 1e-2)} \\
\textsc{Loss} & \multicolumn{4}{c}{MSE (noise prediction) + CTC (proposed)} \\
\bottomrule
\end{tabular}
}
\end{table}

\subsection{Metrics}
\label{Appendix:sub_implement_metrics}
We detail the counterfactual evaluation metrics used in our experiments below.
\paragraph{Effectiveness.}
Effectiveness measures how successfully an intervention alters the intended attributes in counterfactual images. To quantitatively evaluate effectiveness, we leverage an anti-causal predictor trained on the observed data distribution for each parent variable of the image $x$ defined in the causal graph~\citep{monteiro2023measuring}. For CelebA and ADNI, we train deep convolutional regressors as anti-causal predictors for continuous attributes (\textit{e.g.}, brain volume, ventricular volume), and deep convolutional classifiers for categorical attributes (\textit{e.g.}, age, gender, beard, bald, slice class). Both models use a ResNet-18 backbone pretrained on ImageNet, following the implementation of~\citep{melistas2024benchmarking}. For the Pendulum dataset, we train regressors for the four continuous attributes following the implementation of~\citep{komanduri2024causal}.

\paragraph{Composition.}
If an attribute variable $y_i$ is forced to take the same value $\bar{y}_i$ that it would assume without intervention, the intervention should have no effect on any other variables. This corresponds to the \emph{null intervention}, which leaves all variables unchanged and, in the generative setting, can reduce to a standard reconstruction task. To evaluate counterfactual generation under \emph{null intervention}, we perform abduction via DDIM inversion and prediction via DDIM sampling, but skip the action step that edits the original attribute value. For evaluation, we report the MAE distance between the reconstructed and input images, as well as the Learned Perceptual Image Patch Similarity (LPIPS)~\citep{zhang2018unreasonable}, which better reflects perceptual fidelity and content preservation.

\paragraph{Realism.}
Realism evaluates the perceptual quality of counterfactual images by measuring their similarity to real samples. We adopt the Fréchet Inception Distance (FID)~\citep{heusel2017gans}, which quantifies the similarity between the distribution of generated counterfactual images and the real dataset. Specifically, real and counterfactual samples are passed through an Inception v3 network~\citep{szegedy2015going} pretrained on ImageNet to extract high-level semantic feature representations, which are then used to compute the FID score.

\paragraph{Minimality.}
Minimality evaluates whether a counterfactual differs from the factual image only in the intervened parent attribute, ideally leaving all other attributes unaffected. Counterfactual Latent Divergence (CLD) quantifies this by measuring the “distance” between counterfactual and factual images in a latent space~\citep{sanchez2022diffusion}. Intuitively, CLD captures a trade-off: the counterfactual should move sufficiently away from the factual class, but not farther than real samples belonging to the counterfactual class. Following~\citet{melistas2024benchmarking}, we compute CLD using an unconditional VAE. Specifically, we measure the KL divergence between the latent distributions of real and counterfactual images. The metric is minimized when both probabilities remain low, reflecting the balance between departing from the factual class while remaining closer to the counterfactual class than unrelated real samples.
\subsection{Full Regularization Algorithm}
\label{Appendix:sub_full_regularization_Algorithm}

In the following, we present the fully regularized Causal-Adapter in Algorithms~\ref{alg:Causal-Adapter-Full-Train}–\ref{alg:Causal-Adapter-Full-Infer}. Below we summarize the training and inference algorithm in high-level description.

\textbf{Training:}
\begin{enumerate}
    \item \textit{Learn the causal mechanisms $F$:}  
    Using the predefined adjacency matrix $A$ derived from the causal graph and semantic attributes $Y$, the model learns how attributes causally influence one another.  
    This produces a standalone SCM that governs how attributes should update when interventions are applied.

    \item \textit{Train the conditional adapter:}  
    Each semantic attribute $y_i$ is projected into a token embedding $v_i(y_i)$ through Prompt-Aligned Injection (PAI).  
    With the diffusion model frozen, the adapter $\ddot{\epsilon}_{\psi}$ is trained to produce residuals $r_t$ conditioned on the attribute embeddings $V(Y)$.  
    A diffusion loss $\mathcal{L}_{\text{DM}}$ ensures correct denoising behaviour, while a contrastive loss $\mathcal{L}_{\text{CTC}}$ encourages disentanglement and suppresses spurious attribute correlations.
\end{enumerate}

\textbf{Inference (Counterfactual Reasoning):}
\begin{enumerate}
    \item \textit{Abduction:}  
    The input image $x$ is inverted through the diffusion model (via DDIM inversion $H_\theta$) to obtain a latent trajectory $[z_{0}^{\star},\ldots,z_{T}^{\star}]$ consistent with the observed image.

    \item \textit{Action:}  
    A user-specified intervention $y_i'$ is applied to the semantic attributes.  
    The learned causal mechanisms update all causally connected attributes to produce a new attribute set $\bar{y}_i'$ that respects the causal graph.

    \item \textit{Prediction:}  
    Starting from the abducted latent $\bar{z}_T$, the model synthesizes a counterfactual image using the conditional adapter $\ddot{\epsilon}_{\psi}$ and the updated token embeddings
    $V\!\left(Y_{\mathrm{do}(y_i = y_i')}\right)$, yielding the final counterfactual image $\bar{x}$.
\end{enumerate}

\begin{figure}[h]
\vspace{-3mm}
\centering

%================ Training =================
\begin{minipage}[ht]{\textwidth}
\begin{algorithm}[H]
\footnotesize
\caption{Causal-Adapter: Training}
\label{alg:Causal-Adapter-Full-Train}
\begin{algorithmic}[1]

% -------- INPUT / OUTPUT --------
\STATE \textbf{Input:} Image $x$, semantic attributes $Y=\{y_i\}_{i=1}^{K}$, binary adjacency matrix $A\in\{0,1\}^{K\times K}$, frozen modules $\{\mathcal{E}, c_\phi, \epsilon_\theta\}$, projector $G=\{g_i\}_{i=1}^{K}$
\STATE \textbf{Output:} causal mechanisms $F=\{f_i\}_{i=1}^{K}$, updated placeholder embeddings $C=\{c_i\}_{i=1}^{K}$, updated projector $G=\{g_i\}_{i=1}^{K}$, causal adapter $\ddot{\epsilon}_\psi$.

% -------- STAGE 1: CAUSAL MECHANISMS --------

\STATE \textcolor{blue}{\texttt{\# Train causal mechanisms $F$}}
\STATE $\bar{y}_i \coloneqq f_i(A_i \odot Y; \omega_i) + u_i, \quad u_i \sim \mathcal{N}(0, \sigma_i^2),$
\STATE $F\coloneqq\arg\min_{F} \mathcal{L}_{\text{NLL}}(y_i, \bar{y_i})$

% -------- STAGE 2: TOKEN EMBEDDING CONSTRUCTION --------

\STATE \textcolor{blue}{\texttt{\# Construct attribute injected token embeddings}}
\STATE \textbf{initialize}~placeholder embeddings $C=\{c_i\}_{i=1}^{K}$,
\STATE $v_i(y_i) = c_i + g_i(y_i)$, $i=\{1,\cdots,K\}$
\STATE $V(Y) = [v_1(y_1),\ldots,v_K(y_K)]^{\top}$

% -------- STAGE 3: ADAPTER TRAINING --------

\STATE \textcolor{blue}{\texttt{\# Train conditional adapter}}
\FOR{$ t = 1$ to $T$}
    \STATE Encode latent: $z_{t} = \mathcal{E}(x,t)$
    \STATE Compute residual from causal adapter: $r_t = \ddot{\epsilon}_{\psi}(z_t, t, V(Y))$
    \STATE Update parameters: $\psi, G, C \coloneqq \arg\min_{\psi, G, C}\left(\mathcal{L}_{\text{DM}} +\mathcal{L}_{\text{CTC}}\right)$
\ENDFOR

\vspace{1mm}
\STATE \textbf{Return} $F=\{f_i\}_{i=1}^{K}$, $\ddot{\epsilon}_\psi$, $V$
\end{algorithmic}
\end{algorithm}
\end{minipage}
\hfill

%================ Inference =================
\begin{minipage}[t]{\textwidth}
\begin{algorithm}[H]
\footnotesize
\caption{Causal-Adapter: Inference}
\label{alg:Causal-Adapter-Full-Infer}
\begin{algorithmic}[1]

% -------- INPUT / OUTPUT --------
\STATE \textbf{Input:} Image $x$, semantic attributes $Y$, 
       frozen modules $\{\mathcal{E},c_\phi,\epsilon_\theta\}$, 
       trained causal mechanisms $F=\{f_i\}_{i=1}^{K}$, learned placeholder embedding $C=\{c_i\}_{i=1}^{K}$, trained causal adapter $\ddot{\epsilon_{\psi}}$, DDIM Inversion operator $H_{\theta}$ and its generative inverse $H_{\theta}^{-1}$
\STATE \textbf{Output:} Counterfactual image $\bar{x}$

% -------- INITIALIZATION --------

%\STATE $z_{0} = \mathcal{E}(x,0)$
\STATE $v_i(y_i) = c_i + g_i(y_i)$, \quad $i=\{1,\cdots,K\}$
\STATE $V(Y) = [v_1(y_1),\ldots,v_K(y_K)]^\top$

% -------- A: ABDUCTION --------

\STATE \textcolor{blue}{\texttt{\# Abduction – infer inversed latent noise}}
\STATE $z_{0}^{\star} = \mathcal{E}(x,0)$
\FOR{$t = 0$ to $T-1$}
    \STATE $r_t = \ddot{\epsilon}_{\psi}(z_t^{\star}, t, V(Y))$
    \STATE $z_{t+1}^\star = H_{\theta}(z_{t}^{\star}, r_t, V(Y), t)$
\ENDFOR

% -------- B: ACTION --------

\STATE \textcolor{blue}{\texttt{\# Action – apply intervention with $do(y_i=y_{i}^{\prime})$}}
%\STATE Obtain $\bar{Y}$ using $do(y_i=y_{i}^{\prime})$ via Eqn.~\ref{eqn:anm}
\STATE $\bar{y}_{i}^{\prime} \coloneqq f_i(A_i \odot Y_{do(y_i = y_{i}^{\prime})}; \omega_i) + u_i,\quad u_i \sim \mathcal{N}(0, \sigma_i^2),\quad i=\{1,\cdots,K\}$
\STATE $v_i(\bar{y}_{i}^{\prime}) = c_i + g_i(\bar{y}_{i}^{\prime}),\quad i=\{1,\cdots,K\}$
\STATE $V(Y_{do(y_{i} = y_{i}^{\prime})}) = [v_1(\bar{y}_{1}^{\prime}),\ldots,v_K(\bar{y}_{K}^{\prime})]^\top$

% -------- C: PREDICTION --------

\STATE \textcolor{blue}{\texttt{\# Prediction – generate counterfactual}}
\STATE \textbf{initialize} $\bar{z}_T \leftarrow z_T^\star$
\FOR{$t = T$ to $1$}
    \STATE $\bar{r}_t = \ddot{\epsilon}_{\psi}(z_t^\star, t, V(Y_{do(y_{i} = y_{i}^{\prime})}))$
    \STATE $\bar{z}_{t-1} = H_{\theta}^{-1}(\bar{z}_{t}, \bar{r}_t, V(Y_{do(y_{i} = y_{i}^{\prime})}), t)$
\ENDFOR

$\bar{x} = \mathrm{Decode}(\bar{z}_0)$

\STATE \textbf{Return} $\bar{x}$
\end{algorithmic}
\end{algorithm}
\end{minipage}

\vspace{-3mm}
\end{figure}

\clearpage
\section{Extended Ablation Results \footnotesize{\textbf{\hyperref[Appendix:table_of_contents]{[Back to Contents]}}}}
\label{Appendix:extra_ablation}
We further investigate the impact of guidance scale (Appendix~\ref{extra_ablation:effect_guidance_scale}), DDIM steps ( Appendix~\ref{extra_ablation:effect_ddim_steps}),, and attention guidance on counterfactual generation(Appendix~\ref{extra_ablation:investigation_attentions}), and provide additional qualitative evidence (Appendix~\ref{extra_ablation:qualitative_evidence}). To reduce computational cost, these experiments are conducted on the CelebA validation set using 400 random samples.  
\subsection{Effect of Guidance Scale}
\label{extra_ablation:effect_guidance_scale}
\begin{figure}[h]
    \centering
    \includegraphics[width=1\linewidth]{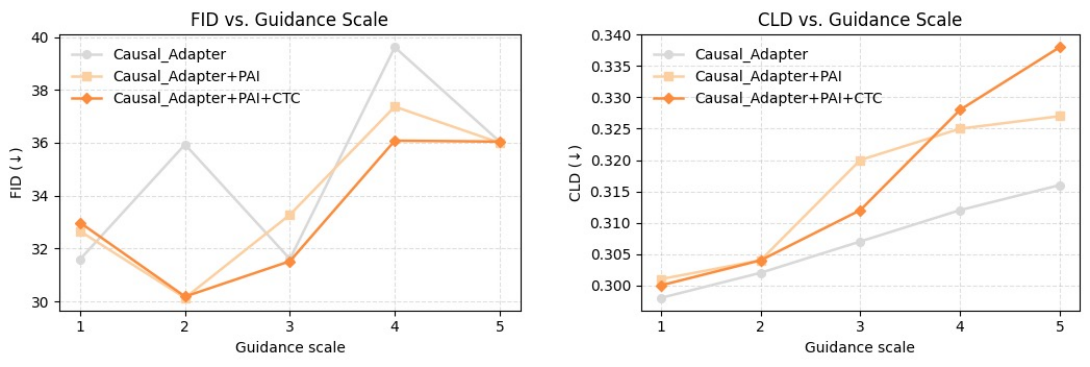}
    \caption{Impact of guidance scale on FID and CLD across three Causal-Adapter variants. Note that FID values here are higher than in the main manuscript due to the smaller evaluation set, which shifts the distribution mean and variance. The implied relative trends in image quality remain consistent.}
    \label{fig:gs_fid}
\end{figure}

\begin{figure}[!htbp]
    \centering
    \includegraphics[width=0.8\linewidth]{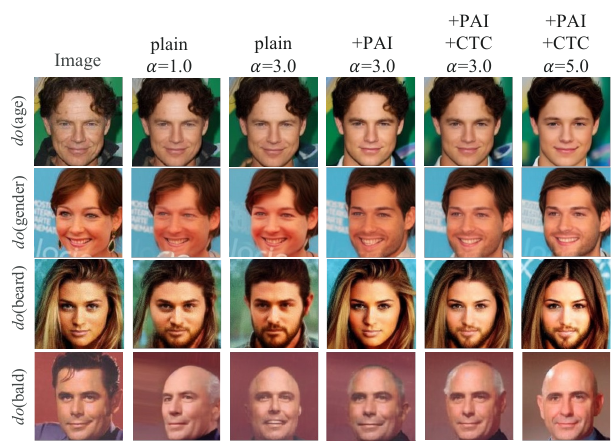}
    \caption{Counterfactuals from Causal-Adapter variants under different guidance scales. 
The plain variant shows weak effectiveness at $\alpha=1.0$ and $3.0$. 
Adding PAI strengthens the editing signal, while further incorporating CTC yields the most effective edits with strong identity preservation. 
At $\alpha=5.0$, the counterfactuals become slightly over-edited.}

    \label{fig:appendix_gs_qualitative}
\end{figure}

We study the influence of the classifier-free guidance scale $\alpha$ on intervention effectiveness, realism, and minimality, as summarized in Table~\ref{tab:guidance_scale_celeba} and Figure~\ref{fig:gs_fid}. Increasing $\alpha$ consistently improves intervention effectiveness across attributes, with all variants showing stronger F1-scores as interventions become more easily separable by the anti-causal predictor (classifier). For instance, the fully regularized model (Table~\ref{tab:guidance_scale_celeba} rows 11-15) improves Age F1 from $42.0$ at $\alpha=1.0$ to $63.4$ at $\alpha=5.0$, with similar improvement for beard and bald interventions. These results suggest that aligning semantic and spatial features and enforcing disentanglement allow more effective editing, particularly at higher guidance scales. This effectiveness improvement comes at a cost: both FID, reflecting reduced realism and causal minimality (Figure~\ref{fig:gs_fid}). Thus, we adopt $\alpha=3.0$ as the default setting, where intervention effectiveness is already strong across all variants (Table~\ref{tab:guidance_scale_celeba} rows 3,8,13) while FID and CLD remain relatively low. Further increasing $\alpha$ beyond this point will leads to marginal improvements in effectiveness but rapidly increases FID and CLD. Qualitative example is shown in Figure~\ref{fig:appendix_gs_qualitative}.
\setcounter{rownumbers}{0}
\begin{table}[h]
\caption{Influence of guidance scale on counterfactual effectiveness.
$\alpha$ denotes the classifier-free guidance scale. 
``Ours'' refers to the plain Causal-Adapter, ``Ours$^{\star}$'' to the regularized Causal-Adapter (+PAI), 
and ``Ours$^{\star\star}$'' to the fully regularized Causal-Adapter (+PAI+CTC).}
    \label{tab:guidance_scale_celeba}
    \centering
    \resizebox{\textwidth}{!}{%
    \begin{tabular}{rl cccc cccc cccc cccc}
    \toprule
    &\multirow{2}{*}{Method}
    & \multicolumn{4}{c}{\textbf{Age (a) F1} $\uparrow$}
    & \multicolumn{4}{c}{\textbf{Gender (g) F1} $\uparrow$}
    & \multicolumn{4}{c}{\textbf{Beard (br) F1} $\uparrow$}
    & \multicolumn{4}{c}{\textbf{Bald (bl) F1} $\uparrow$} \\
    \cmidrule(lr){3-6}\cmidrule(lr){7-10}\cmidrule(lr){11-14}\cmidrule(lr){15-18}
    && $do(a)$ & $do(g)$ & $do(br)$ & $do(bl)$
    & $do(a)$ & $do(g)$ & $do(br)$ & $do(bl)$
    & $do(a)$ & $do(g)$ & $do(br)$ & $do(bl)$
    & $do(a)$ & $do(g)$ & $do(br)$ & $do(bl)$ \\
    \midrule
    \rownum &Ours$(\alpha=1.0)$
    &  38.0 & 79.9 & 86.0 & 77.8
    & 96.1 & 97.1 & 65.1& 86.6
    &95.4	&75	&32.2	&93.7
    &100	&39.7	&66.7	&22.9 \\
    \rownum &Ours$(\alpha=2.0)$
    & 38.1	&80.2	&95.8	&75.5	&95.5	&96.7	&61	&82.3
    &95	&78	&39.2	&93	&100	&50	&85.7	&32
    \\
    \rownum &Ours$(\alpha=3.0)$
  &38.2	&81.1	&85.3	&73.9	&95.3	&95.6	&59.7	&77.8
    &93.9	&81.3	&42.1	&92.2	&100	&61.3	&85.7	&39.8
    \\
    \rownum &Ours$(\alpha=4.0)$
    &37.2	&80.1	&84.8	&72.6	&93.7	&95.2	&58.7	&75.4
    &92.8	&82.4	&45.2	&91.3	&100	&66.7	&75	&50.2
    \\
    \rownum &Ours$(\alpha=5.0)$
    &37.1	&80.7	&83.1	&71	&92.8	&94.8	&57.8	&71.5
     &92.3	&84.1	&43.4	&90.1	&80	&67	&85.7	&54.9 \\
    \midrule
    \rownum &Ours$^{\star}$$(\alpha=1.0)$
    &37.5	&78.1	&87.6	&70.8	&96.1	&96.8	&71.7	&86.1
    &95.1	&77.4	&29	&93.3	&100	&44.2	&85.7	&22.5
    \\
    \rownum &Ours$^{\star}$$(\alpha=2.0)$
    &46.8	&85.4	&93	&84.3	&99	&99.8	&72.8	&95.5
    &98.6	&93.7	&43.8	&97	&50	&84.8	&50	&47.2 
    \\
    \rownum &Ours$^{\star}$$(\alpha=3.0)$
 &48.5	&90.8	&92.7	&89.4	&99	&100	&72.1	&97
    &99.4	&97.4	&50.7	&97.3	&50	&91.8	&66.7	&55.8
    \\
    \rownum &Ours$^{\star}$$(\alpha=4.0)$
    &48.8	&90.2	&92.2	&90.6	&99.7	&100	&71.1	&97.5
    &99.7	&98.6	&53	&98.6	&50	&94.7	&54.5	&57.7
    
    \\
    \rownum &Ours$^{\star}$$(\alpha=5.0)$
    &48.6	&90	&91.6	&91.8	&100	&100	&70.5	&97.5
     &100	&99.1	&54.1	&99	&66.7	&94.3	&40	&58.2
     \\
    \midrule
    \rownum &Ours$^{\star\star}$$(\alpha=1.0)$
    &  42.0 & 80.2 & 86.0 & 73.6
    & 96.2 & 97.7 & 67.8 & 81.5
    & 94.9 & 78.0 & 31.5 & 92.9
    & 1  & 45.9 & 75.0  & 29.4 \\
    \rownum &Ours$^{\star\star}$$(\alpha=2.0)$
    & 51.9 & 88.0 & 93.4 & 84.9
    & 98.1 & 99.6 & 72.3 & 90.1
    & 98.8 & 91.0 & 46.4 & 97.0
    & 80.0  & 84.2 & 75.0  & 55.1 \\
    \rownum &Ours$^{\star\star}$$(\alpha=3.0)$
    & 58.8 & 91.3 & 94.4 & 89.4
    & 100 & 99.8 & 74.5 & 91.1
    & 99.7 & 96.4 & 53.4 & 97.8
    & 80.0  & 92.4 & 66.7  & 59.8 \\
    \rownum &Ours$^{\star\star}$$(\alpha=4.0)$
    & 61.3 & 89.9 & 94.7 & 91.5
    & 99.7 & 100 & 77.4 & 92.2
    & 100& 97.2 & 55.6 & 97.6
    & 75.0 & 92.4 & 66.7 & 59.8 \\
    \rownum &Ours$^{\star\star}$$(\alpha=5.0)$
    & 63.4 & 90.7 &95.2& 91.4
    & 99.7 & 100 &79.0  & 93.5
    & 100 & 98.0 & 57.0 & 97.8
    & 75.0 & 92.4 &66.7  & 63.0 \\
    
    \bottomrule
    
    \end{tabular}
    }
\end{table}

\clearpage
\subsection{Effect of DDIM Steps}
\label{extra_ablation:effect_ddim_steps}
We further evaluate the effect of DDIM steps on counterfactual effectiveness using our fully regularized model (Table~\ref{tab:timesteps_celeba}). As $T$ increases, intervention effectiveness remains stable, indicating that performance is insensitive to the number of denoising steps. Realism and minimality also show negligible variation. The main drawback of larger $T$ lies in increased inference time: generating one counterfactual takes 5–7 seconds at $T=100$, about 20 seconds at $T=200$, and nearly 40 seconds at $T=500$. Balancing efficiency and accuracy, we adopt $T=100$ as the default setting for all experiments.
\setcounter{rownumbers}{0}
\begin{table}[h]
    \caption{Influence of DDIM steps under $\alpha=3.0$ on counterfactual effectiveness. $T$ denote the used steps.}
    \label{tab:timesteps_celeba}
    \centering
    \resizebox{\textwidth}{!}{%
    \begin{tabular}{rl cccc cccc cccc cccc}
    \toprule
    &\multirow{2}{*}{Method}
    & \multicolumn{4}{c}{\textbf{Age (a) F1} $\uparrow$}
    & \multicolumn{4}{c}{\textbf{Gender (g) F1} $\uparrow$}
    & \multicolumn{4}{c}{\textbf{Beard (br) F1} $\uparrow$}
    & \multicolumn{4}{c}{\textbf{Bald (bl) F1} $\uparrow$} \\
    \cmidrule(lr){3-6}\cmidrule(lr){7-10}\cmidrule(lr){11-14}\cmidrule(lr){15-18}
    && $do(a)$ & $do(g)$ & $do(br)$ & $do(bl)$
    & $do(a)$ & $do(g)$ & $do(br)$ & $do(bl)$
    & $do(a)$ & $do(g)$ & $do(br)$ & $do(bl)$
    & $do(a)$ & $do(g)$ & $do(br)$ & $do(bl)$ \\
    \midrule
    \rownum &Ours$^{\star\star}$$(T=50)$
    &  57.1 & 90.5 & 93.8 & 89.4
    & 99.7 & 99.6 & 75.7 & 91.6
    & 99.7 & 96.1 & 49.1 & 98
    & 80.0  & 89.4 & 75.0  & 60 \\
    \rownum &Ours$^{\star\star}$$(T=100)$
    & 58.8 & 91.3 & 94.4 & 89.4
    & 100 & 99.8 & 74.5 & 91.1
    & 99.7 & 96.4 & 53.4 & 97.8
    & 80.0  & 92.4 & 66.7  & 61.8 \\
    \rownum &Ours$^{\star\star}$$(T=200)$
    & 60.5 & 91.7 & 94.9 & 89.6
    & 100 & 99.8 & 74.1 & 91.4
    & 99.7 & 96.4 & 54.3 & 97.4
    & 80.0  & 91.9 & 66.7  & 61.3 \\
    \rownum &Ours$^{\star\star}$$(T=500)$
    & 60.7 & 91.6 & 94.7 & 90.0
    & 100 & 99.8 & 73.8 & 91.9
    & 99.7 & 96.1 & 54.8 & 98.0
    & 85.7  & 91.9 & 66.7  & 61.3 \\
    
    \bottomrule
    
    \end{tabular}
    }
\end{table}
\begin{figure}[h]
    \centering
\includegraphics[width=1\linewidth]{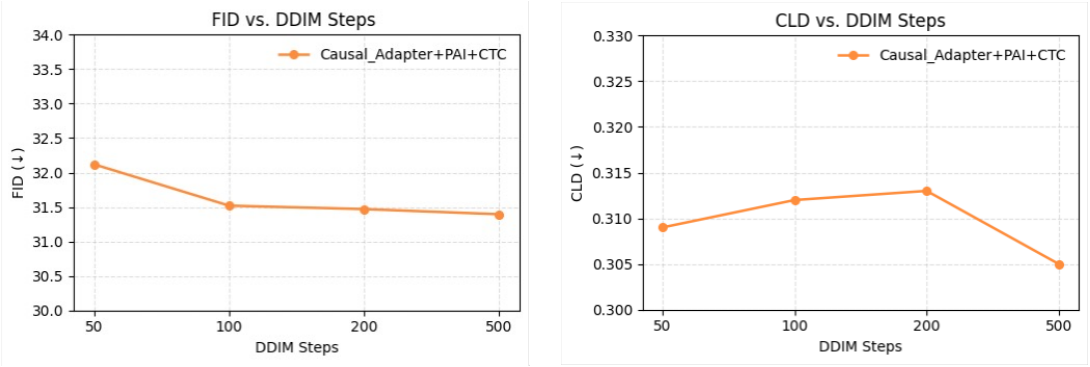}
    \caption{Impact of DDIM steps on FID and CLD}
    \label{fig:step_fid}
\end{figure}
\clearpage
\subsection{Investigation of Attention Guidance}
\label{extra_ablation:investigation_attentions}
\begin{figure}[!htbp]
    \centering
    \includegraphics[width=0.8\linewidth]{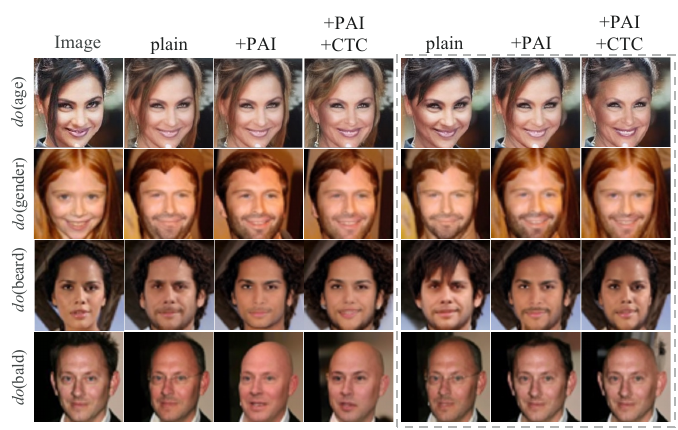}
    \caption{Full ablation visualizations with optional attention guidance (AG). 
Causal-Adapter with the two regularizers reduces spurious correlations, AG can enhance identity preservation through localized editing. 
Dotted boxes indicate results with AG-based localized edits.}
    \label{fig:appendix_ag_qualitative}
\end{figure}

\begin{figure}[!ht]
    \centering
    \includegraphics[width=1.0\linewidth]{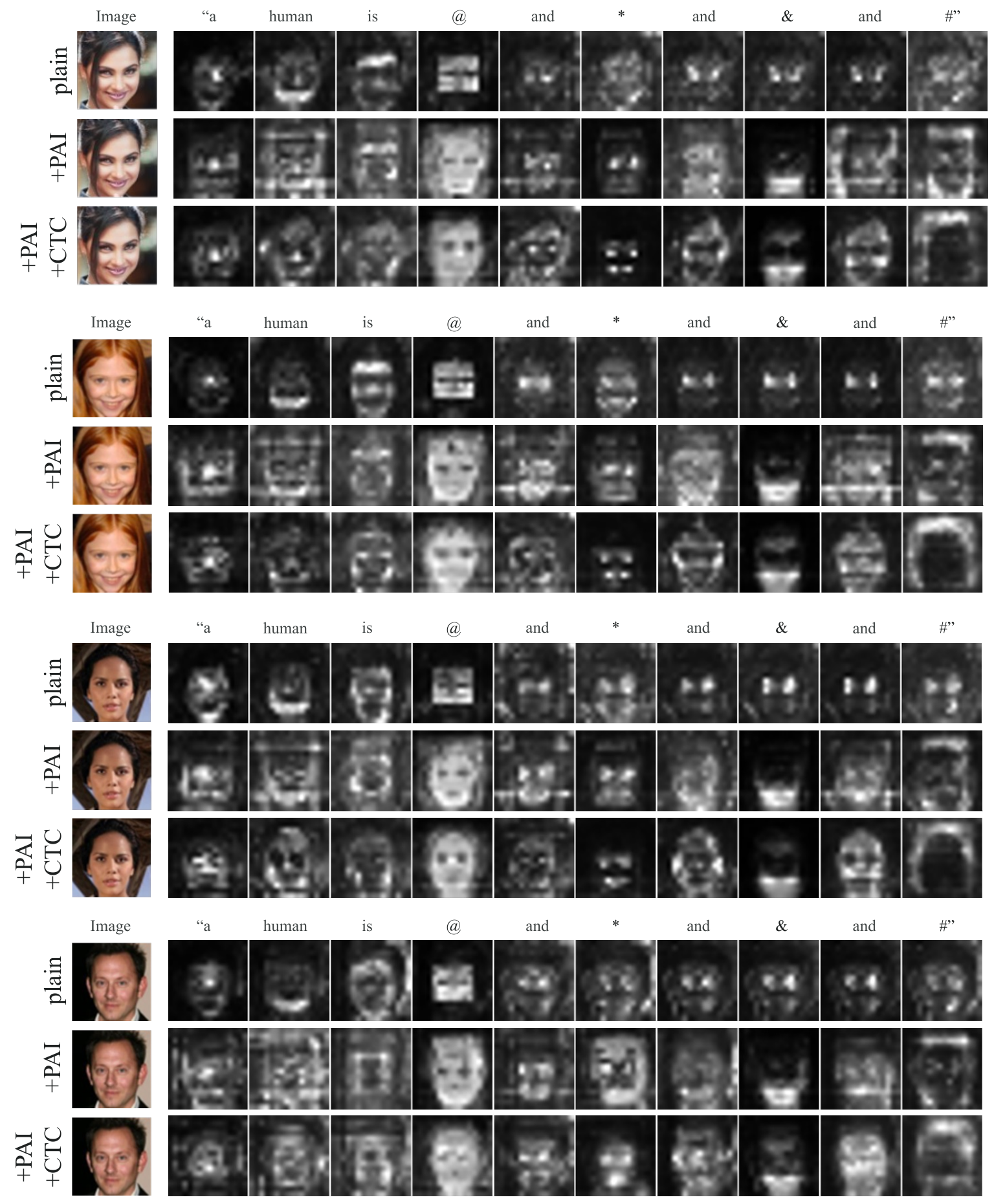}
    \caption{Average cross-attention maps from Causal-Adapter variants. Tokens denote attributes: ``@'' for age, ``*'' for gender, ``\&'' for beard, and ``\#'' for bald.
The plain adapter fails to align semantics with spatial features, producing poor maps. 
Adding PAI improves alignment but some tokens (e.g., “*”, “\#”) remain entangled. 
With both PAI and CTC, token embedding disentanglement is enforced, and attentions are clearly localized to the semantic regions.}

    \label{fig:appendix_attentions_qualitative}
\end{figure}

Attention guidance (AG) has been proposed as an external mechanism to enforce localized edits. While our primary contribution lies in Causal-Adapter and its regularizers, we further examine whether AG can complement different variants of our model. As shown in Tables~\ref{tab:attention_guidance_celeba}–\ref{tab:celea_fid_attention_guidance}, incorporating AG sometimes reduces intervention effectiveness, particularly for the plain adapter (Table~\ref{tab:attention_guidance_celeba} rows 1,3). For example, Age F1 drops sharply from $38.2$ to $28.1$ when switching from global to local editing, reflecting misaligned attention maps that disrupt counterfactual consistency. In contrast, the regularized variants exhibit only mild decreases (Table~\ref{tab:attention_guidance_celeba} rows 2,5), indicating that aligning semantic and spatial features in diffusion latents via PAI is a prerequisite for stable local editing.
Our fully regularized model (Table~\ref{tab:attention_guidance_celeba} rows 3,6) not only maintains strong intervention effectiveness but also achieves the lowest FID (31.216) with AG, highlighting that PAI and CTC help produce more precise attention maps. By comparison, applying AG to the plain adapter actually worsens FID and CLD (Table~\ref{tab:celea_fid_attention_guidance} row 4), suggesting that noisy or misaligned attention can harm editing quality. Qualitative evidence is provided in Figures~\ref{fig:appendix_ag_qualitative}–\ref{fig:appendix_attentions_qualitative}.

We emphasize that while AG can be beneficial for identity preservation in human faces (CelebA), in other domains such as ADNI or Pendulum, our model already produces accurate, faithful, and identity-preserving counterfactuals without AG. This underscores that our contribution lies not in AG itself, but in designing Causal-Adapter such that causal attributes are naturally aligned with token embeddings, enabling both disentanglement and, if desired, effective integration with AG for localized editing.
\setcounter{rownumbers}{0}
\begin{table}[h]
    \caption{Effectiveness of Causal-Adapter variants with and without attention guidance.}
    \label{tab:attention_guidance_celeba}
    \centering
    \resizebox{\textwidth}{!}{%
    \begin{tabular}{rl cccc cccc cccc cccc}
    \toprule
    &\multirow{2}{*}{Method}
    & \multicolumn{4}{c}{\textbf{Age (a) F1} $\uparrow$}
    & \multicolumn{4}{c}{\textbf{Gender (g) F1} $\uparrow$}
    & \multicolumn{4}{c}{\textbf{Beard (br) F1} $\uparrow$}
    & \multicolumn{4}{c}{\textbf{Bald (bl) F1} $\uparrow$} \\
    \cmidrule(lr){3-6}\cmidrule(lr){7-10}\cmidrule(lr){11-14}\cmidrule(lr){15-18}
    && $do(a)$ & $do(g)$ & $do(br)$ & $do(bl)$
    & $do(a)$ & $do(g)$ & $do(br)$ & $do(bl)$
    & $do(a)$ & $do(g)$ & $do(br)$ & $do(bl)$
    & $do(a)$ & $do(g)$ & $do(br)$ & $do(bl)$ \\
    \midrule
    \multicolumn{18}{l}{\emph{without attention guidance for global editing}} \\
     \rownum &Ours
    &38.2	&81.1	&85.3	&73.9	&95.3	&95.6	&59.7	&77.8
    &93.9	&81.3	&42.1	&92.2	&100	&61.3	&85.7	&39.8 \\
    \rownum &Ours$^{\star}$
     &48.5	&90.8	&92.7	&89.4	&99	&100	&72.1	&97
        &99.4	&97.4	&50.7	&97.3	&50	&91.8	&66.7	&55.8 \\
    \rownum &Ours$^{\star\star}$
    & 58.8 & 91.3 & 94.4 & 89.4
    & 100 & 99.8 & 74.5 & 91.1
    & 99.7 & 96.4 & 53.4 & 97.8
    & 80.0  & 92.4 & 66.7  & 61.8 \\
    \midrule
    \multicolumn{18}{l}{\emph{with attention guidance for local editing}} \\
     \rownum &Ours
    &28.1	&79.4	&83.2	&63.3	&94.7	&93.4	&62.4	&78.1
    &96.5	&80.6	&40.4	&89.4	&66.7	&70.6	&33.3	&34.5
    \\
    \rownum &Ours$^{\star}$
     &49.4	&88.8	&91.2	&73.9	&98.8	&96.4	&71.1	&92.8
       &99.5	&96.1	&47.4	&94.6	&22.2	&76.2	&40	&52.4
        \\
    \rownum &Ours$^{\star\star}$
    &57.1	&88.7	&94	&77	&99.4	&100	&73.3	&90.4
    &99.7	&95.6	&49.4	&95.1	&66.7	&80.7	&50	&54.4
    \\
    
    \bottomrule
    
    \end{tabular}
    }
\end{table}
\setcounter{rownumbers}{0}
\begin{table}[h]
\centering
        \caption{Realism and minimality of Causal-Adapter variants with and without attention guidance} 
        \resizebox{0.5\linewidth}{!}{ 
            \begin{tabular}{rl c c}
            \toprule
            &\multirow{2}{*}{Method} 
            & \textbf{Realism}  
            & \textbf{Minimality} \\ 
            \cmidrule(lr){3-3}\cmidrule(lr){4-4}
            &&FID $\downarrow$ &CLD $\downarrow$  \\
            \midrule
            \multicolumn{4}{l}{\emph{without attention guidance for global editing}} \\
            \rownum &Ours   & 31.604 & 0.307 \\
            
            \rownum &Ours$^{\star}$   & 33.278 & 0.320 \\
            
            \rownum &Ours$^{\star\star}$   & 31.52 & 0.312 \\
            \midrule
            \multicolumn{4}{l}{\emph{with attention guidance for localized editing}} \\
            \rownum &Ours & 33.849 & 0.311 \\
            \rownum &Ours$^{\star\star}$   & 31.885 & 0.299 \\
            
            \rownum &Ours$^{\star\star}$   & 31.216 & 0.300 \\
            
            \bottomrule
            \end{tabular}
            
        }
        \label{tab:celea_fid_attention_guidance}
\end{table}
\clearpage
\subsection{Qualitative Evidence}
\label{extra_ablation:qualitative_evidence}

\clearpage
\section{Additional Results \footnotesize{\textbf{\hyperref[Appendix:table_of_contents]{[Back to Contents]}}}}
\label{Appendix:addition_results}
\subsection{Pendulum}
\label{Appendix:extra_pend}

\begin{figure}[h]
    \centering
    \includegraphics[width=\linewidth]{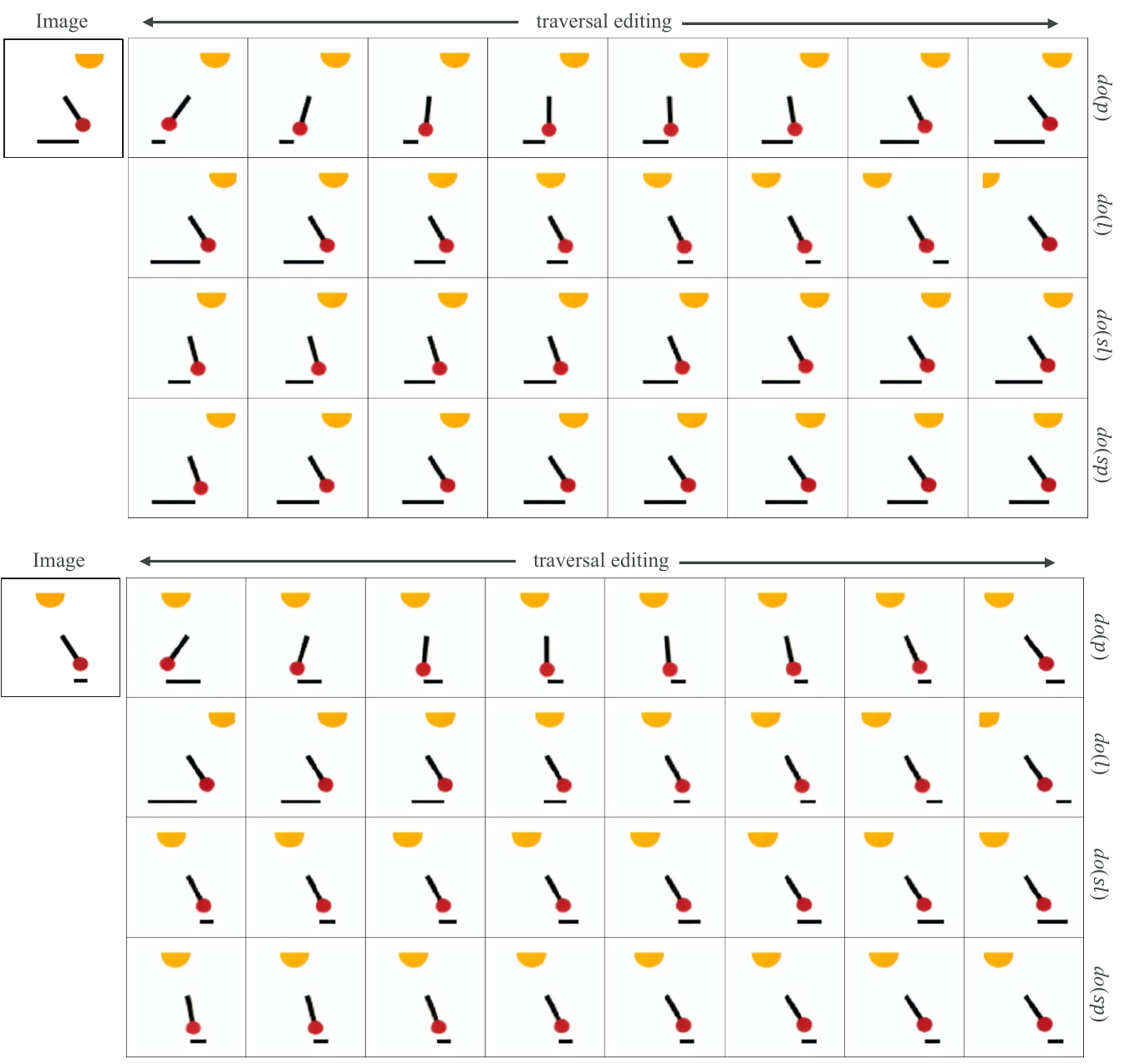}
    \caption{Pendulum counterfactuals from Causal-Adapter. $p$ for pendulum, $l$ for light, $sl$ for shadow length and $sp$ for shadow position.Traversal editing across four attributes
demonstrates that our method produces high-quality, fine-grained generations of attribute variations.}
    \label{fig:pend_extra_quality_1}
\end{figure}
\begin{figure}[!ht]
    \centering
    \includegraphics[width=\linewidth]{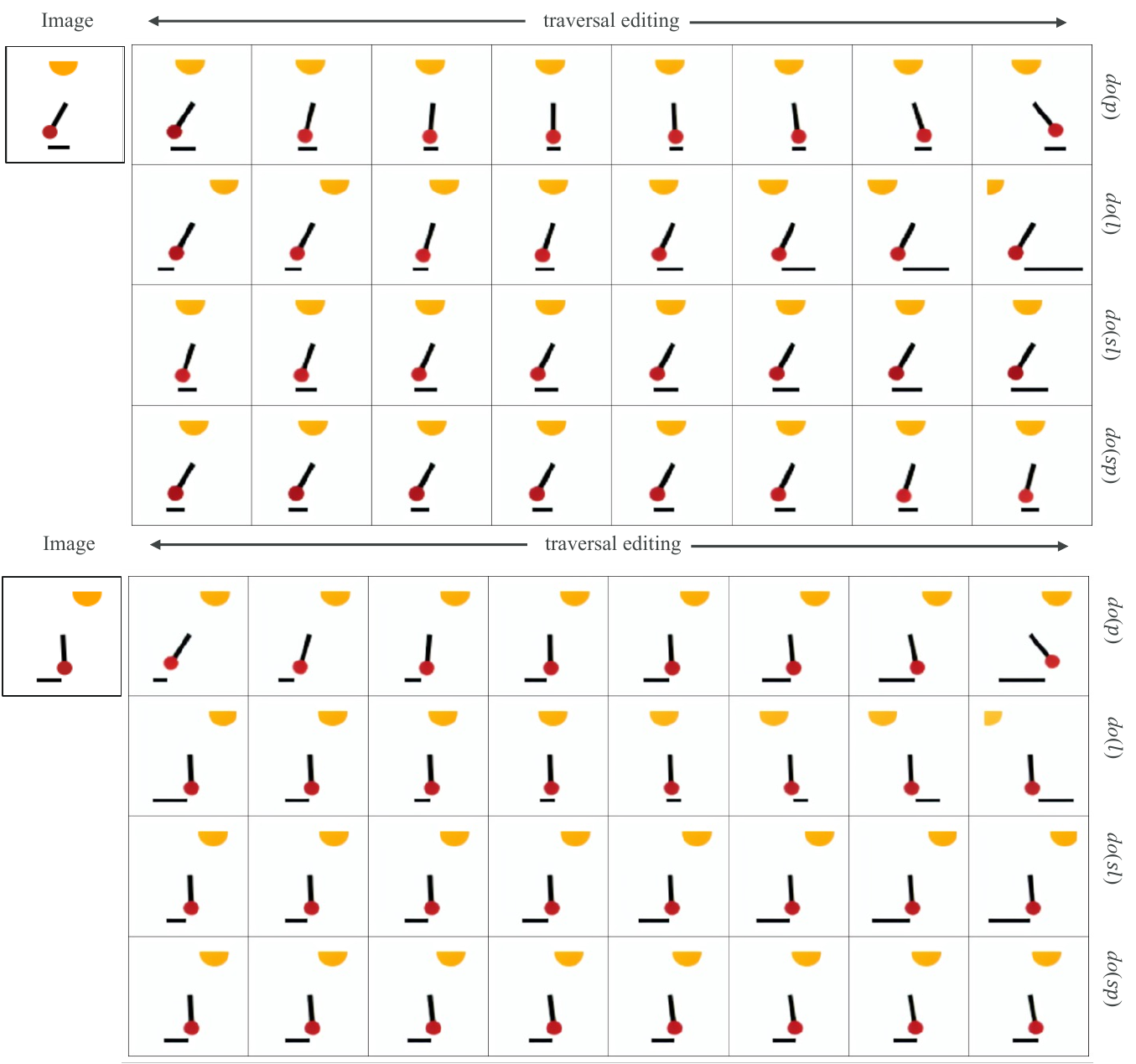}
    \caption{Pendulum counterfactuals from Causal-Adapter. $p$ for pendulum, $l$ for light, $sl$ for shadow length and $sp$ for shadow position.}
    \label{fig:pend_extra_quality_2}
\end{figure}

\clearpage
\subsection{CelebA}
\label{Appendix:extra_celeba}
\begin{figure}[h]
    \centering
    \includegraphics[width=\linewidth]{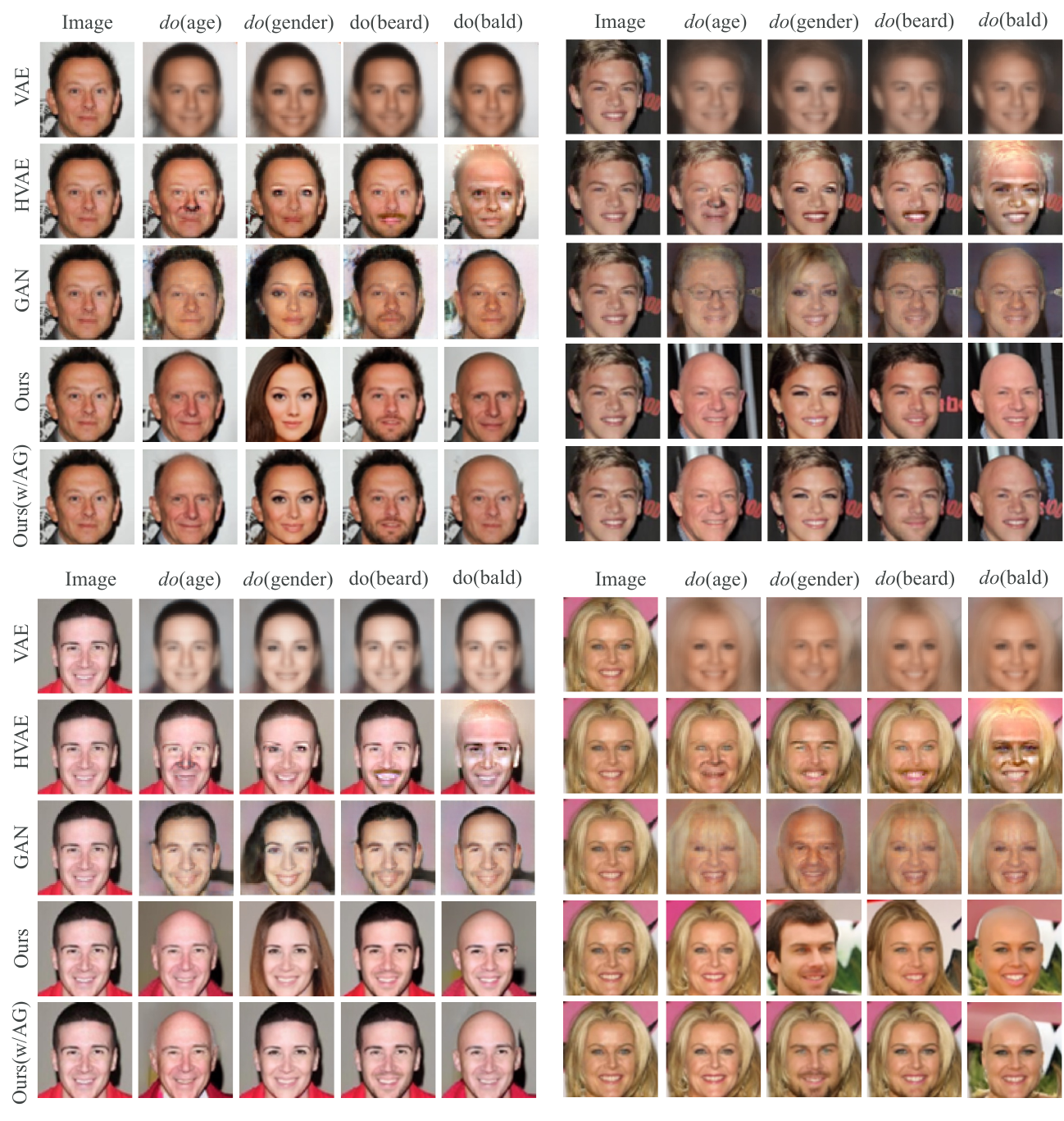}
    \caption{Additional counterfactual results on the CelebA dataset (with edit samples selected in a non-cherrypicked manner following~\citep{melistas2024benchmarking}). Our Causal-Adapter effectively disentangles target attributes compared with prior methods and achieve faithful counterfactual generations.}
    \label{fig:celeba_extra_quality_1}
\end{figure}

We report the full CelebA results with the corresponding means and standard deviations in Tables~\ref{tab:celeba_age_gender}--\ref{tab:celea_fid_std}. Compared with VAE, HVAE, and GAN, our method achieves best performance across most interventions, including up to an $86\%$ reduction in LPIPS (composition, from 0.122 to 0.017) and an $79\%$ reduction in FID (realism, from 27.861 to 8.152). Besides, our method delivers significant improvements under $do(a)$ and $do(br)$ on F1 score of bald attribute, demonstrating causal faithfulness in representing baldness across diverse individuals. As illustrated in Figure~\ref{fig:celeba_extra_quality_1}, Causal-Adapter can successfully add baldness to both male and female under $do(bl)$, and under interventions on causal parents such as $do(a)$, it can jointly edit age and baldness, whereas baselines fail to maintain causal consistency.

Table~\ref{tab:celea_fid_std} additionally reports the LPIPS distance between each counterfactual and its original image, measuring identity preservation under different attribute interventions. 
% We conduct an additional composition evaluation by iterative reconstruction, 
% where each output is recursively fed back as input for 10 cycles following~\citep{melistas2024benchmarking}. 
% As shown in Table~\ref{tab:celea_composition_10}, our method achieves the lowest LPIPS score, 
% indicating better preservation of identity information after 10 reconstructions, 
% while also delivering competitive MAE performance. Additional qualitative results are presented in Figure~\ref{fig:celeba_extra_quality_1}-~\ref{fig:celeba_extra_quality_2}.

\setcounter{rownumbers}{0}
\begin{table}[ht!]
     \caption{Intervention effectiveness on CelebA test set. Age and Gender F1 under four interventions.}
    \scriptsize
    \label{tab:celeba_age_gender}
    \centering
    \resizebox{\textwidth}{!}{%
    \begin{tabular}{rlcccccccc}
    \toprule
    &\multirow{2}{*}{Method}
    & \multicolumn{4}{c}{\textbf{Age $(a)$ F1} $\uparrow$}
    & \multicolumn{4}{c}{\textbf{Gender $(g)$ F1} $\uparrow$} \\
    \cmidrule(lr){3-6}\cmidrule(lr){7-10}
    && $do(a)$ & $do(g)$ & $do(br)$ & $do(bl)$
     & $do(a)$ & $do(g)$ & $do(br)$ & $do(bl)$ \\
    \midrule

    \rownum &VAE
    & 35.0$_{0.04}$ & 78.2$_{0.02}$ & 81.6$_{0.02}$ & 81.9$_{0.02}$
    & 97.7$_{0.01}$ & 90.9$_{0.02}$ & 95.9$_{0.02}$ & \textbf{97.3}$_{0.01}$ \\

    \rownum &HVAE
    & \textbf{65.4}$_{0.10}$ & 89.3$_{0.04}$ & 90.8$_{0.03}$ & \textbf{89.9}$_{0.03}$
    & 98.8$_{0.02}$ & 94.9$_{0.03}$ & \textbf{99.4}$_{0.01}$ & 95.0$_{0.03}$ \\

    \rownum &GAN
    & 41.3$_{0.04}$ & 71.0$_{0.02}$ & 81.8$_{0.02}$ & 79.9$_{0.01}$
    & 95.2$_{0.01}$ & 98.2$_{0.01}$ & 92.0$_{0.01}$ & 96.1$_{0.01}$ \\

    \rownum &Ours
    & 58.5$_{0.14}$ & \textbf{89.9}$_{0.35}$ & 94.0$_{0.07}$ & 89.4$_{0.00}$
    & \textbf{99.6}$_{0.00}$ & \textbf{99.9}$_{0.00}$ & 74.5$_{0.07}$ & 92.5$_{0.21}$ \\

    \midrule
    \multicolumn{10}{l}{\emph{with attention guidance for localized editing}} \\

    \rownum &Ours (AG)
    & 57.1$_{0.14}$ & 89.5$_{1.76}$ & \textbf{94.5}$_{0.00}$ & 81.5$_{0.63}$
    & \textbf{99.6}$_{0.00}$ & 99.7$_{0.00}$ & 73.8$_{0.56}$ & 89.2$_{0.14}$ \\

    \bottomrule
    \end{tabular}
    }
\end{table}
\setcounter{rownumbers}{0}
\begin{table}[ht!]
    
    \caption{Intervention effectiveness on CelebA test set. Beard and Bald F1 under four interventions.}
    \scriptsize
    \label{tab:celeba_beard_bald}
    \centering
    \resizebox{\textwidth}{!}{%
    \begin{tabular}{rlcccccccc}
    \toprule
    &\multirow{2}{*}{Method}
    & \multicolumn{4}{c}{\textbf{Beard $(br)$ F1} $\uparrow$}
    & \multicolumn{4}{c}{\textbf{Bald $(bl)$ F1} $\uparrow$} \\
    \cmidrule(lr){3-6}\cmidrule(lr){7-10}
    && $do(a)$ & $do(g)$ & $do(br)$ & $do(bl)$
     & $do(a)$ & $do(g)$ & $do(br)$ & $do(bl)$ \\
    \midrule

    \rownum &VAE
    & 94.4$_{0.01}$ & 82.8$_{0.03}$ & 29.6$_{0.05}$ & 94.5$_{0.02}$
    & 2.3$_{0.03}$ & 49.6$_{0.05}$ & 4.5$_{0.04}$ & 41.2$_{0.03}$ \\

    \rownum &HVAE
    & 95.2$_{0.03}$ & 95.1$_{0.03}$ & 44.1$_{0.11}$ & 91.6$_{0.04}$
    & 2.0$_{0.05}$ & 86.0$_{0.05}$ & 4.5$_{0.07}$ & \textbf{61.1}$_{0.04}$ \\

    \rownum &GAN
    & 90.8$_{0.01}$ & 83.8$_{0.02}$ & 23.3$_{0.03}$ & 90.7$_{0.01}$
    & 2.1$_{0.02}$ & 82.0$_{0.02}$ & 5.5$_{0.02}$ & 49.2$_{0.02}$ \\

    \rownum &Ours
    & \textbf{99.7}$_{0.00}$ & 96.1$_{0.00}$ & \textbf{52.1}$_{0.21}$ & \textbf{98.1}$_{0.07}$
    & \textbf{59.2}$_{0.91}$ & \textbf{91.2}$_{0.28}$ & \textbf{74.5}$_{2.75}$ & 58.8$_{0.35}$ \\

    \midrule
    \multicolumn{10}{l}{\emph{with attention guidance for localized editing}} \\

    \rownum &Ours (AG)
    & 99.7$_{0.07}$ & \textbf{96.8}$_{0.00}$ & 48.2$_{0.14}$ & 96.6$_{0.07}$
    & 38.7$_{1.90}$ & 78.4$_{0.49}$ & 47.7$_{1.76}$ & 51.4$_{0.49}$ \\

    \bottomrule
    \end{tabular}
    }
\end{table}

\setcounter{rownumbers}{0}
\begin{table}[ht]
    \centering
    \caption{
        Identity preservation, realism, and minimality on the CelebA test set.
        VAE, HVAE, and GAN are reported as single-run results from prior work,
        while our methods report mean$_{\text{std}}$ over three random seeds.
        ``--'' indicates that LPIPS was not reported for the corresponding method.
    }
    \scriptsize
    \resizebox{\linewidth}{!}{ 
        \begin{tabular}{rlcccccccc}
            \toprule
            & \multirow{2}{*}{Method} 
            & \multicolumn{4}{c}{\textbf{Identity Preservation} (LPIPS $\downarrow$)} 
            & \multicolumn{2}{c}{\textbf{Composition}} 
            & \textbf{Realism} 
            & \textbf{Minimality} \\ 
            \cmidrule(lr){3-6}\cmidrule(lr){7-8}\cmidrule(lr){9-9}\cmidrule(lr){10-10}
            && Age & Gender & Bearded & Bald & MAE $\downarrow$ & LPIPS $\downarrow$& FID $\downarrow$ & CLD $\downarrow$  \\
            \midrule
            \rownum & VAE          
                   & -- & -- & -- & -- 
                   & 18.695 & 0.282 & 59.393 & \textbf{0.299} \\
            \rownum & HVAE         
                   & -- & -- & -- & -- 
                   & 7.143 & 0.122& 35.712 & 0.305 \\
            \rownum & GAN          
                   & -- & -- & -- & -- 
                   & 60.120 & 0.276& 27.861 & 0.304 \\
            \rownum & Ours         
                   & 0.087$_{0.0066}$ & 0.157$_{0.0067}$ 
                   & 0.0797$_{0.0060}$ & 0.157$_{0.0073}$ & \textbf{3.535$_{0.000}$} & \textbf{0.017$_{0.000}$}
                   & 8.152$_{0.365}$ & 0.310$_{0.001}$ \\
            \midrule
            \multicolumn{8}{l}{\emph{Ours with attention guidance for improved realism and minimality}} \\
            \rownum & Ours (AG)    
                   & \textbf{0.061}$_{0.0171}$ & \textbf{0.072}$_{0.0178}$ 
                   & \textbf{0.0330}$_{0.0088}$ & \textbf{0.109}$_{0.0192}$ 
                   & -- & --
                   & \textbf{5.213}$_{0.173}$ & 0.301$_{0.001}$ \\
            \bottomrule
        \end{tabular}    
    }
    \label{tab:celea_fid_std}
\end{table}

\begin{figure}[h]
    \centering
    \includegraphics[width=\linewidth]{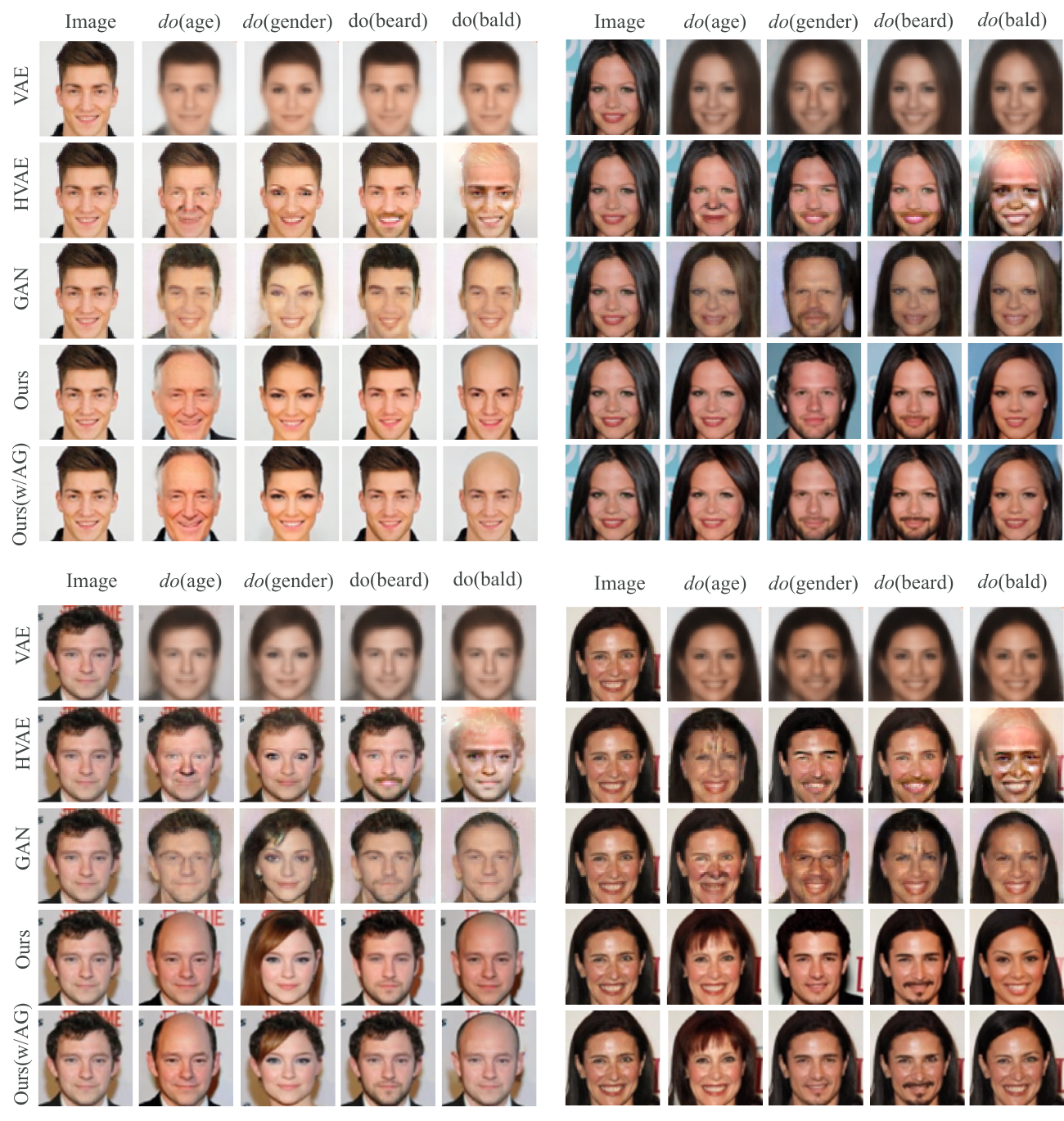}
    \caption{Additional counterfactual results on the CelebA dataset (with edit samples selected in a non-cherrypicked manner following~\citep{melistas2024benchmarking}).}
    \label{fig:celeba_extra_quality_2}
\end{figure}
\clearpage
\subsection{ADNI}
\label{Appendix:extra_adni}
\begin{figure}[!ht]
    \centering
    \includegraphics[width=0.93\linewidth]{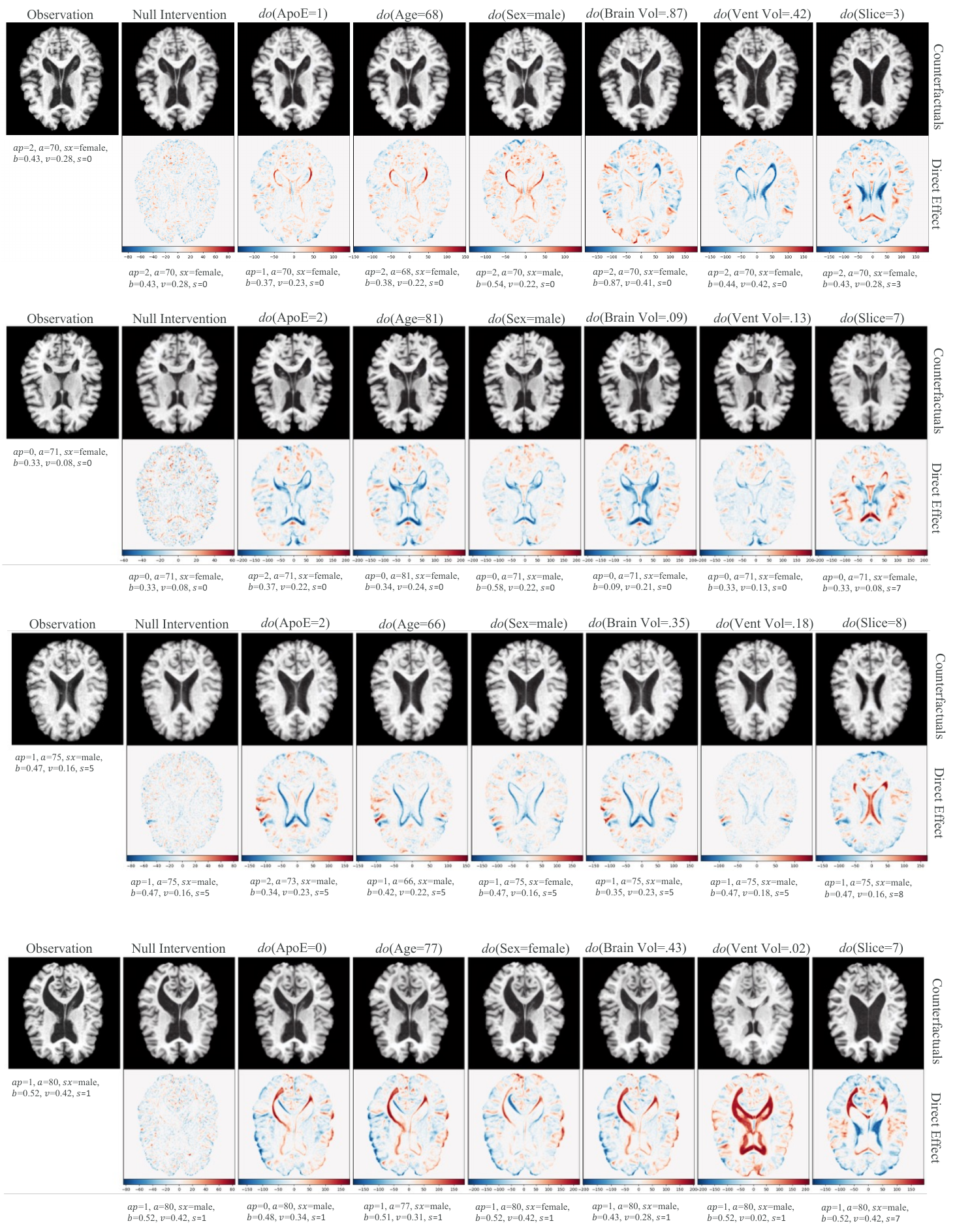}
    \caption{Additional counterfactual results from random interventions on each attribute in the ADNI dataset (non-cherrypicked). We observe localized changes consistent with the performed interventions and the assumed causal graph. Importantly, the identity of the original observation is well preserved, demonstrating the effectiveness of Causal-Adapter.}
    \label{fig:adni_extra_quality_1}
\end{figure}

\begin{figure}[!ht]
    \centering
    \includegraphics[width=0.93\linewidth]{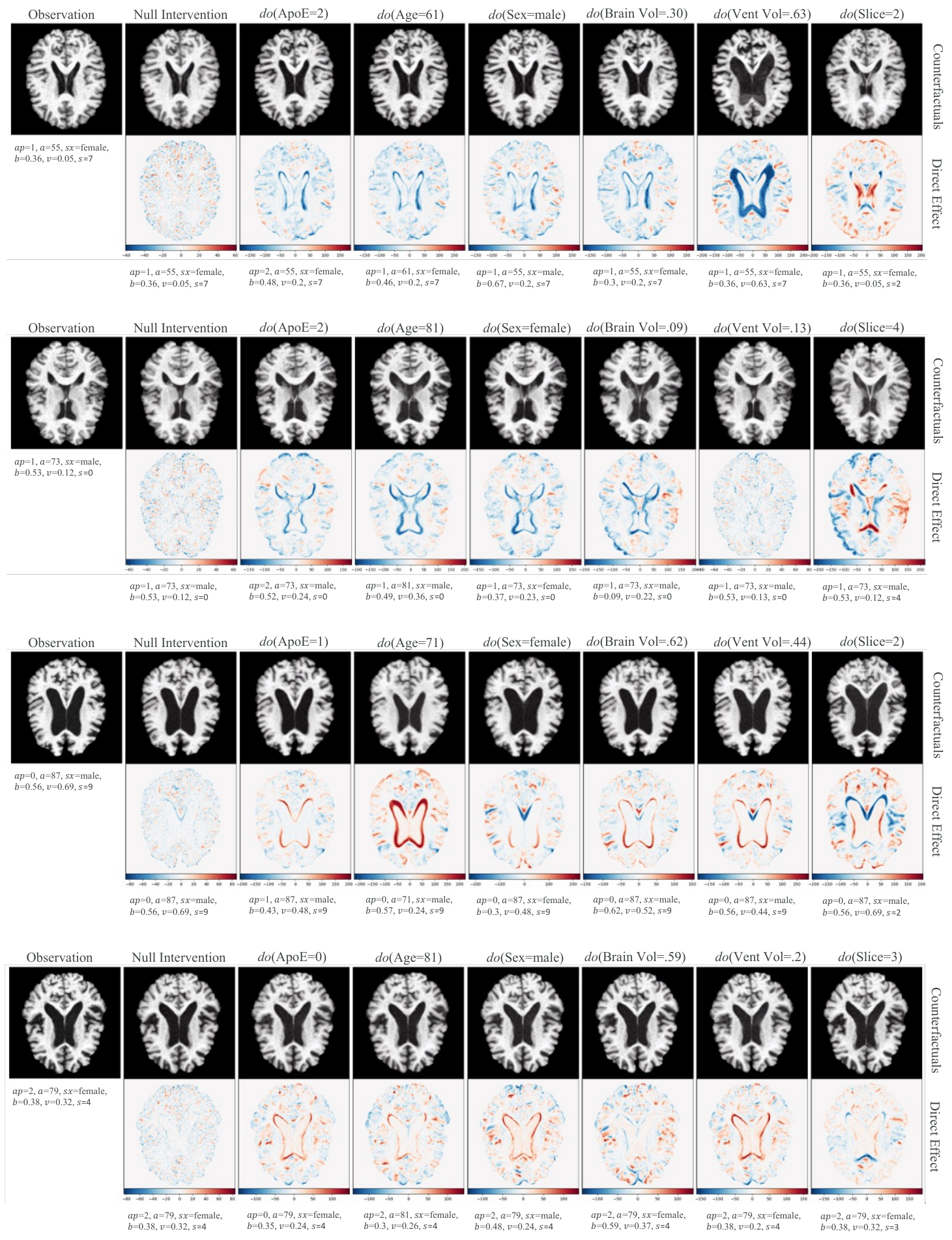}
    \caption{Additional counterfactual results from random interventions on each attribute in the ADNI dataset}
    \label{fig:adni_extra_quality_2}
\end{figure}

We intervene on three generative conditioning attributes (Brain Volume, Ventricular Volume, Slice) and report the results in Table~\ref{tab:effectiveness_benchmark_adni_full}. Our approach achieves best performance in intervention effectiveness and minimality, while also delivering strong realism. Qualitative results in Figure~\ref{fig:adni_extra_quality_1}-\ref{fig:adni_extra_quality_2} further show that our model produces sharp and localized interventional changes consistent with the causal graph. For example, edits to ApoE, Age, or Sex appropriately influence Brain and Ventricular Volumes. In particular, fine-grained edits to Ventricular Volume visibly adjust the ventricle region while faithfully preserving subject identity.

\setcounter{rownumbers}{0}
\begin{table*}[h]
\caption{Effectiveness on ADNI test set, evaluated with MAE/F1 from pretrained regressors/classifiers (std. dev. reported). Composition, realism, and minimality metrics are also included.}
\centering
\resizebox{\textwidth}{!}{%
\begin{tabular}{rlccc ccc ccc | cc c c}
\toprule
&\multirow{2}{*}{Method} 
& \multicolumn{3}{c}{\textbf{Brain volume (b) MAE} $\downarrow$} 
& \multicolumn{3}{c}{\textbf{Ventricular volume (v) MAE} $\downarrow$}  
& \multicolumn{3}{c}{\textbf{Slice (s) F1} $\uparrow$} 
& \multicolumn{2}{c}{\textbf{Composition} } 
& \textbf{Realism} 
& \textbf{Minimality} \\
\cmidrule(lr){3-5} \cmidrule(lr){6-8} \cmidrule(lr){9-11} \cmidrule(lr){12-13}\cmidrule(lr){14-14}\cmidrule(lr){15-15}
& & $do(b)$ & $do(v)$ & $do(s)$ 
 & $do(b)$ & $do(v)$ & $do(s)$ 
 & $do(b)$ & $do(v)$ & $do(s)$ 
 & MAE $\downarrow$& LPIPS$\downarrow$ & FID$\downarrow$  &CLD$\downarrow$  \\
\midrule
\rownum &VAE   
& 0.17$_{0.03}$ & 0.15$_{0.06}$ & 0.15$_{0.06}$ 
& 0.08$_{0.05}$ & 0.20$_{0.04}$ & 0.08$_{0.05}$ 
& 0.52$_{0.15}$ & 0.48$_{0.15}$ & 0.46$_{0.10}$ 
& 18.88 & 0.306 & 278.245 & 0.352 \\
\rownum &HVAE  
& 0.09$_{0.03}$ & 0.12$_{0.06}$ & 0.13$_{0.06}$ 
& 0.06$_{0.04}$ & 0.04$_{0.01}$ & 0.06$_{0.04}$ 
& 0.38$_{0.15}$ & 0.41$_{0.16}$ & 0.41$_{0.11}$ 
& \textbf{3.38} & 0.101 & 74.696 & 0.347 \\
\rownum &GAN   
& 0.17$_{0.02}$ & 0.16$_{0.07}$ & 0.16$_{0.06}$ 
& 0.12$_{0.02}$ & 0.22$_{0.03}$ & 0.12$_{0.03}$ 
& 0.14$_{0.03}$ & 0.16$_{0.03}$ & 0.05$_{0.02}$ 
& 24.26 & 0.268 & 113.749 & 0.353 \\
\midrule
\rownum &Ours  
& \textbf{0.09$_{0.01}$} & \textbf{0.11$_{0.03}$} & \textbf{0.11$_{0.03}$} 
& \textbf{0.03$_{0.01}$} & \textbf{0.03$_{0.01}$} & \textbf{0.03$_{0.01}$} 
& 0.53$_{0.09}$ & 0.55$_{0.09}$ & \textbf{0.48$_{0.06}$} 
& 3.54 & \textbf{0.035} & 9.130 & 0.346 \\
\midrule
        \multicolumn{15}{l}{\emph{with attention guidance for localized editing}} \\
\rownum &Ours  
& 0.10$_{0.01}$ & 0.14$_{0.04}$ & 0.14$_{0.06}$ 
& 0.10$_{0.02}$ & 0.04$_{0.01}$ & 0.10$_{0.02}$ 
& \textbf{0.55}$_{0.08}$ & \textbf{0.57}$_{0.08}$ & 0.46$_{0.08}$ 
& - & - & \textbf{9.066} & \textbf{0.332} \\
\bottomrule
\end{tabular}
}
\label{tab:effectiveness_benchmark_adni_full}
\end{table*}

% On the ADNI dataset, the 10-cycle composition results in Table~\ref{tab:celea_composition_10} show that our method achieves competitive reconstruction MAE and LPIPS. 
% For counterfactual generation, however, iterative reconstruction is not the primary requirement; it can nevertheless be strengthened by increasing the number of diffusion steps. 
% Additional qualitative results are provided in Figure~\ref{fig:adni_extra_quality_1}--\ref{fig:adni_extra_quality_2}.

% \begin{table}[h]
% \centering
% \caption{Composition$^{10}$ results on CelebA test set.} 
% \resizebox{0.4\linewidth}{!}{ 
%     \begin{tabular}{c c c}
%     \toprule
%     \multirow{2}{*}{Method} & \multicolumn{2}{c}{\textbf{Composition}$^{10}$} \\ 
%     \cmidrule(lr){2-3} 
%      & MAE \(\downarrow\) & LPIPS \(\downarrow\) \\
%     \midrule
%     VAE  & 30.250 & 0.384 \\
%     HVAE & \textbf{7.456} & \textbf{0.156} \\
%     GAN  & 32.794 & 0.323 \\
%     $\text{Ours}$   & 23.400 & 0.193 \\
%     \bottomrule
%     \end{tabular}
% }
% \label{tab:adni_composition_10}
% \end{table}

\clearpage
\subsection{CelebA-HQ}
\label{Appendix:extra_celebahq}
We compare Causal-Adapter with recent state-of-the-art counterfactual methods including VCI~\citep{wu2024counterfactual}, HVAE~\citep{pmlr-v202-de-sousa-ribeiro23a}, and DiffCounter~\citep{rasal2025diffusion} on high-resolution human face dataset CelebA-HQ. We follow the same settings in DiffCounter for fair comparison, focusing on three categorical variables (glasses, smile, mouth-open). The quantitative results are presented in Table~\ref{tab:celeba-hq_full}, with an additional reversibility metric that measures how well the generated counterfactuals can be recovered back to the original observations. Our Causal-Adapter (row 4) achieves intervention effectiveness comparable to DiffCounter, while substantially improving both reversibility and identity preservation. For example, under the smile intervention, our method reduces LPIPS by 57\% (from 0.66 to 0.028), and for glasses reversal, it reduces $L_1$ by 73\% (from 0.185 to 0.049). The reversibility analysis empirically indicates that our model can recover counterfactuals back to their original observations (Appendix~\ref{Appendix:Counterfactual Identifiability}), suggesting potential for achieving counterfactual identifiability~\citep{ribeiro2025counterfactual}.
\setcounter{rownumbers}{0}
\begin{table*}[ht]
    \centering
    \footnotesize
    \caption{Soundness of CelebA-HQ counterfactual images generated by the proposed Causal-Adapter. Effectiveness is evaluated using F1-scores from pre-trained classifiers for eyeglasses ($g$) and smiling ($s$), while reversibility (Rev.) and compositional consistency (Comp.) are measured using $L_1$. Identity preservation (IDP) is assessed using LPIPS.}
    \label{tab:celeba-hq_full}

    \resizebox{\textwidth}{!}{
    \begin{tabular}{r l cc c c cc c c c}
        \toprule
        && \multicolumn{4}{c}{\textsc{Eyeglasses Intervention $(do(g))$}}
        & \multicolumn{4}{c}{\textsc{Smiling Intervention $(do(s))$}}
        & \textsc{Null} \\
        \cmidrule(lr){3-6} \cmidrule(lr){7-10}
        && \multicolumn{2}{c}{\textsc{Effectiveness}}
        & \textsc{Rev.} & \textsc{IDP}
        & \multicolumn{2}{c}{\textsc{Effectiveness}}
        & \textsc{Rev.} & \textsc{IDP}
        & \textsc{Comp.} \\
        \cmidrule(lr){3-4} \cmidrule(lr){5-5} \cmidrule(lr){6-6}
        \cmidrule(lr){7-8} \cmidrule(lr){9-9} \cmidrule(lr){10-10}\cmidrule(lr){11-11}

        &\textsc{Method}
        & $\text{F1}(s)\uparrow$ & $\text{F1}(g)\uparrow$
        & $L_1\downarrow$ & $\text{LPIPS}\downarrow$
        & $\text{F1}(s)\uparrow$ & $\text{F1}(g)\uparrow$
        & $L_1\downarrow$ & $\text{LPIPS}\downarrow$
        & $L_1\downarrow$ \\
        \midrule

        \rownum &VCI 
        & 97.84 & 3.39 & - & - 
        & 33.81 & \textbf{99.85} & - & - & - \\

        \rownum &HVAE 
        & 90.05 & 65.31 & - & - 
        & 75.33 & 95.82 & - & - & - \\

        \rownum &DiffCounter 
        & \textbf{99.09} & 96.86 & 0.185 & 0.096
        & \textbf{94.93} & 99.45 & 0.183 & 0.066 & 0.130 \\

        \rownum &Ours 
        & 96.89 & \textbf{99.26} & \textbf{0.049} & 0.084
        & 94.15 & 99.15 & \textbf{0.028} & \textbf{0.028} & 0.010 \\
        \midrule

        \rownum &Ours (DiT) 
        & 98.19 & 97.39 & 0.086 & \textbf{0.060}
        & 94.71 & 99.71 & 0.089 & 0.035 & \textbf{0.001} \\

        \bottomrule
    \end{tabular}
    }
\end{table*}

\begin{figure}[!ht]
    \centering
    \includegraphics[width=0.93\linewidth]{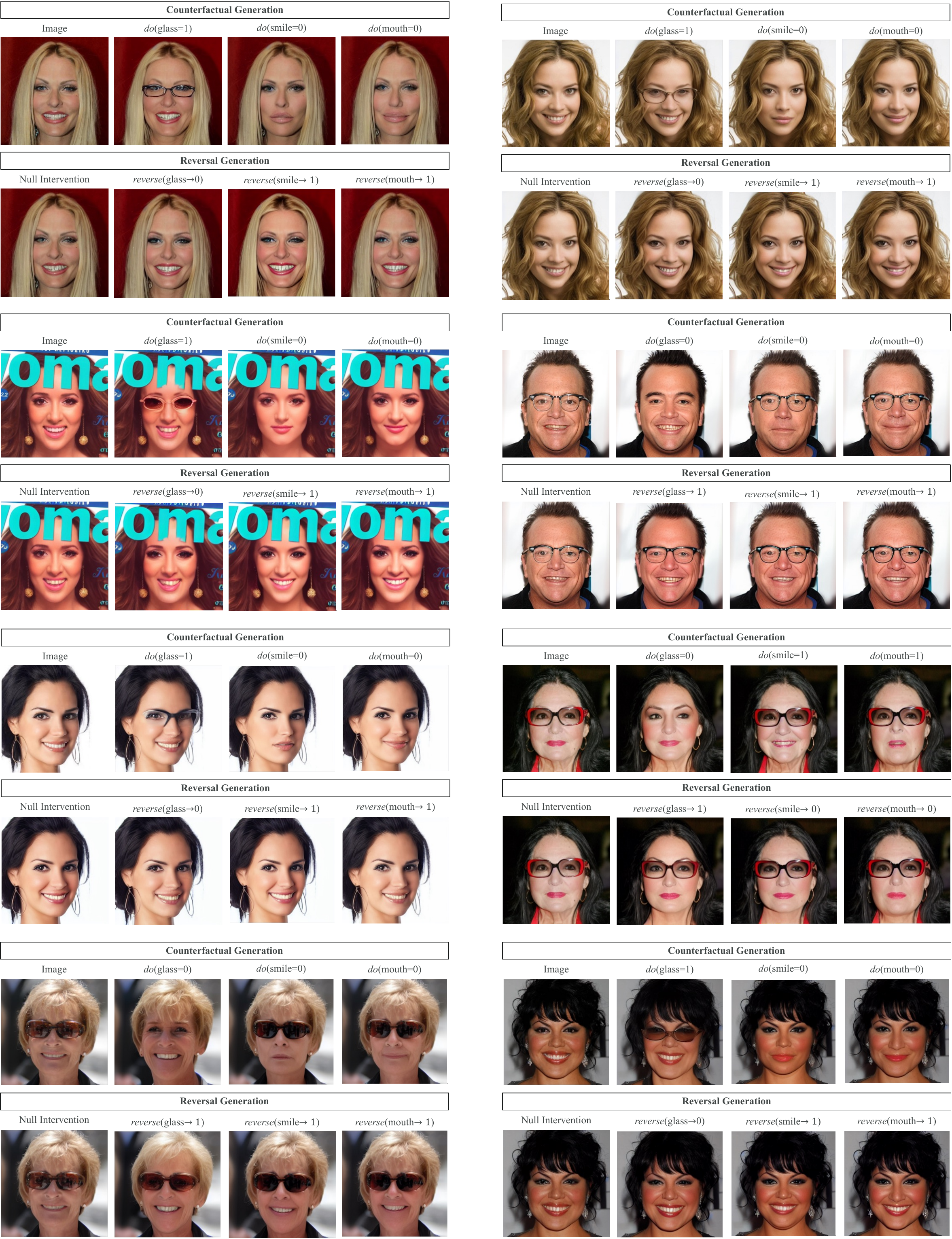}
    \caption{Additional counterfactual and reversal results (256$\times$256) on CelebA-HQ.}
\label{fig:celebahq_extra_reversal_1}
\end{figure}

\begin{figure}[!ht]
    \centering
    \includegraphics[width=0.93\linewidth]{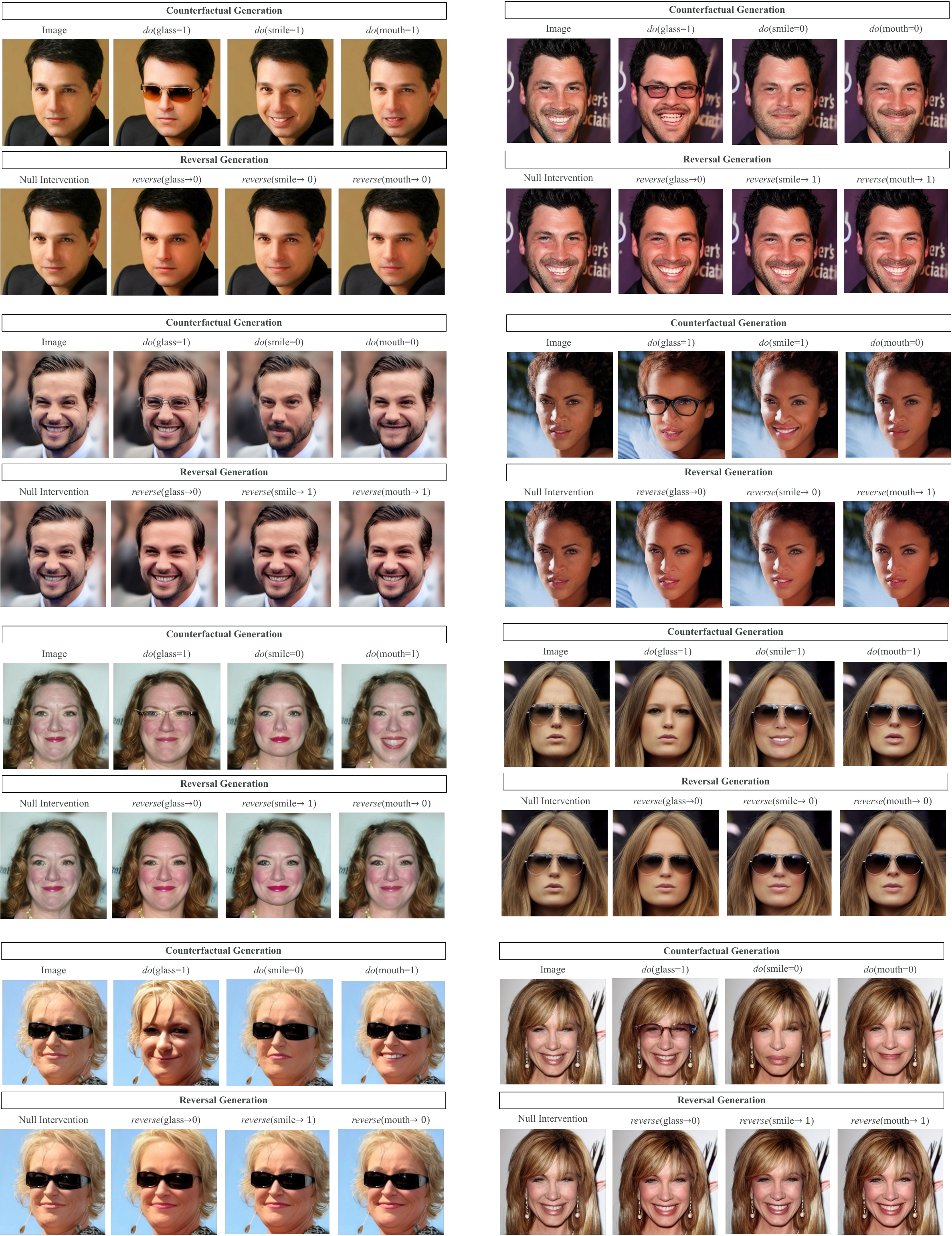}
    \caption{Additional counterfactual and reversal results (256$\times$256) on CelebA-HQ.}
\label{fig:celebahq_extra_reversal_2}
\end{figure}

\paragraph{Additional counterfactual and reversal results.}  
Supplementary counterfactual and reversal examples generated by Causal-Adapter are provided in Figure~\ref{fig:celebahq_extra_reversal_1}–\ref{fig:celebahq_extra_reversal_2}. These visualizations demonstrate faithful interventions and strong identity preservation across diverse attribute edits.

\begin{figure}[!ht]
    \centering
    \includegraphics[width=0.83\linewidth]{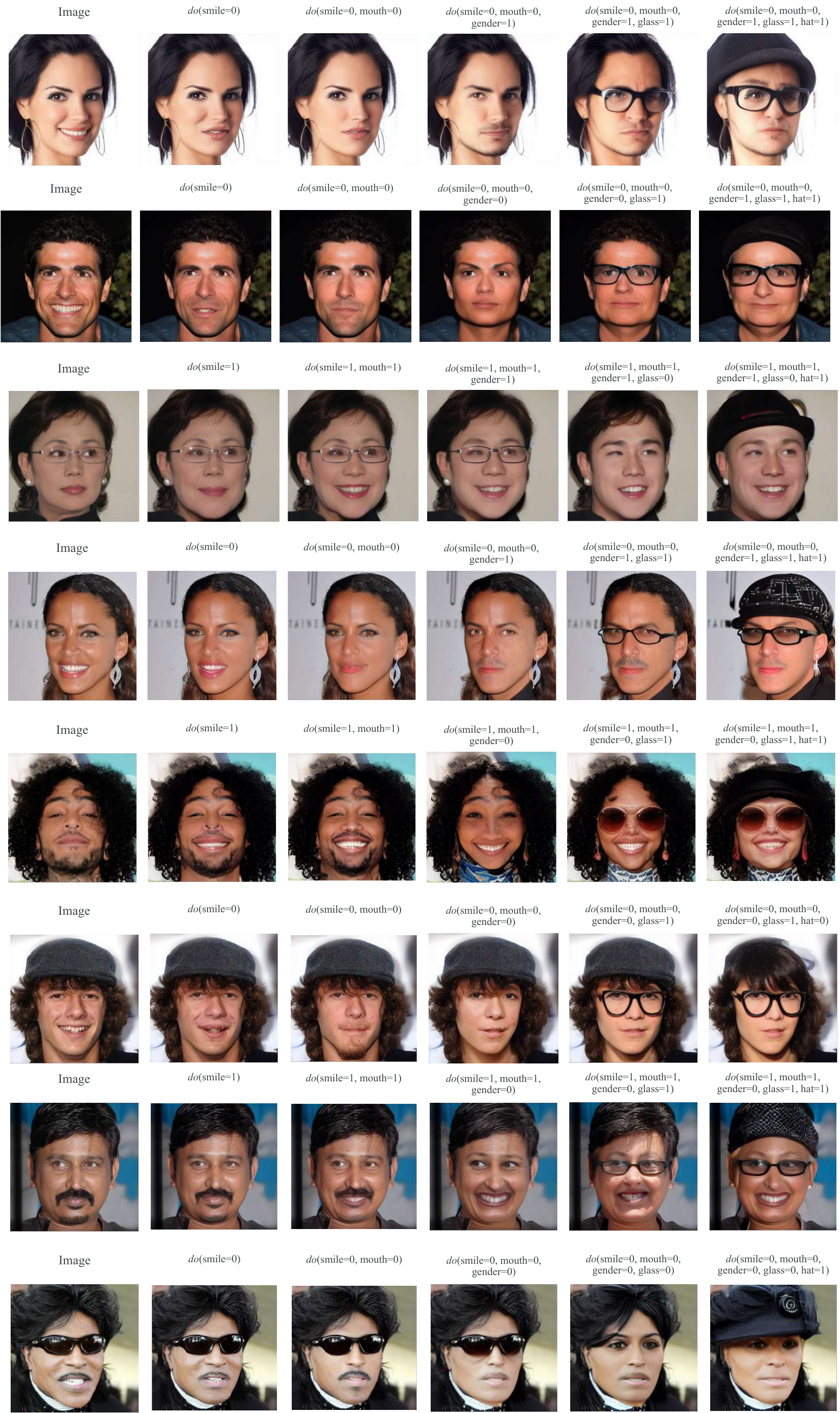}
    \caption{Stress test with compositional interventions: five attributes (smile, mouth-open, gender, glasses, hat) are progressively intervened. For example, the second column applies $do(\text{smile})$; the third column applies $do(\text{smile}, \text{mouth})$; subsequent columns add further interventions in sequence.}
\label{fig:celebahq_extra_compos_1}
\end{figure}

\begin{figure}[!ht]
    \centering
    \includegraphics[width=0.83\linewidth]{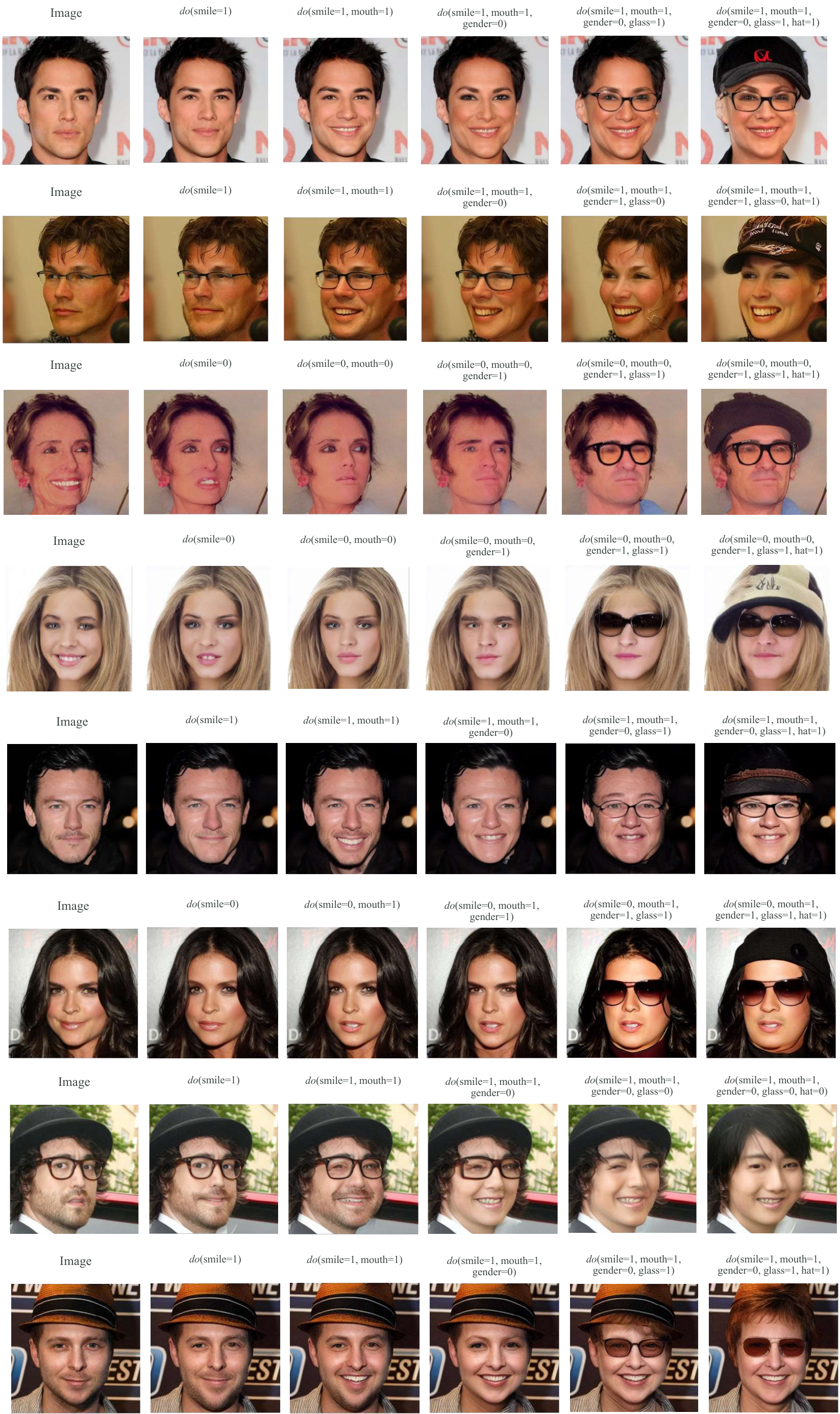}
    \caption{Stress test with compositional interventions: five attributes (smile, mouth-open, gender, glasses, hat) are progressively intervened. For example, the second column applies $do(\text{smile})$; the third column applies $do(\text{smile}, \text{mouth})$; subsequent columns add further interventions in sequence.}
\label{fig:celebahq_extra_compos_2}
\end{figure}

\paragraph{Stress test with compositional interventions.}  
Following the settings of~\citet{rasal2025diffusion}, we concatenated four confounding attributes (Male, Wearing Lipstick, Bald, Wearing Hat) during training to mitigate cross-attribute spurious correlations. Based on the disentangled attributes, we perform a compositional stress test in which we select five non-overlapping attributes (smile, mouth-open, gender, glasses, hat) and progressively apply multiple interventions, i.e., $do(\text{smile})$, $do(\text{smile}, \text{mouth})$, etc. Results in Figure~\ref{fig:celebahq_extra_compos_1}–\ref{fig:celebahq_extra_compos_2} show that Causal-Adapter produces plausible counterfactuals under progressively complex intervention sets while maintaining strong identity consistency.

\paragraph{Generalization test (SD3 backbone).}
To further evaluate the generalizability of Causal-Adapter, we apply it to Stable Diffusion 3 (SD3)~\citep{esser2024scaling}, a diffusion backbone based on the Diffusion Transformer (DiT) architecture. 
Unlike SD~1.5, SD3 employs three text encoders, including two CLIP encoders and one T5 encoder. 
Accordingly, PAI injects attribute information into all three token-embedding streams. 
For a dataset with $N$ attributes, we learn $N$ attribute tokens for each encoder, resulting in $3N$ injected embeddings that guide the SD3 MMDiT backbone through the adapter. 

Following the SD3 training recipe, we replace the standard diffusion loss with the flow-matching loss $\mathcal{L}_{\text{FM}}$. 
To maintain semantic alignment across the three heterogeneous embedding spaces, we apply CTC to the injected tokens of each encoder and optimize the following objective:
\begin{equation}
\mathcal{L}_{\text{total}} = \mathcal{L}_{\text{FM}} + \lambda \sum_{i=1}^{3} \mathcal{L}_{\text{CTC}_i},
\end{equation}
where $\mathcal{L}_{\text{CTC}_i}$ denotes the contrastive token loss for the injected embeddings of the $i$-th encoder.

During inference, we adopt FlowEdit~\citep{kulikov2025flowedit}, an editing method designed for flow-matching models. 
We construct the source condition by injecting the original attributes of the input image into the text embeddings, and construct the target condition by injecting the SCM-intervened attributes. 
We use 50 inversion and denoising steps with a target guidance scale of 1.5 to generate counterfactual images.

\begin{figure}[!ht]
    \centering
    \includegraphics[width=\linewidth]{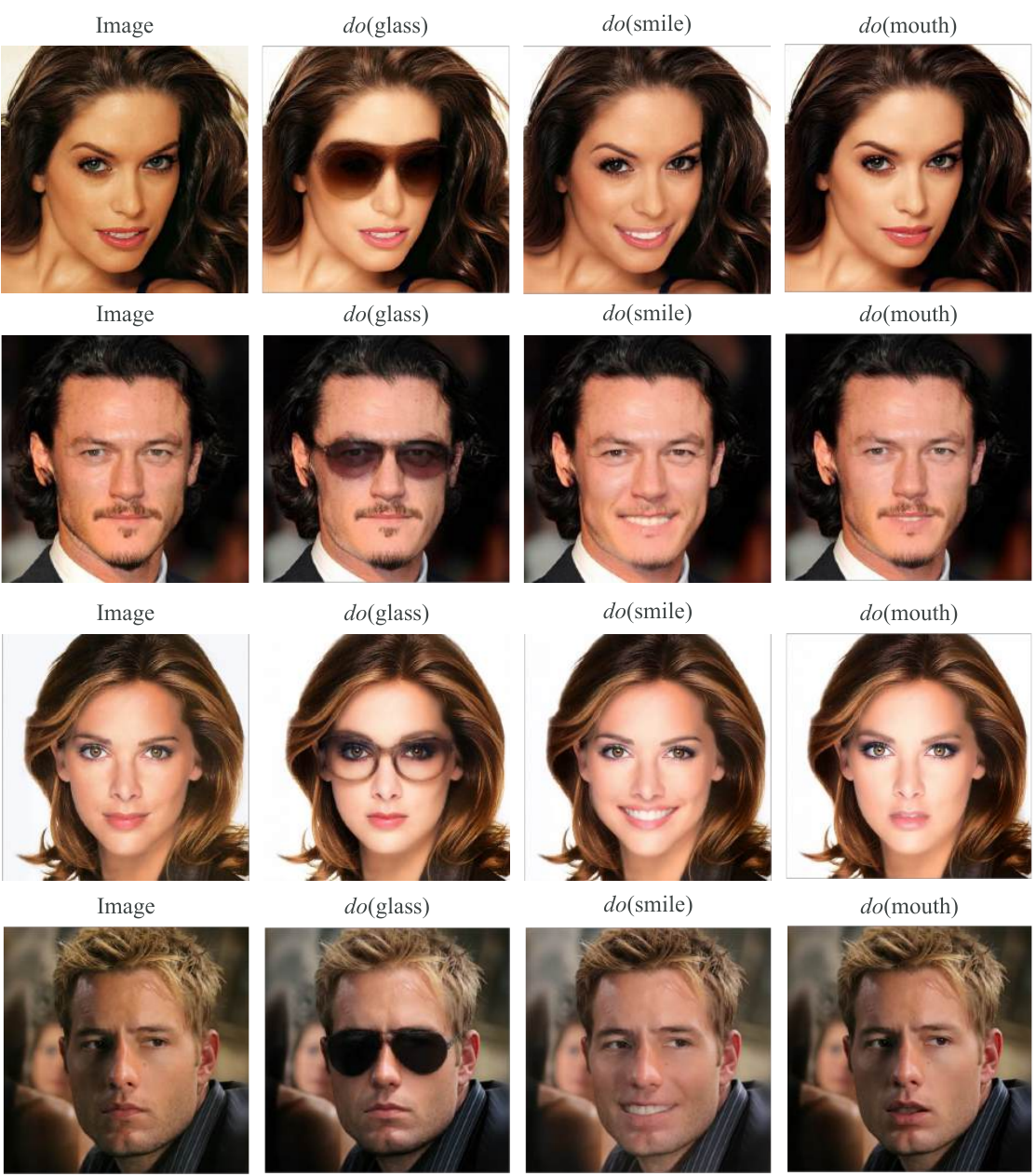}
    \caption{Additional counterfactuals (512$\times$512) generated by using SD3 Backbone.}
\label{fig:celebahq_extra_sd3_2}
\end{figure}

\begin{figure}[h]
    \centering
    \includegraphics[width=\linewidth]{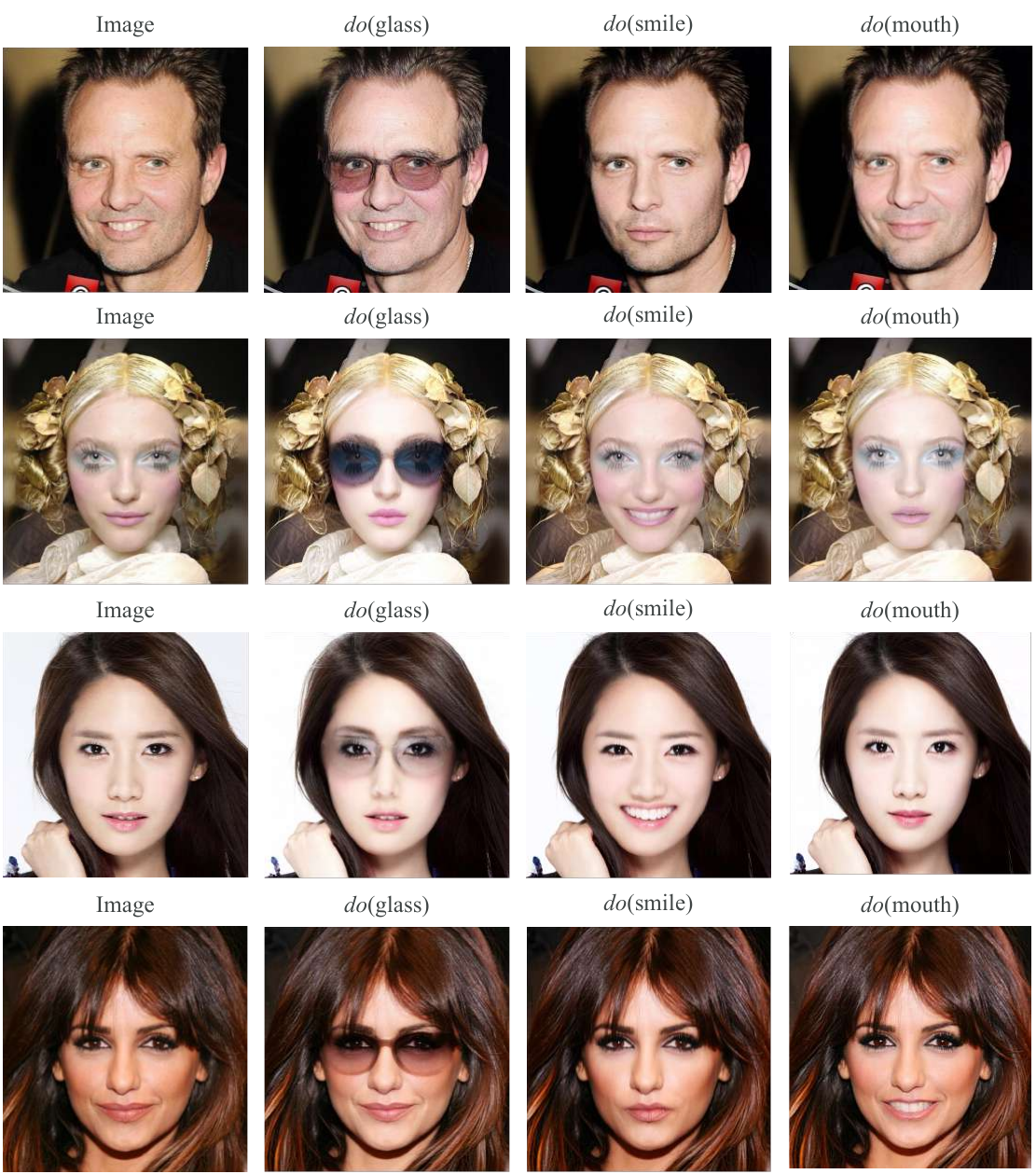}
    \caption{Additional counterfactuals (512$\times$512) generated by using SD3 Backbone.}
\label{fig:celebahq_extra_sd3_3}
\end{figure}

\begin{figure}[h]
    \centering
    \includegraphics[width=\linewidth]{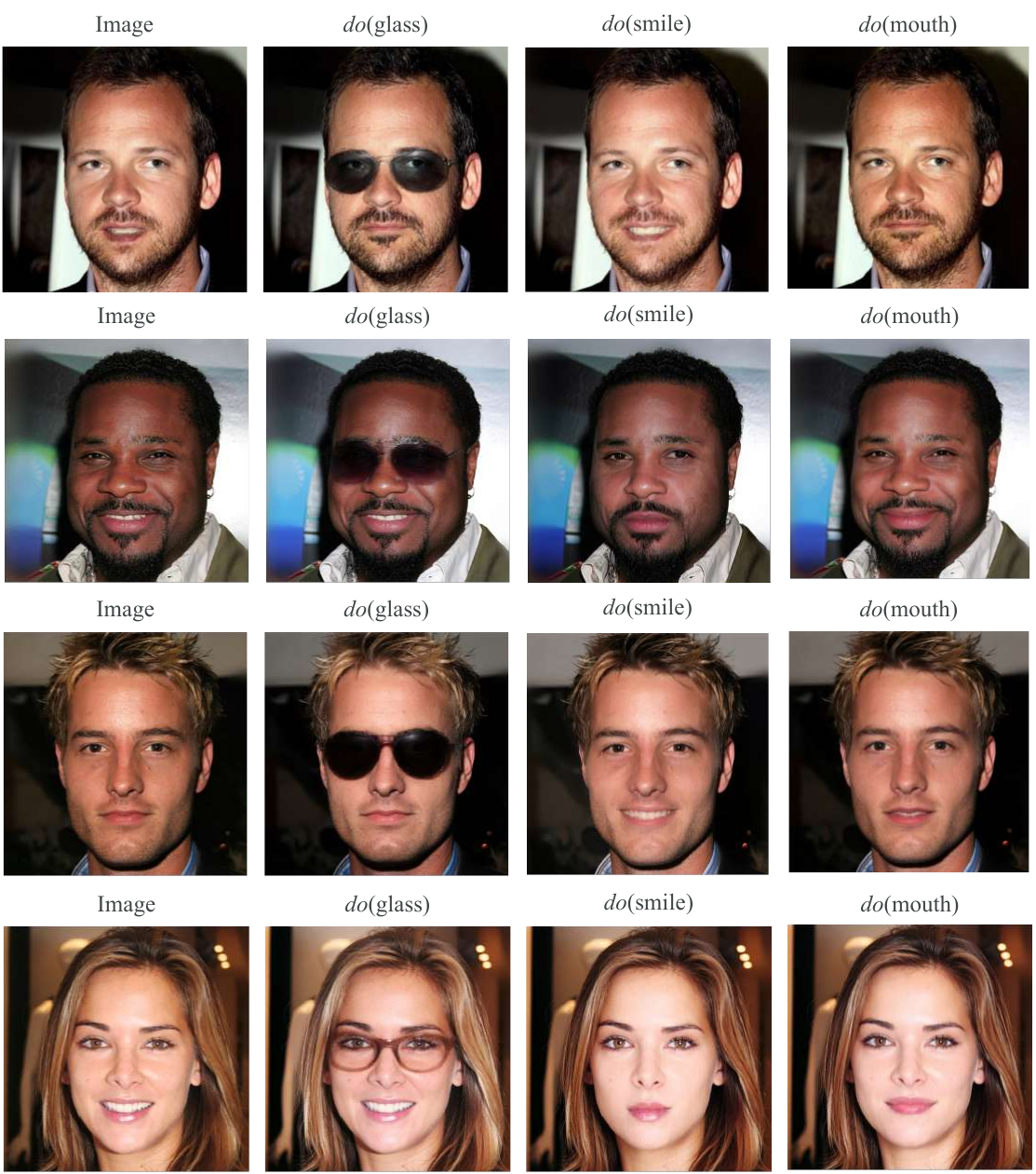}
    \caption{Additional counterfactuals (512$\times$512) generated by using SD3 Backbone.}
\label{fig:celebahq_extra_sd3_4}
\end{figure}

As shown in Figure~\ref{fig:celebahq_extra_sd3_2}--\ref{fig:celebahq_extra_sd3_4}, Causal-Adapter successfully adapts to the SD3 backbone and produces causally faithful counterfactuals. 
These results demonstrate that our method is not tied to DDIM inversion or a specific diffusion architecture, but can naturally extend to different T2I generative families and editing paradigms. 
Although other inversion techniques, such as null-text inversion~\citep{mokady2023null}, can also be integrated, they require per-sample optimization and are therefore less suitable for large-scale benchmarks.

\clearpage
\subsubsection{Qualitative Results with the SD1.5 Backbone}
\label{Appendix:celebahq_sd15}
\clearpage
\subsubsection{Qualitative Results with the SD3 Backbone}
\label{Appendix:celebahq_sd3}
% \begin{figure}[h]
%     \centering
%     \includegraphics[width=\linewidth]{rebuttal/CelebaHQ-DIT-Counterfactuals/sd3_qualitative1.pdf}
%     \caption{Additional counterfactuals (512$\times$512) generated by using SD3 Backbone.}
% \label{fig:celebahq_extra_sd3_1}
% \end{figure}

\clearpage
\subsection{Attention Maps}
\label{Appendix:additional_attention_maps}
 % we present additional attention maps obtained from the fully regularized Causal-Adapter. These maps are collected from the cross-attention layers of the frozen diffusion denoiser $\epsilon$ and our adapter encoder $\ddot{\epsilon}_{\psi}$.

\begin{figure}[!htbp]
    \centering
    \includegraphics[width=0.87\linewidth]{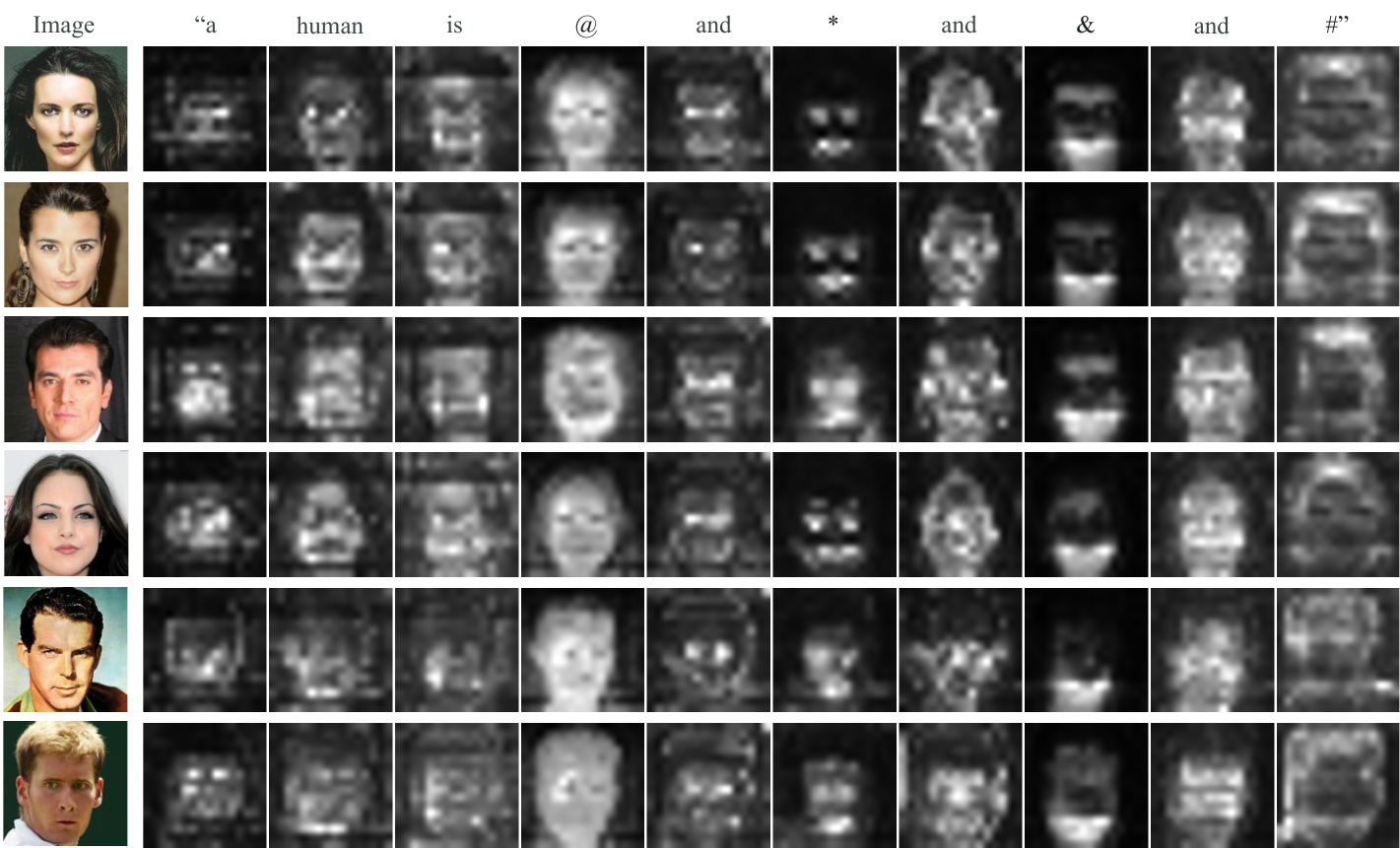}
    \caption{Average cross-attention maps from Causal-Adapter on CelebA dataset. Token denote attributes: ``@'' for age, ``*'' for gender, ``\&'' for beard, and ``\#'' for bald.}
    \label{fig:celeba_attention_maps}
\end{figure}

\begin{figure}[!htbp]
    \centering
    \includegraphics[width=0.87\linewidth]{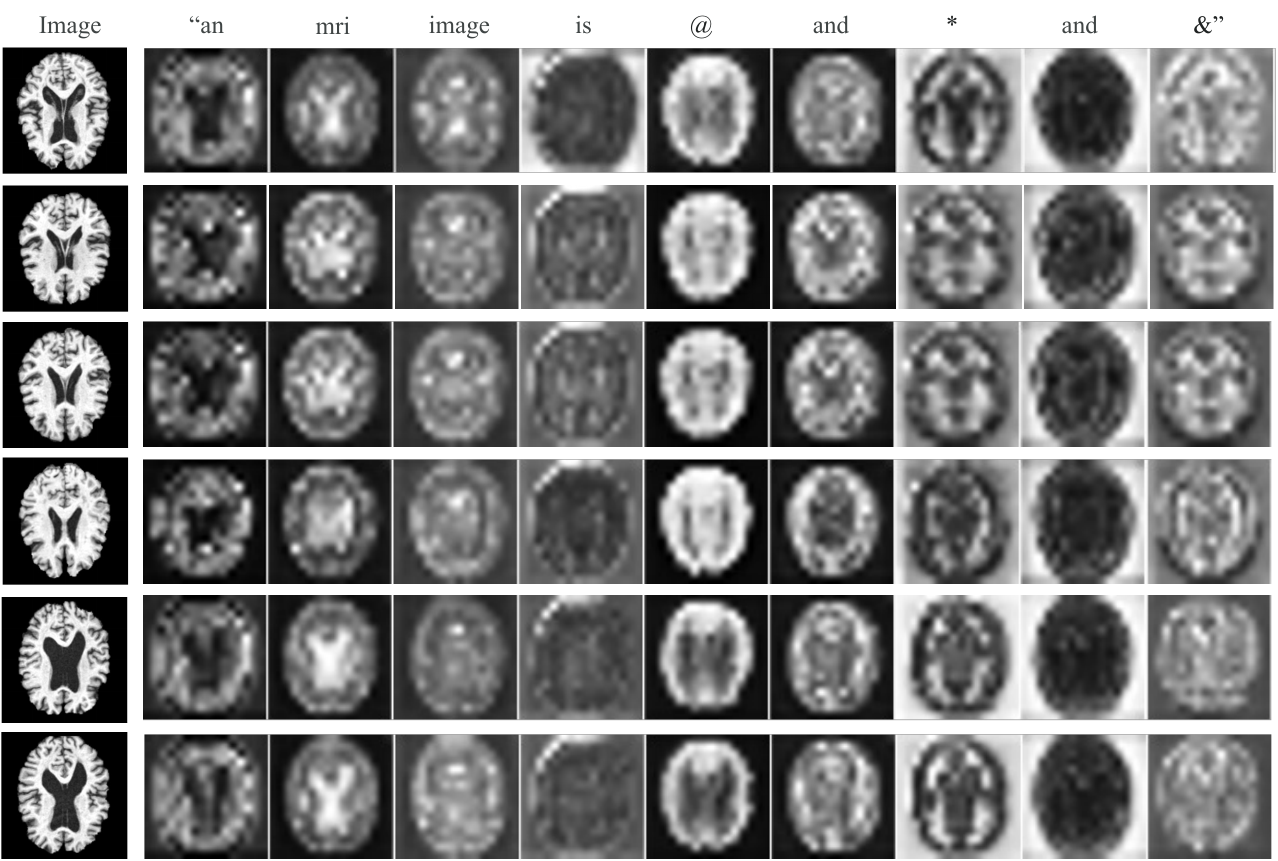}
    \caption{Average cross-attention maps from Causal-Adapter on ADNI dataset. Token denote attributes: ``@'' for brain volume, ``*'' for ventricular volume, ``\&'' for slice.}
    \label{fig:adni_attention_maps}
\end{figure}

\begin{figure}[!htbp]
    \centering
    \includegraphics[width=0.87\linewidth]{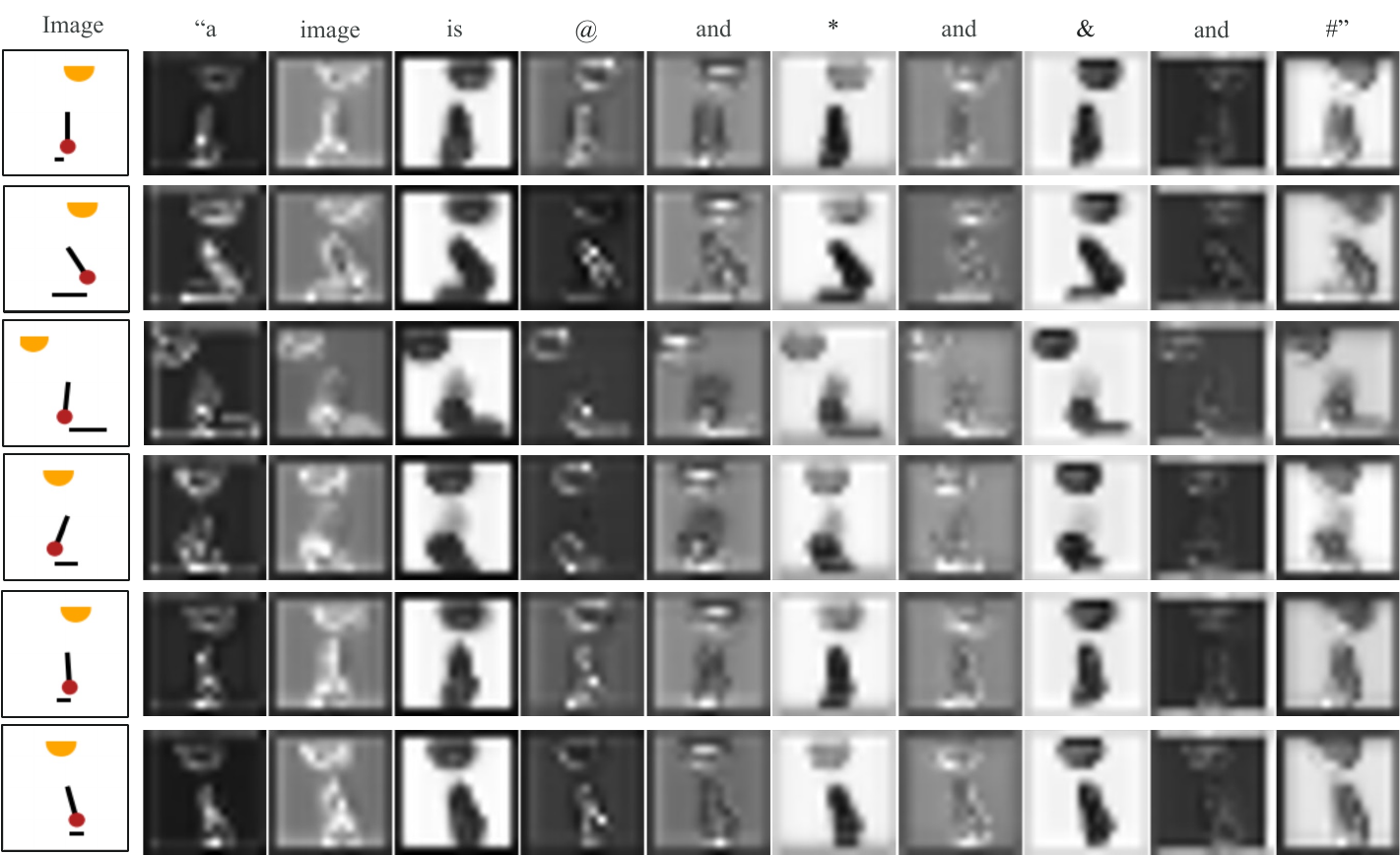}
    \caption{Average cross-attention maps from Causal-Adapter on Pendulum dataset. Token denote attributes: ``@'' for pendulum, ``*'' for light, ``\&'' for shadow length, and ``\#'' for shadow position.}
    \label{fig:pend_attention_maps}
\end{figure}

\clearpage
\subsection{Comparison with Commercial Foundation Models}
\label{Appendix:compare_commercial_model}

\begin{figure}[!htbp]
    \centering
    \includegraphics[width=1\linewidth]{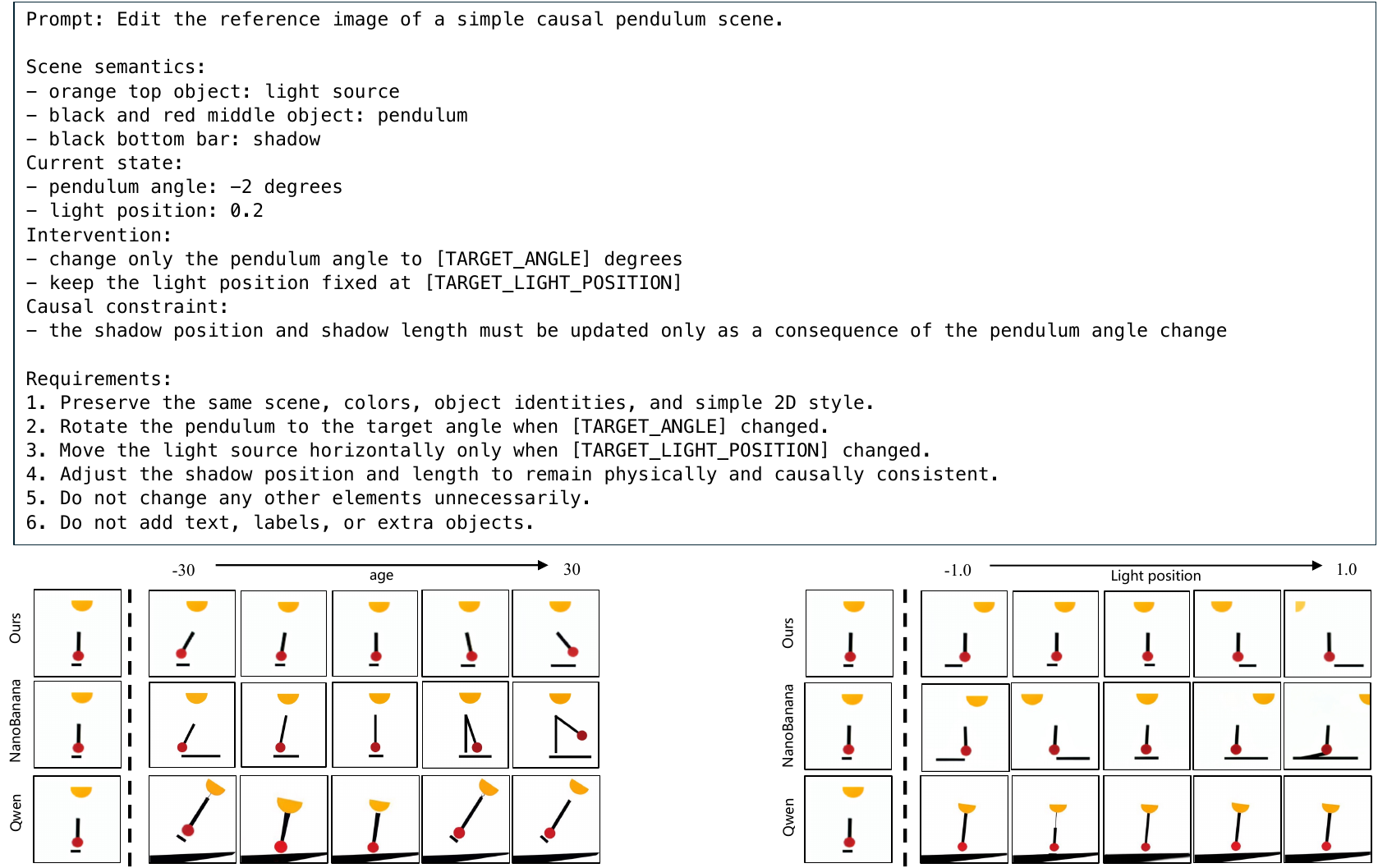}
    \caption{
    Qualitative comparison with commercial foundation editing models on the Pendulum dataset.
    }
    \label{fig:foundation_model_comparison_pendulum}
\end{figure}

\begin{figure}[!htbp]
    \centering
    \includegraphics[width=1\linewidth]{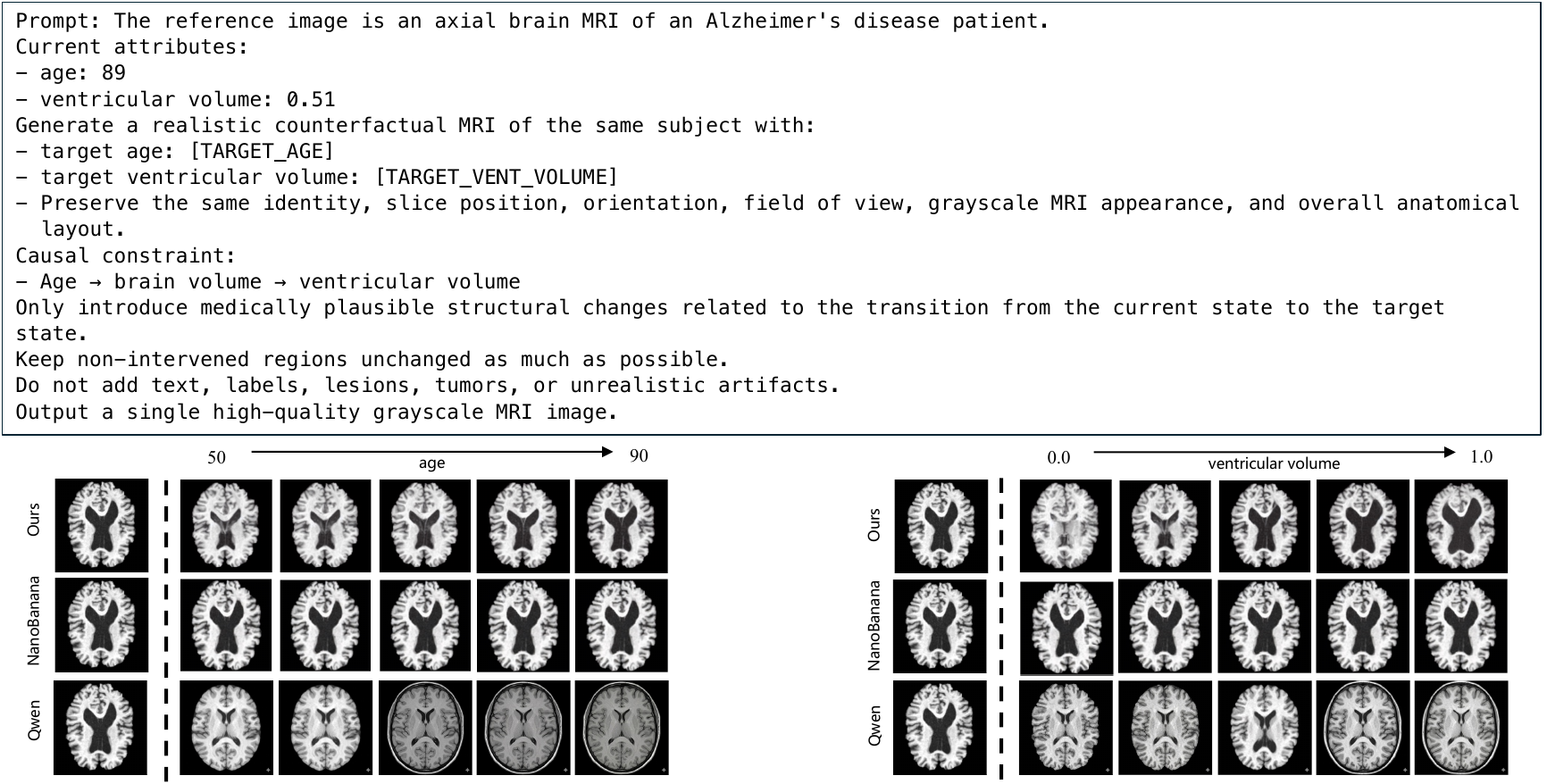}
    \caption{
    Qualitative comparison with commercial foundation editing models on ADNI.
    }
    \label{fig:foundation_model_comparison_adni}
\end{figure}

\begin{figure}[!htbp]
    \centering
    \includegraphics[width=0.87\linewidth]{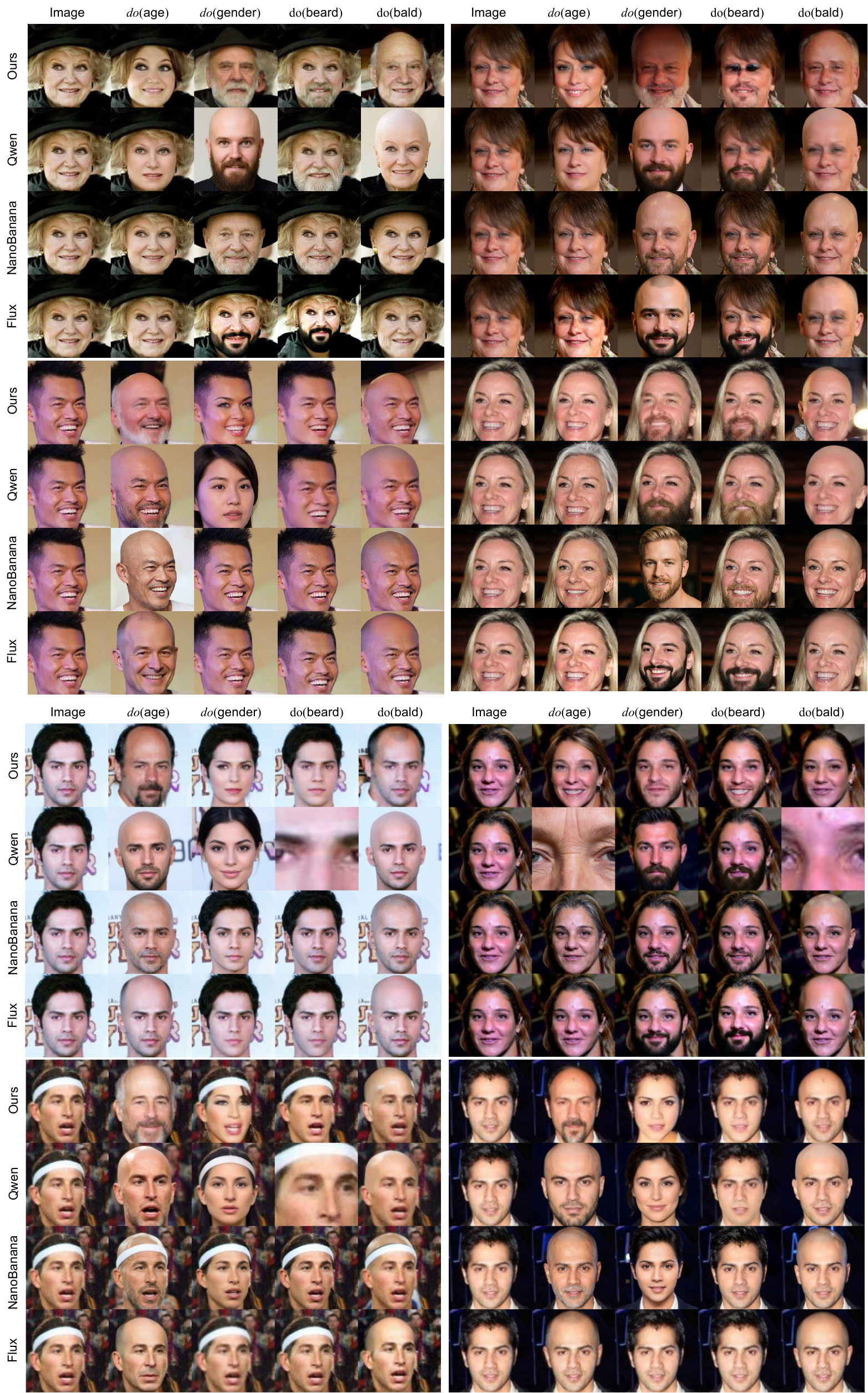}
    \caption{
    Qualitative comparison with commercial foundation editing models on CelebA.
    }
    \label{fig:foundation_model_comparison_celeba1}
\end{figure}

To further assess the practical effectiveness of Causal-Adapter, we compare it with recent commercial foundation editing models, including Nano Banana 2, Qwen-Image-Edit, and Flux.1-Kontext-Dev. 
We provide qualitative comparisons on Pendulum, ADNI, and CelebA in Figures~\ref{fig:foundation_model_comparison_pendulum}--\ref{fig:foundation_model_comparison_celeba1}. 
Since these foundation models do not expose explicit attribute-level causal controls, editing is performed by updating their text prompts. 
In contrast, Causal-Adapter performs counterfactual editing by intervening on attributes through the SCM and injecting the resulting causal conditions into the text-embedding space.

We also provide quantitative comparisons on a CelebA subset. 
Tables~\ref{tab:commercial_age_gender} and~\ref{tab:commercial_beard_bald} report classifier-based accuracy for the target and related attributes, together with LPIPS for visual similarity. 
Compared with large-scale commercial editing models trained with complex pipelines, Causal-Adapter remains simple and efficient while achieving competitive performance under interventions on age, gender, and beard. 
These results are consistent with our main findings: recent foundation editing models can produce visually plausible edits, but they still struggle with fine-grained causal interventions, especially for continuous or causally entangled attributes. 
They often modify visually salient regions rather than faithfully following the intended causal intervention, which limits their suitability for counterfactual image generation.

\setcounter{rownumbers}{0}
\begin{table}[ht!]
\centering
\caption{
Quantitative comparison with commercial foundation editing models on CelebA under age and gender interventions. 
We report classifier accuracy for Age, Gender, Beard, and Bald, together with LPIPS for visual similarity. 
Higher accuracy is better, while lower LPIPS is better.
}
\label{tab:commercial_age_gender}
\scriptsize
\resizebox{\textwidth}{!}{%
\begin{tabular}{rlcccccccccc}
\toprule
&\multirow{2}{*}{Method}
& \multicolumn{2}{c}{\textbf{Age Acc. $\uparrow$}}
& \multicolumn{2}{c}{\textbf{Gender Acc. $\uparrow$}}
& \multicolumn{2}{c}{\textbf{Beard Acc. $\uparrow$}}
& \multicolumn{2}{c}{\textbf{Bald Acc. $\uparrow$}}
& \multicolumn{2}{c}{\textbf{LPIPS $\downarrow$}} \\
\cmidrule(lr){3-4}
\cmidrule(lr){5-6}
\cmidrule(lr){7-8}
\cmidrule(lr){9-10}
\cmidrule(lr){11-12}
&& $do(a)$ & $do(g)$
& $do(a)$ & $do(g)$
& $do(a)$ & $do(g)$
& $do(a)$ & $do(g)$
& $do(a)$ & $do(g)$ \\
\midrule

\rownum &Ours
& \textbf{72} & 90
& \textbf{100} & \textbf{100}
& \textbf{100} & 96
& 90 & 92
& 0.0725 & 0.0526 \\

\rownum &Qwen-Image-Edit
& 62 & 86
& 84 & 90
& 96 & 96
& 86 & \textbf{98}
& 0.1496 & 0.1579 \\

\rownum &Flux.1-Kontext-Dev
& 34 & 84
& 98 & 46
& 96 & 88
& 92 & 90
& \textbf{0.0560} & \textbf{0.0480} \\

\rownum &Nano Banana 2
& 44 & \textbf{94}
& 98 & 68
& 94 & \textbf{100}
& \textbf{92} & \textbf{98}
& 0.0750 & 0.0820 \\

\bottomrule
\end{tabular}
}
\end{table}

\setcounter{rownumbers}{0}
\begin{table}[ht!]
\centering
\caption{
Quantitative comparison with commercial foundation editing models on CelebA under beard and bald interventions. 
We report classifier accuracy for Age, Gender, Beard, and Bald, together with LPIPS for visual similarity. 
Higher accuracy is better, while lower LPIPS is better.
}
\label{tab:commercial_beard_bald}
\scriptsize
\resizebox{\textwidth}{!}{%
\begin{tabular}{rlcccccccccc}
\toprule
&\multirow{2}{*}{Method}
& \multicolumn{2}{c}{\textbf{Age Acc. $\uparrow$}}
& \multicolumn{2}{c}{\textbf{Gender Acc. $\uparrow$}}
& \multicolumn{2}{c}{\textbf{Beard Acc. $\uparrow$}}
& \multicolumn{2}{c}{\textbf{Bald Acc. $\uparrow$}}
& \multicolumn{2}{c}{\textbf{LPIPS $\downarrow$}} \\
\cmidrule(lr){3-4}
\cmidrule(lr){5-6}
\cmidrule(lr){7-8}
\cmidrule(lr){9-10}
\cmidrule(lr){11-12}
&& $do(br)$ & $do(bl)$
& $do(br)$ & $do(bl)$
& $do(br)$ & $do(bl)$
& $do(br)$ & $do(bl)$
& $do(br)$ & $do(bl)$ \\
\midrule

\rownum &Ours
& \textbf{98} & \textbf{70}
& 48 & \textbf{86}
& 80 & 92
& \textbf{100} & 40
& 0.0282 & 0.1087 \\

\rownum &Qwen-Image-Edit
& 84 & 56
& 36 & 82
& \textbf{92} & \textbf{96}
& 98 & \textbf{94}
& 0.1326 & 0.1892 \\

\rownum &Flux.1-Kontext-Dev
& 78 & 60
& 44 & 80
& 88 & 92
& 98 & 76
& 0.0665 & \textbf{0.0976} \\

\rownum &Nano Banana 2
& 92 & 66
& \textbf{50} & \textbf{86}
& 86 & \textbf{96}
& \textbf{100} & \textbf{94}
& \textbf{0.0215} & 0.1200 \\

\bottomrule
\end{tabular}
}
\end{table}

\clearpage
\subsection{Generalization Beyond Aligned Images}
\label{Appendix:human_body_generalization}

To further examine whether Causal-Adapter can generalize beyond tightly aligned datasets such as CelebA, ADNI, and Pendulum, we evaluate it on full-body human images with larger variations in pose, background, scale, and composition. 
Specifically, we use a public human-body image dataset\footnote{\url{https://www.kaggle.com/datasets/snmahsa/human-images-dataset-men-and-women/data}}, randomly select 100 samples, and manually annotate three attributes: age, gender, and beard. 
Since constructing a comprehensive causal graph for open-domain human images is challenging, we perform disentangled editing without using an SCM in this experiment. 
We compare Causal-Adapter with Qwen-Image-Edit, Nano Banana 2, and Flux.1-Kontext-Dev.

\begin{figure}[!htbp]
    \centering
    \includegraphics[width=\linewidth]{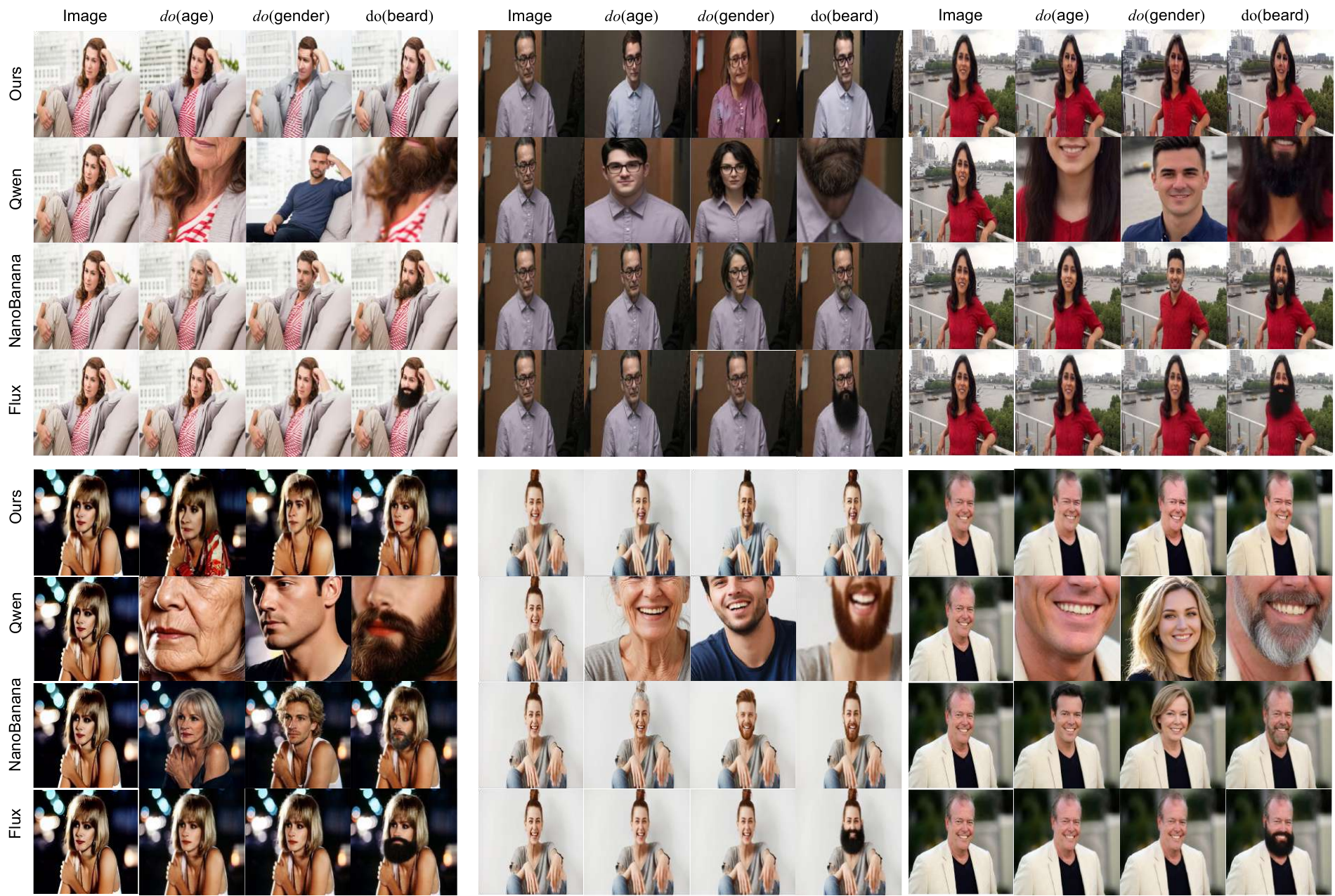}
    \caption{
    Qualitative comparison with commercial foundation editing models on full-body human images. 
    Causal-Adapter performs disentangled editing without using an explicit causal graph in this experiment.
    }
    \label{fig:human_body_generalization}
\end{figure}

As shown in Figure~\ref{fig:human_body_generalization}, commercial foundation editing models can generate visually plausible results, but they often struggle to perform precise attribute-level interventions while preserving the overall image structure. 
For example, Qwen-Image-Edit tends to over-edit the image and sometimes introduces partial cropping artifacts, while Flux.1-Kontext-Dev often fails to reliably edit age and gender. 
In contrast, Causal-Adapter produces clear changes under age and gender interventions while better preserving pose, clothing layout, and background. 
These results suggest that the proposed framework retains a degree of disentangled editing ability beyond highly aligned face datasets.

This experiment is not intended to claim fully open-domain causal generation, as reliable causal editing in general scenes would require richer attribute annotations and more comprehensive causal knowledge. 
Nevertheless, the results indicate that Causal-Adapter can potentially scale to more complex domains when sufficiently labeled attributes and training samples are available. 
They also support our main claim that the framework can still function as a disentangled editing method even without an explicit causal graph, while the SCM can be incorporated when reliable causal structure is available.

\clearpage
\section{Counterfactual Identifiability \footnotesize{\textbf{\hyperref[Appendix:table_of_contents]{[Back to Contents]}}}}
\label{Appendix:Counterfactual Identifiability}
The abduction step in counterfactual generation implicitly assumes that the frozen diffusion model provides a counterfactually valid inverse mapping that DDIM inversion can approximately recover the exogenous noise consistent with the true data-generating process. This relates to the broader challenge of counterfactual identifiability in high-dimensional generative models~\citep{ribeiro2025counterfactual,komanduri2023identifiable,nasr2023counterfactual}, where recovering latent exogenous variables from observations is generally non-trivial and often non-identifiable without additional assumptions.

To empirically assess identifiability, we perform a reversibility analysis, following the principle that counterfactual outputs should be reconstructable from the observed image distribution.  
In our CelebA-HQ experiments, we compare Causal-Adapter with DiffCounter and observe substantially stronger reversibility under our approach.  
Representative examples for three interventions (glasses, smile, mouth) are shown in Figure~\ref{fig:celebahq_extra_reversal_1}–\ref{fig:celebahq_extra_reversal_2}.  
These results suggest that our adapter improves the model’s ability to produce counterfactuals that remain consistent with the underlying observational manifold.

However, we note that strong empirical recovery does not imply a formal identifiability guarantee.  
DDIM inversion is approximate, and imperfect reconstruction may cause information loss, especially under complex or compound interventions.  
Thus, even our Causal-Adapter “tames’’ causal priors into the frozen T2I backbone and yields practically robust counterfactual behavior, exact theoretical identifiability remains an open challenge.

Achieving formal counterfactual identifiability would require improved inverse operators or diffusion models explicitly designed to recover exogenous noise, may potentially leveraging recent progress in flow matching.  
We regard this as an important direction for future work, complementary to our empirical findings.

\clearpage

\clearpage
\section{Relaxing the Assumption of a Known Causal Graph 
\footnotesize{\textbf{\hyperref[Appendix:table_of_contents]{[Back to Contents]}}}}
\label{subsec:mitigate_causal_graph}

Existing counterfactual generation frameworks typically assume that a pre-defined causal graph is provided as part of the input~\citep{pearl2009causality,pawlowski2020deep,yang2021causalvae,pmlr-v202-de-sousa-ribeiro23a,wu2024counterfactual,komanduri2023identifiable,komanduri2024causal,rasal2025diffusion,xia2025decoupled}. 
This assumption enables active interventions and facilitates comparable evaluation across methods, but it can be restrictive in real-world scenarios where the causal graph may be misspecified, partially known, or entirely unavailable. 
To examine this issue, we conduct two complementary studies: 
(1) a causal discovery study that explores whether Causal-Adapter can learn the causal structure from data, and 
(2) a graph ablation study that evaluates the effect of using no graph, an incorrect graph, and the correct graph.

\paragraph{Causal Discovery as a Mitigation Study.}
Causal-Adapter is modular and can be extended with minimal modification to jointly learn the causal graph. 
Instead of fixing the adjacency matrix $A$, we treat $A$ as a learnable parameter initialized to zeros. 
Following differentiable DAG-based methods such as NOTEARS~\citep{zheng2018dags} and DAGMA~\citep{bello2022dagma}, we impose an acyclicity constraint, such as a log-determinant penalty, directly on $A$. 
This requires no additional networks or architectural changes, but only introduces an extra structural loss term.

\begin{figure}[!htbp]
    \centering
    \includegraphics[width=0.85\linewidth]{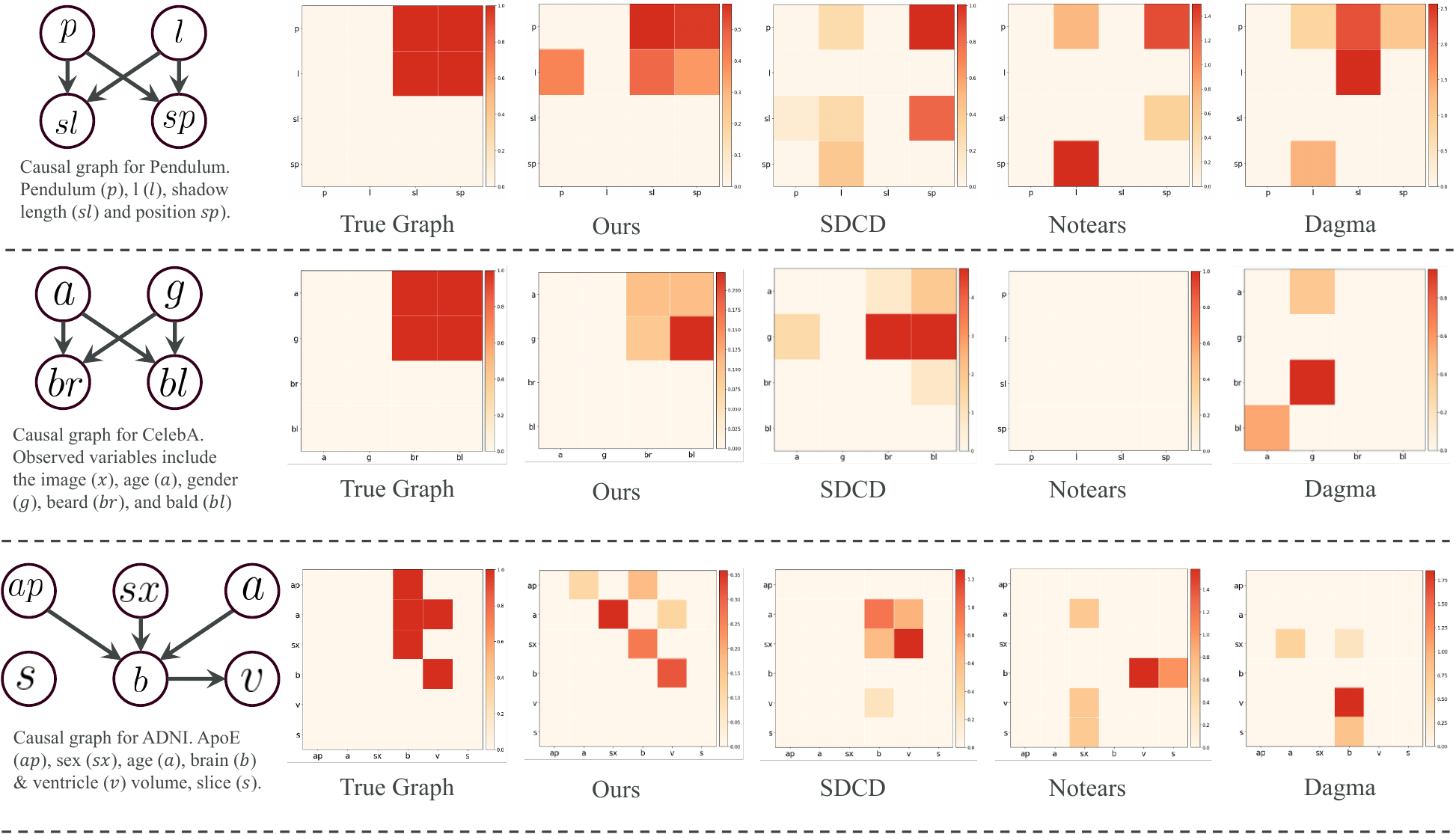}
    \caption{
    Causal discovery performance of Causal-Adapter compared with SDCD, NOTEARS, and DAGMA across three benchmarks.  
    Using the predefined graph as ground truth, our method recovers more true edges than competing methods, demonstrating the potential to unify causal discovery and counterfactual generation within a single framework.
    }
    \label{fig:causal_discovery_comparison}
\end{figure}

\begin{figure}[!htbp]
    \centering
    \includegraphics[width=0.81\linewidth]{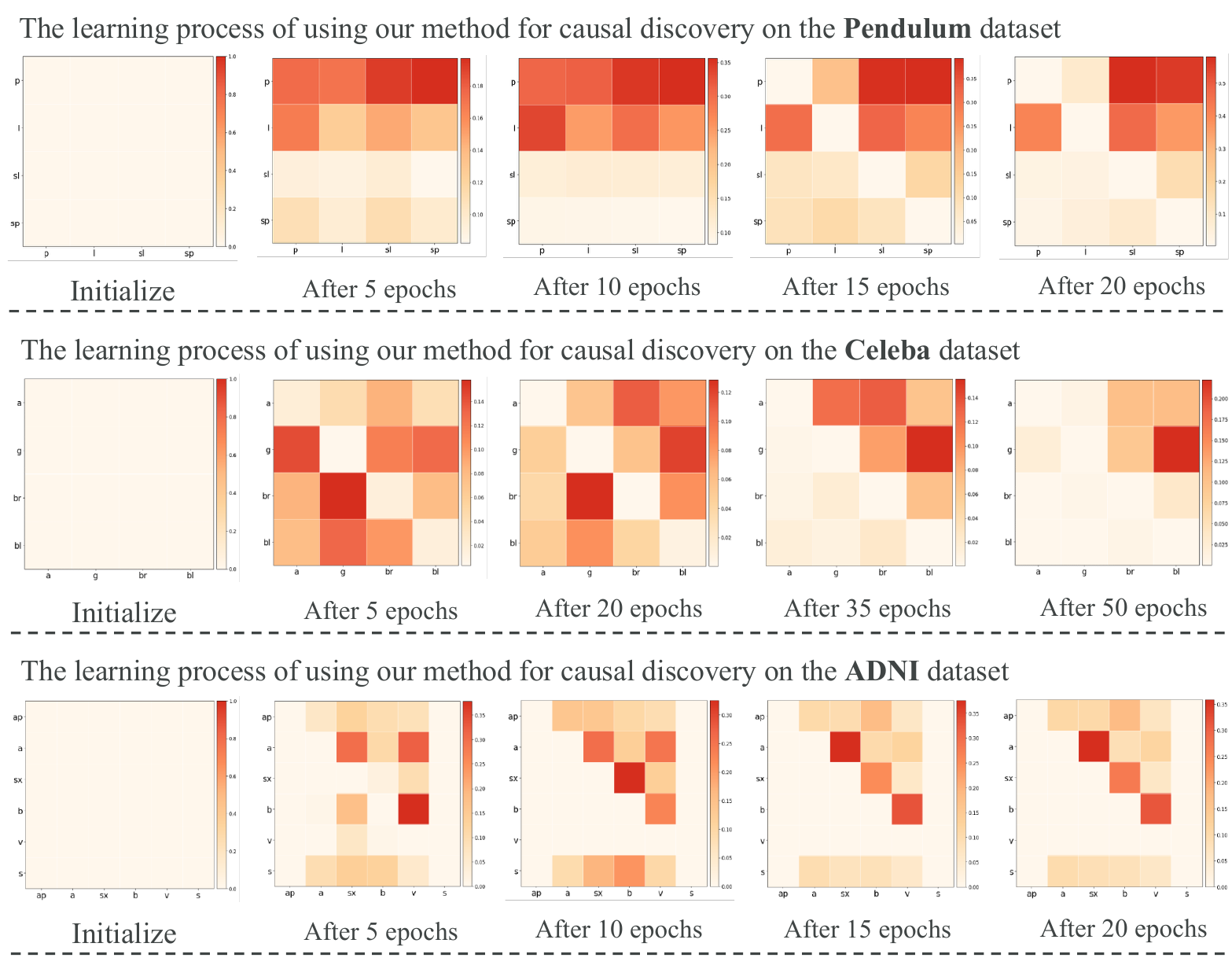}
    \caption{
    Learning trajectory of the adjacency matrix $A$.  
    As training progresses, the learned graph progressively converges toward the ground-truth structure.  
    A fixed threshold of 0.1 is applied across benchmarks for fair comparison.
    }
    \label{fig:causal_discovery_learning}
\end{figure}

As shown in Figure~\ref{fig:causal_discovery_comparison}, this extension enables Causal-Adapter to perform causal structure learning and achieve competitive or superior performance compared with state-of-the-art differentiable causal discovery methods, including SDCD~\citep{nazaret2024stable}. 
Across three benchmark settings, our method recovers more true edges than competing baselines. 
The learning dynamics of $A$ are further visualized in Figure~\ref{fig:causal_discovery_learning}, showing stable convergence toward the ground-truth graph and demonstrating interpretable behavior. 
Nevertheless, consistent with prior studies~\citep{nazaret2024stable,olko2025since}, recovering the full causal graph from purely observational data, such as CelebA or ADNI, remains fundamentally challenging due to the lack of interventional signals. 
These results suggest that Causal-Adapter has the potential to integrate causal discovery and counterfactual generation within a single, simple, and efficient framework.

\paragraph{Ablation on Causal Graph Specification.}
We further evaluate the robustness of Causal-Adapter under different causal graph specifications. 
Specifically, we consider three settings: 
(1) without SCM, 
(2) with an incorrect causal graph, denoted as SCM1, and 
(3) with the correct causal graph, denoted as SCM2. 
This study examines whether the framework can still perform disentangled editing without an explicit graph, and whether the correct graph improves graph-faithful counterfactual generation.

\begin{figure}[!htbp]
    \centering
    \includegraphics[width=1\linewidth]{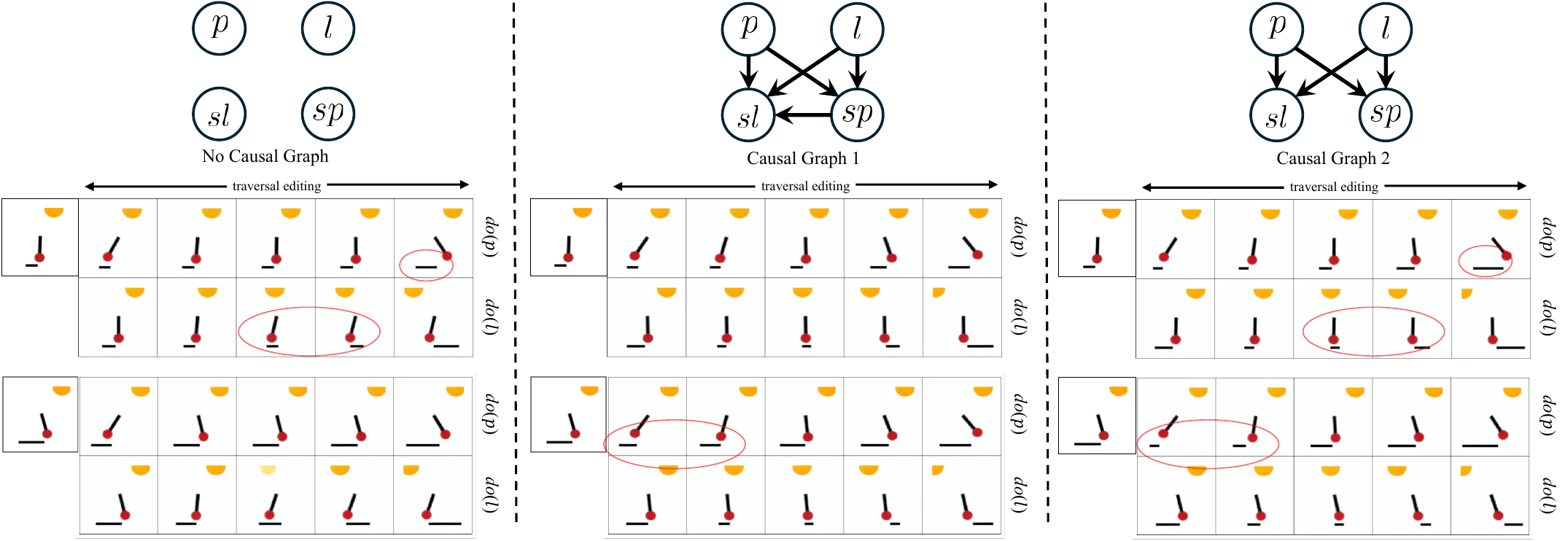}
    \caption{
    Pendulum counterfactuals under three graph settings: no causal graph, an incorrect causal graph (SCM1), and the correct causal graph (SCM2).
    }
    \label{fig:non_causal_pendulum}
\end{figure}

For the Pendulum dataset, qualitative results are shown in Figure~\ref{fig:non_causal_pendulum}, and quantitative results are reported in Table~\ref{tab:pendulum_graph_ablation_appendix}. 
Without SCM, interventions can still modify shadow-related variables because pendulum, light, and shadow attributes are strongly correlated in the training data. 
However, these changes are primarily correlation-driven rather than graph-faithful: intervening on one factor may spuriously affect another non-intervened factor, such as changing the light while also moving the pendulum. 
With SCM1, the incorrectly specified causal edge leads to an inconsistent propagation pattern and produces a different shadow evolution behavior. 
In contrast, with the correct graph SCM2, the intervention effects propagate more consistently with the underlying physical mechanisms.

\setcounter{rownumbers}{0}
\begin{table}[ht!]
\centering
\caption{
Ablation on causal graph specification for the Pendulum dataset. 
We report MAE under interventions on pendulum $do(p)$ and light $do(l)$. 
Lower values indicate better intervention effectiveness.
}
\label{tab:pendulum_graph_ablation_appendix}
\scriptsize
\resizebox{\textwidth}{!}{%
\begin{tabular}{rlcccccccc}
\toprule
&\multirow{2}{*}{Method}
& \multicolumn{2}{c}{\textbf{Pendulum $(p)$ MAE $\downarrow$}}
& \multicolumn{2}{c}{\textbf{Light $(l)$ MAE $\downarrow$}}
& \multicolumn{2}{c}{\textbf{Shadow Length $(sl)$ MAE $\downarrow$}}
& \multicolumn{2}{c}{\textbf{Shadow Position $(sp)$ MAE $\downarrow$}} \\
\cmidrule(lr){3-4}
\cmidrule(lr){5-6}
\cmidrule(lr){7-8}
\cmidrule(lr){9-10}
&& $do(p)$ & $do(l)$
& $do(p)$ & $do(l)$
& $do(p)$ & $do(l)$
& $do(p)$ & $do(l)$ \\
\midrule

\rownum &$\text{w/o~SCM}$
& 0.159 & 0.183
& 0.060 & 0.173
& 0.143 & 0.235
& 0.086 & 0.155 \\

\rownum &$\text{w/~SCM1}$
& \textbf{0.011} & 0.040
& 0.048 & 0.062
& 0.039 & 0.104
& 0.031 & \textbf{0.033} \\

\rownum &$\text{w/~SCM2}$
& 0.014 & \textbf{0.035}
& \textbf{0.045} & \textbf{0.041}
& \textbf{0.028} & \textbf{0.051}
& \textbf{0.030} & \textbf{0.033} \\

\bottomrule
\end{tabular}
}
\end{table}

\begin{figure}[!htbp]
    \centering
    \includegraphics[width=1\linewidth]{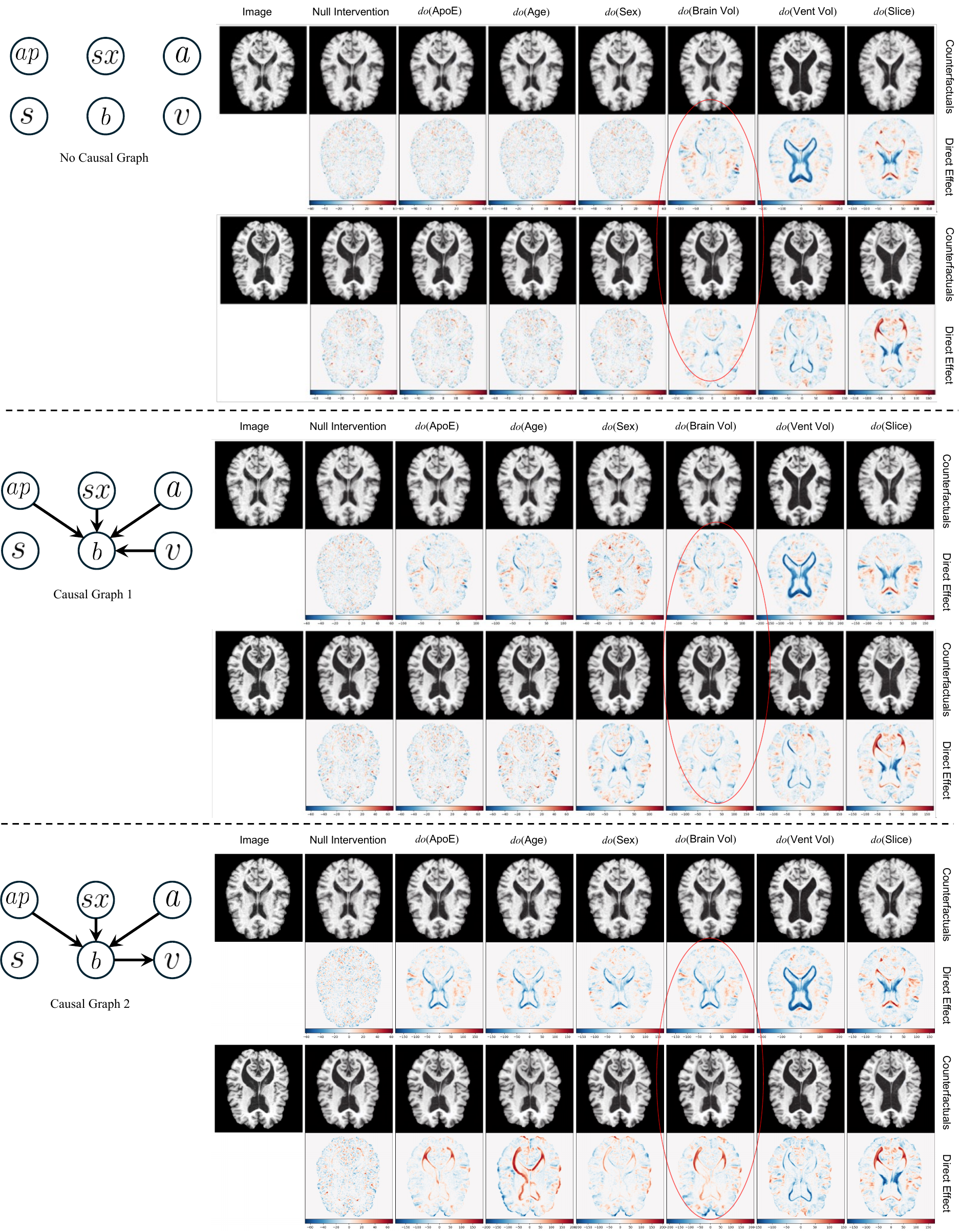}
    \caption{
    ADNI counterfactuals under three graph settings: no causal graph, an incorrect causal graph (SCM1), and the correct causal graph (SCM2).
    }
    \label{fig:non_causal_adni}
\end{figure}

We also perform a graph ablation study on ADNI. 
The qualitative results are shown in Figure~\ref{fig:non_causal_adni}, and the quantitative results are reported in Table~\ref{tab:adni_graph_ablation_appendix}. 
In the non-causal setting, intervening on brain volume rarely affects the ventricular region. 
With SCM1, where the edge between brain volume and ventricular volume is incorrectly specified, interventions on parent factors no longer clearly influence ventricular volume and contour. 
By contrast, the correct graph better preserves the expected propagation pattern between brain and ventricular structures.

\setcounter{rownumbers}{0}
\begin{table}[ht!]
\centering
\caption{
Ablation on causal graph specification for ADNI. 
We report MAE for brain and ventricular volume under interventions on brain volume $do(b)$, ventricular volume $do(v)$, and sex $do(s)$. 
Lower values indicate better intervention effectiveness.
}
\label{tab:adni_graph_ablation_appendix}
\scriptsize
\resizebox{0.85\textwidth}{!}{%
\begin{tabular}{rlcccccc}
\toprule
&\multirow{2}{*}{Method}
& \multicolumn{3}{c}{\textbf{Brain $(b)$ MAE $\downarrow$}}
& \multicolumn{3}{c}{\textbf{Ventricle $(v)$ MAE $\downarrow$}} \\
\cmidrule(lr){3-5}
\cmidrule(lr){6-8}
&& $do(b)$ & $do(v)$ & $do(s)$
& $do(b)$ & $do(v)$ & $do(s)$ \\
\midrule

\rownum &$\text{w/o~SCM}$
& 0.142 & \textbf{0.110} & 0.110
& \textbf{0.028} & \textbf{0.030} & 0.030 \\

\rownum &$\text{w/~SCM1}$
& 0.145 & 0.144 & \textbf{0.030}
& 0.032 & 0.032 & \textbf{0.029} \\

\rownum &$\text{w/~SCM2}$
& \textbf{0.090} & \textbf{0.110} & 0.110
& 0.030 & \textbf{0.030} & 0.030 \\

\bottomrule
\end{tabular}
}
\end{table}

Overall, these studies show that our design deliberately separates SCM modeling from conditional attribute learning and disentanglement. 
This allows the framework to adapt to different causal graphs at inference time by replacing the SCM, without retraining the diffusion backbone. 
When no causal graph is used, Causal-Adapter can still function as a disentangled editing method. 
When a reliable causal graph is available, the generated counterfactuals better match the underlying data-generating structure and achieve stronger quantitative performance.

\newpage

\vspace{-5mm}

\end{document}